\documentclass[letterpaper]{article} 
\usepackage[]{aaai25}  
\usepackage{times}  
\usepackage{helvet}  
\usepackage{courier}  
\usepackage[hyphens]{url}  
\usepackage{graphicx} 
\usepackage{natbib}  
\usepackage{caption} 
\usepackage{algorithm}
\usepackage[noend]{algorithmic}

\usepackage{newfloat}
\usepackage{listings}
\DeclareCaptionStyle{ruled}{labelfont=normalfont,labelsep=colon,strut=off} 
\floatstyle{ruled}
\newfloat{listing}{tb}{lst}{}
\floatname{listing}{Listing}
\usepackage[dvipsnames]{xcolor}

\newcommand{\algorithmicinput}{\textbf{Input:}}
\newcommand{\algorithmicoutput}{\textbf{Output:}}
\newcommand{\INPUT}{\item[\algorithmicinput]}
\newcommand{\OUTPUT}{\item[\algorithmicoutput]}
\newcommand{\algorithmicassumptions}{\textbf{Assumptions:}}
\newcommand{\ASSUMPTIONS}{\item[\algorithmicassumptions]}

\usepackage{amsmath}
\usepackage{amssymb}
\usepackage{mathtools}
\usepackage{amsthm}
\theoremstyle{plain}
\newtheorem{theorem}{Theorem}

\theoremstyle{definition}
\newtheorem{definition}{Definition}

\newtheorem{remark}{Remark}

\usepackage{framed} 

\usepackage{tikz}
\usetikzlibrary {arrows.meta,arrows,patterns,decorations.pathreplacing,chains,positioning,shapes,shapes.geometric,calc,backgrounds,automata}
\usepackage{pgfplots}
\pgfplotsset{compat=1.15}
\usepackage{pifont}
\usepackage{tikz-cd}

\newcommand{\rebuttal}[1]{
{\color{black} {#1}}}
\newcommand{\revision}[1]{
{\color{black} {#1}}}

\usepackage{enumitem}
\setlist[itemize]{noitemsep, topsep=0pt}
\setlist[enumerate]{noitemsep, topsep=0pt}

\usepackage{amsmath}
\newcommand{\ind}{\perp\!\!\!\perp} 
\newcommand{\nind}{\not\!\perp\!\!\!\perp} 

\newcommand{\g}{\mathcal{G}}
\newcommand{\z}{\mathbf{Z}}
\newcommand{\cde}{\mathbf{A}_{\text{\textsc{DE}}}}

\usepackage{colortbl} 
\usepackage{array} 
\usepackage{adjustbox}  
\usepackage{ragged2e}
\usepackage{multirow} 
\usepackage{booktabs} 
\usepackage{tablefootnote}
\usepackage{anyfontsize}

\title{Local Causal Discovery for Structural Evidence of Direct Discrimination} 

\author {
    Jacqueline Maasch\textsuperscript{\rm 1},
    Kyra Gan\textsuperscript{\rm 1},
    Violet Chen\textsuperscript{\rm 2},
    Agni Orfanoudaki\textsuperscript{\rm 3},
    Nil-Jana Akpinar\textsuperscript{\rm 4}\textsuperscript{\textdagger},
    Fei Wang\textsuperscript{\rm 5}
}
\affiliations {
    \textsuperscript{\rm 1}Cornell Tech\\
    \textsuperscript{\rm 2}Stevens Institute of Technology\\
    \textsuperscript{\rm 3}University of Oxford\\
    \textsuperscript{\rm 4}Amazon AWS AI/ML\\
    \textsuperscript{\rm 5}Weill Cornell Medicine
}

\begin{document}

\maketitle


\begin{abstract}
    Identifying the causal pathways of unfairness is a critical objective for improving policy design and algorithmic decision-making. Prior work in causal fairness analysis often requires knowledge of the causal graph, hindering practical applications in complex or low-knowledge domains. Moreover, global discovery methods that learn causal structure from data can display unstable performance on finite  samples, preventing robust fairness conclusions. To mitigate these challenges, we introduce \textit{local discovery for direct discrimination} (LD3): a method that uncovers structural evidence of direct unfairness by identifying the causal parents of an outcome variable. LD3 performs a linear number of conditional independence tests relative to variable set size, and allows for latent confounding under the sufficient condition that all parents of the outcome are observed. We show that LD3 returns a valid adjustment set (VAS) under a new graphical criterion for the \textit{weighted controlled direct effect}, a qualitative indicator of direct discrimination. LD3 limits unnecessary adjustment, providing interpretable VAS for assessing unfairness. We use LD3 to analyze causal fairness in two complex decision systems: criminal recidivism prediction and liver transplant allocation. LD3 was more time-efficient and returned more plausible results on real-world data than baselines, which took 46$\times$ to 5870$\times$ longer to execute.
\end{abstract}


\section{Introduction}
Fairness holds fundamental importance in
policy design and algorithmic decision-making, especially in high-stakes domains such as healthcare and policing \citep{starke2022fairness,corbett2023measure}. Various criteria have been proposed for measuring unfairness with respect to protected or sensitive attributes, such as gender and ethnicity \citep{verma2018fairness}. Legal doctrines generally differentiate between direct discrimination and indirect or spurious forms of unfairness, codifying the notion that the \textit{mechanism} of unfairness matters \citep{barocas2016big, carey2022causal}. However, fairness criteria based solely on statistical associations cannot disentangle
these mechanisms \citep{kilbertus2017avoiding,makhlouf2020survey}, limiting their informativeness and actionability for policy interventions. Consequently, there is a growing emphasis on applying causal reasoning  in fairness analysis \citep{kilbertus2017avoiding}, shifting the focus from associations to interventions and counterfactual outcomes. 

\textit{Causal fairness analysis} (CFA) provides a theoretical framework for disentangling the mechanisms of unfairness using the language of \textit{structural causal models} (SCMs).
Previous works in CFA \citep{zhang_fairness_2018, plecko_causal_2023} and the allied field of mediation analysis \citep{pearl_interpretation_2014,vanderweele2016mediation} generally assume significant prior structural knowledge in order to decompose direct, indirect, and spurious effects. In practice, structural knowledge is often incomplete, absent, or contentious for complex domains, even among experts \citep{petersen2023constructing}. 
Furthermore, the identifiability of direct and indirect effects has been a topic of extensive debate among theoreticians \citep{pearl_interpretation_2014}, raising barriers to entry for applied researchers \citep{vanderweele_controlled_2011}. Thus, existing methodologies in CFA can be challenging to apply in complex systems.

Among the many fairness measures proposed in CFA, the \emph{controlled direct effect} (CDE) is a 
relatively straightforward
qualitative indicator of direct discrimination that takes non-zero values
only when the exposure is a direct cause of the outcome \citep{zhang_fairness_2018}. The CDE has often been favored in policy evaluation over alternative direct effect measures \citep{vanderweele_controlled_2011,vanderweele2013policy} 
as it is more interpretable for real-world interventions and requires fewer untestable assumptions and less prior structural knowledge \citep{pearl_direct_2001,shpitser_complete_2011}. 
To increase the practicality of CFA in complex domains, we choose to focus on the CDE as a starting point for this work.



\begin{figure}[!t]
    \centering
\begin{tikzpicture}[every edge quotes/.style = {font=\scriptsize, fill=white,sloped}]
  \node[circle,black,thick,draw,scale=0.8] (X) at (0, 0) {$X$};
  \node[circle,black,thick,draw,scale=0.8] (Y) at (2, 0) {$Y$};
  \node[circle,black,thick,draw,scale=0.8] (C) at (1, 1) {$\mathbf{C}$};
  \node[circle,black,thick,draw,scale=0.8] (M) at (1, -1) {$\mathbf{M}$};
  \draw[-{Stealth[width=6.5pt,length=6.5pt]},ultra thick,auto,WildStrawberry]  (X) edge node {} (Y);
  \draw[-{Stealth[width=5pt,length=5pt]},thick,auto,black]  (C) edge node {} (X);
  \draw[-{Stealth[width=5pt,length=5pt]},thick,auto,black]  (C) edge node {} (Y);
  \draw[-{Stealth[width=5pt,length=5pt]},thick,auto,black]  (X) edge node {} (M);
  \draw[-{Stealth[width=5pt,length=5pt]},thick,auto,black]  (M) edge node {} (Y);
  \draw[-{Stealth[width=5pt,length=5pt]},thick,auto,black]  (C) edge node {} (M);
  \draw[{Stealth[width=5pt,length=5pt]}-{Stealth[width=5pt,length=5pt]},thick,gray!80,dotted] (M) edge[bend left=20] (C) node[] {};
  \draw[{Stealth[width=5pt,length=5pt]}-{Stealth[width=5pt,length=5pt]},thick,gray!80,dotted] (X) edge[bend left=50] (C) node[] {};
  \draw[{Stealth[width=5pt,length=5pt]}-{Stealth[width=5pt,length=5pt]},thick,gray!80,dotted] (C) edge[bend left=50] (Y) node[] {};
  \draw[{Stealth[width=5pt,length=5pt]}-{Stealth[width=5pt,length=5pt]},thick,gray!80,dotted] (X) edge[bend right=50] (M) node[] {};
  \draw[{Stealth[width=5pt,length=5pt]}-{Stealth[width=5pt,length=5pt]},thick,gray!80,dotted] (M) edge[bend right=50] (Y) node[] {}; 
  \draw[{Stealth[width=5pt,length=5pt]}-{Stealth[width=5pt,length=5pt]},thick,gray!80,dotted] (X) edge[bend left=20] (Y) node[] {};
  \node[black,below=of M,yshift=0.8cm] {\textsc{Direct}};
\end{tikzpicture}
\hspace{1mm}
\begin{tikzpicture}[every edge quotes/.style = {font=\scriptsize, fill=white,sloped}]
  \node[circle,black,thick,draw,scale=0.8] (X) at (0, 0) {$X$};
  \node[circle,black,thick,draw,scale=0.8] (Y) at (2, 0) {$Y$};
  \node[circle,black,thick,draw,scale=0.8] (C) at (1, 1) {$\mathbf{C}$};
  \node[circle,black,thick,draw,scale=0.8] (M) at (1, -1) {$\mathbf{M}$};
  \draw[-{Stealth[width=5pt,length=5pt]},thick,auto,black]  (X) edge node {} (Y);
  \draw[-{Stealth[width=5pt,length=5pt]},thick,auto,black]  (C) edge node {} (X);
  \draw[-{Stealth[width=5pt,length=5pt]},thick,auto,black]  (C) edge node {} (Y);
  \draw[-{Stealth[width=6.5pt,length=6.5pt]},ultra thick,auto,WildStrawberry]  (X) edge node {} (M);
  \draw[-{Stealth[width=6.5pt,length=6.5pt]},ultra thick,auto,WildStrawberry]  (M) edge node {} (Y);
  \draw[-{Stealth[width=5pt,length=5pt]},thick,auto,black]  (C) edge node {} (M);
  \draw[{Stealth[width=5pt,length=5pt]}-{Stealth[width=5pt,length=5pt]},thick,gray!80,dotted] (M) edge[bend left=20] (C) node[] {};
  \draw[{Stealth[width=5pt,length=5pt]}-{Stealth[width=5pt,length=5pt]},thick,gray!80,dotted] (X) edge[bend left=50] (C) node[] {};
  \draw[{Stealth[width=5pt,length=5pt]}-{Stealth[width=5pt,length=5pt]},thick,gray!80,dotted] (C) edge[bend left=50] (Y) node[] {};
  \draw[{Stealth[width=5pt,length=5pt]}-{Stealth[width=5pt,length=5pt]},thick,gray!80,dotted] (X) edge[bend right=50] (M) node[] {};
  \draw[{Stealth[width=5pt,length=5pt]}-{Stealth[width=5pt,length=5pt]},thick,gray!80,dotted] (M) edge[bend right=50] (Y) node[] {};
  \draw[{Stealth[width=5pt,length=5pt]}-{Stealth[width=5pt,length=5pt]},thick,gray!80,dotted] (X) edge[bend left=20] (Y) node[] {};
  \node[black,below=of M,yshift=0.8cm] {\textsc{Indirect}};
\end{tikzpicture}
\hspace{1mm}
\begin{tikzpicture}[every edge quotes/.style = {font=\scriptsize, fill=white,sloped}]
  \node[circle,black,thick,draw,scale=0.8] (X) at (0, 0) {$X$};
  \node[circle,black,thick,draw,scale=0.8] (Y) at (2, 0) {$Y$};
  \node[circle,black,thick,draw,scale=0.8] (C) at (1, 1) {$\mathbf{C}$};
  \node[circle,black,thick,draw,scale=0.8] (M) at (1, -1) {$\mathbf{M}$};
  \draw[-{Stealth[width=5pt,length=5pt]},thick,auto,black]  (X) edge node {} (Y);
  \draw[-{Stealth[width=6.5pt,length=6.5pt]},ultra thick,auto,WildStrawberry]  (C) edge node {} (X);
  \draw[-{Stealth[width=6.5pt,length=6.5pt]},ultra thick,auto,WildStrawberry]  (C) edge node {} (Y);
  \draw[-{Stealth[width=5pt,length=5pt]},thick,auto,black]  (X) edge node {} (M);
  \draw[-{Stealth[width=5pt,length=5pt]},thick,auto,black]  (M) edge node {} (Y);
  \draw[-{Stealth[width=5pt,length=5pt]},thick,auto,black]  (C) edge node {} (M);
  \draw[{Stealth[width=5pt,length=5pt]}-{Stealth[width=5pt,length=5pt]},thick,gray!80,dotted] (M) edge[bend left=20] (C) node[] {};
  \draw[{Stealth[width=5pt,length=5pt]}-{Stealth[width=5pt,length=5pt]},thick,gray!80,dotted] (X) edge[bend left=50] (C) node[] {};
  \draw[{Stealth[width=5pt,length=5pt]}-{Stealth[width=5pt,length=5pt]},thick,gray!80,dotted] (C) edge[bend left=50] (Y) node[] {};
  \draw[{Stealth[width=5pt,length=5pt]}-{Stealth[width=5pt,length=5pt]},thick,gray!80,dotted] (X) edge[bend right=50] (M) node[] {};
  \draw[{Stealth[width=5pt,length=5pt]}-{Stealth[width=5pt,length=5pt]},thick,gray!80,dotted] (M) edge[bend right=50] (Y) node[] {};
  \draw[{Stealth[width=5pt,length=5pt]}-{Stealth[width=5pt,length=5pt]},thick,gray!80,dotted] (X) edge[bend left=20] (Y) node[] {};
  \node[black,below=of M,yshift=0.8cm] {\textsc{Spurious}};
\end{tikzpicture}
    \caption{The \textit{standard fairness model} (SFM) is compactly represented as a local subgraph around protected attribute $X$ and outcome $Y$ 
    \citep{plecko_causal_2024}. 
    Variables that are irrelevant to CFA
    are abstracted away, leaving confounders ($\mathbf{C}$) and mediators ($\mathbf{M}$). 
    Directed edges represent the existence of active directed paths, and
    bidirected edges denote potential confounding. This work aims to identify \textit{direct} mechanisms of unfairness in a data-driven way. 
    }
    \label{fig:sfm_de_ie_se}
\end{figure}

In low-knowledge domains, we can support CFA by learning causal structure directly from observational data. However, \textit{global causal discovery} 
is challenging in finite data due to high sample complexity \citep{spirtes2001causation} 
and exponential time complexity in unconstrained search spaces \citep{chickering_large-sample_2004,claassen_learning_2013}. Learned causal graphs often disagree with expert knowledge in complex domains \citep{shen_challenges_2020,petersen2023constructing}, and can yield conflicting causal fairness conclusions in CFA (\citealt{binkyte2023causal} and
Section \ref{sec:causal_fairness_analysis} of this paper). 

While global discovery learns the relations among all observed variables,
\emph{local causal discovery} 
only learns the substructures relevant for downstream tasks, such as causal effect estimation \citep{gupta_local_2023, maasch2024local, shah2024front} 
or feature selection \citep{yu_causality-based_2021, yu_unified_2021}.  
As the \emph{standard fairness model} (SFM; Figure \ref{fig:sfm_de_ie_se}) is represented as a local subgraph 
\citep{plecko_causal_2024}, local discovery offers a natural framework for CFA. 
However, since task-specific local discovery algorithms are definitionally one-size-\textit{does}-\textit{not}-fit-all, existing methods may not be optimal for fairness tasks.

\paragraph{Contributions} 
This work aims to increase the practicality of CFA for direct discrimination in complex domains with unknown causal graphs. Our contributions are three-fold.
\begin{enumerate}
    \item \textbf{Local discovery for direct discrimination (LD3)}. This local causal discovery method leverages the problem structure in CFA to efficiently detect graphical signatures of direct discrimination.\footnote{Code on GitHub: \texttt{https://github.com/jmaasch/LD3}} LD3 discovers the parents of an outcome variable in a linear number of conditional independence tests with respect to variable set size. 
    \item \textbf{A graphical criterion for the weighted controlled direct effect (WCDE).} This criterion is sufficient to identify a valid adjustment set (VAS) for the WCDE, a qualitative indicator of direct discrimination. This criterion is satisfied by the knowledge returned by LD3.
    \item \textbf{Real-world fairness analysis.} We deploy LD3 for two fairness problems: (1) racial discrimination in recidivism prediction and (2) sex-based discrimination in liver transplant allocation. LD3 recovered more plausible causal relations than local and global baselines, which performed 11–1021$\times$ more tests and took 46–5870$\times$ longer to run.
\end{enumerate}




\section{Preliminaries} \label{sec:prelim}
Let capital letters denote univariate random variables (e.g., $X$), with their values in lowercase (e.g., $X = x$). Multivariate random variables or sets are denoted by boldface capital letters (e.g., $\mathbf{X}$), with vector values in bold lowercase (e.g., $\mathbf{X} = \mathbf{x}$). Graphs or function sets are denoted by calligraphic letters (e.g., $\mathcal{F}$). Let $pa(\cdot)$ and $de(\cdot)$ denote the parent and descendant sets for a variable in causal graph $\mathcal{G}$, respectively. 

\subsection{Causal Fairness Analysis}

CFA can be framed in the language of SCMs and their graphical representations \citep{plecko_causal_2024}. 

\begin{definition}[Structural causal model,
\citealt{bareinboim2022pearl}] \label{def:scm}
    An SCM is a 4-tuple $\langle \mathbf{V},\mathbf{U},\mathcal{F}, p(\mathbf{u}) \rangle$ where $\mathbf{U} = \{U_i\}_{i=1}^n$ denotes a set of exogenous variables determined by factors external to the model, $\mathbf{V} = \{V_i\}_{i=1}^n$ denotes a set of observed endogenous variables determined by $\mathbf{U} \cup \mathbf{V}$,  $\mathcal{F} = \{f_i\}_{i=1}^n$ denotes a set of structural functions
    such that $V_i = f_i(pa(V_i), U_i)$, and $p(\mathbf{u})$ is the distribution over $\mathbf{U}$.
\end{definition}

Every SCM corresponds to a causal graphical model. To facilitate CFA, the true causal graph for an SCM can be 
compactly represented using the SFM (Figure \ref{fig:sfm_de_ie_se}). 

\begin{definition}[Standard fairness model, \citealt{plecko_causal_2024}] \label{def:sfm}
    Let $\mathcal{G} = (\mathbf{V}, \mathbf{E})$ be a causal graph with vertices $\mathbf{V}$ and edges $\mathbf{E}$. Let $\mathcal{G}_\mathrm{SFM}$ be the \emph{projection} of $\mathcal{G}$ onto the SFM, which is obtained by (1) selecting a protected attribute-outcome pair $\{X,Y\} \subset \mathbf{V}$ and (2) identifying sets $\mathbf{M},\mathbf{C} \subseteq \mathbf{V} \setminus \{X,Y\}$ that meet the following conditions:
    \begin{itemize}
        \item $\mathbf{M}$ is the set of mediators with respect to $X$ and $Y$;
        \item $\mathbf{C}$ is the set of confounders with respect to $X$ and $Y$. 
    \end{itemize}
    Note that $\mathbf{C}$ or $\mathbf{M}$ can be the empty set, and confounding can exist among $\mathbf{C}$, $\mathbf{M}$, $X$, $Y$ (Figure \ref{fig:sfm_de_ie_se}). 
\end{definition}

\paragraph{Structural Fairness Criteria}

Multiple criteria have been proposed for evaluating fairness from graphical structures.
Here, we focus on a criterion for direct discrimination.

\begin{definition}[Structural direct criterion (SDC), 
\citealt{plecko_causal_2024}] \label{def:sdc}
An SCM is fair 
with respect to
direct discrimination if and only if the SDC evaluates to 0:
    \begin{align}
        SDC &= 
        \begin{cases}
            1 \quad \text{if $X$ is a parent of $Y$,} \\
            0 \quad \text{if $X$ is not a parent of $Y$.}
        \end{cases}
    \end{align}
\end{definition}

\subsection{Controlled Direct Effect}
\label{sec:cde_preliminaries}

The CDE can be used to test
for 
direct discrimination, as the true value is non-zero if and only if there is a direct path from the protected attribute to the outcome \citep{zhang_fairness_2018}. Let $X = x$ and $X = x^*$ be the exposure values corresponding to treatment and no treatment, 
respectively.

\revision{
\begin{definition}[CDE,
\citealt{pearl_interpretation_2014}] 
\label{def:cde}
The CDE measures the expected change in outcome as the exposure changes when mediators $\mathbf{M}$ are uniformly fixed to a constant value $\mathbf{m}$:
    \begin{align}
        \mathrm{CDE}(\mathbf{m}) &\coloneqq 
        \mathbb{E}[Y|do(x,\mathbf{m})] - \mathbb{E}[Y|do(x^*, \mathbf{m})].
    \end{align}
\end{definition}
}

\revision{
\begin{definition}[Identifiability conditions of the CDE, \citealt{vanderweele_controlled_2011}] \label{def:id_cde}


When we have access to a covariate set $\mathbf{S}$ that controls for confounding of both $\{X,Y\}$ and $\{\mathbf{M},Y\}$, 
the CDE is identifiable under the following conditions. Let $Y_{\check{x},\check{\mathbf{m}}}$ denote the value of $Y$ under $X=\check{x},\mathbf{M}=\check{\mathbf{m}}$. 
\begin{enumerate}[leftmargin=*, itemsep=0pt, parsep=0pt]
    \item There is no latent confounding of the exposure and outcome given $\mathbf{S}$, 
    i.e., $Y_{\check{x},\check{\mathbf{m}}} \ind X | \mathbf{S}$ for all $\check{x}, \check{\mathbf{m}}$.
    \item There is no latent confounding of the mediators and outcome given $\{X,\mathbf{S}\}$, i.e., $Y_{\check{x},\check{\mathbf{m}}} \ind \mathbf{M} | X, \mathbf{S}$ for all $\check{x}, \check{\mathbf{m}}$.
\end{enumerate}
Then, we obtain $CDE(\mathbf{m})$ by\footnote{When $\mathbf{S}$ is not sufficient for valid control of confounding for both $\{X,Y\}$ and $\{\mathbf{M},Y\}$, alternative formulae may be required \citep{vanderweele_controlled_2011,pearl_interpretation_2014}. 
}  
\begin{align}
     \sum_{\mathbf{s}} \big( \mathbb{E}[Y \mid x, \mathbf{s}, \mathbf{m}] - \mathbb{E}[Y \mid x^*, \mathbf{s}, \mathbf{m}] \big) P(\mathbf{s}). \label{eq:reduced_cde}
\end{align} 
Note that the subset of mediators that are parents of $Y$ is sufficient for CDE identification \citep{pearl_direct_2001}.
\end{definition}
}

\paragraph{Comparison to Alternative Measures} Several estimands can capture direct effects, including
CDE \citep{pearl_direct_2001}, natural direct effect (NDE; \citealt{pearl_direct_2001}), and  counterfactual direct effect (Ctf-DE; \citealt{zhang_fairness_2018}). While CDE is an interventional quantity, NDE and Ctf-DE are counterfactual quantities. 
In this work, we favor  CDE 
as it requires fewer untestable assumptions over the data generating process \citep{shpitser_complete_2011} and less structural knowledge than NDE and Ctf-DE \citep{pearl_direct_2001, vanderweele_controlled_2011,zhang_fairness_2018}. The main objective of this work is to assess whether the protected attribute is a direct cause of the outcome. \textit{The CDE, NDE, and Ctf-DE are all non-zero if and only if 
$X$ directly points toward $Y$.}
Thus, we advocate for this simpler estimand for practicality. See Appendix \ref{appendix:direct_effects} for an extended comparison of estimands.

\revision{
\begin{remark}[Assumptions on Confounding]
\label{remark:assumptions_confounding}
    Unlike the NDE, CDE identification does not forbid the existence of confounders for $\{\mathbf{M},Y\}$ that are descended from $X$ \citep{pearl_interpretation_2014}. 
As done previously \citep{vanderweele_conceptual_2009}, this work assumes structures in which adjusting for confounders of $\{\mathbf{M},Y\}$ does not induce post-treatment bias, allowing us to identify the CDE with the expression provided in Equation \ref{eq:reduced_cde}. If this assumption does not hold, unbiased CDE estimates can still be obtained given  alternative estimators \citep{petersen_estimation_2006}. 
\end{remark}
}

 
\paragraph{Weighted CDE} When interaction between the mediator and exposure exists, 
the value of CDE may vary across different mediator values
\citep{pearl_interpretation_2014}. 
To avoid assumptions about interaction while still obtaining a unique estimate, we define the \textit{weighted CDE} (WCDE) as the following expectation over $\mathbf{M}$. For simplicity,
we assume discrete $\mathbf{M}$.\footnote{Definition~\ref{def:cde_bar} generalizes to continuous $\mathbf{M}$ by integrating over its possible values. We note that LD3 operates independently of the variable type of $\mathbf{M}$.}  

\revision{
\begin{definition}[WCDE,
\citealt{pearl2000models}\footnote{Though unnamed, this expression is noted on page 131.}] \label{def:cde_bar}
We define WCDE as
\begin{align}
    \sum_{\mathbf{m}'} \big( \mathbb{E}[Y \mid do(x,\mathbf{m}')] - \mathbb{E}[Y \mid do(x^*, \mathbf{m}')]\big)  P(\mathbf{m}'),
\end{align}
where $\mathbf{M}' \subseteq \mathbf{M}$ are parents of $Y$. Per Equation \ref{eq:reduced_cde}, WCDE is identifiable as
    \begin{align}
        \hspace{-0.5mm}\sum_{\mathbf{m}'} \sum_{\mathbf{s}}  \big( \mathbb{E}[Y|x, \mathbf{s}, \mathbf{m}'] - \mathbb{E}[Y | x^*, \mathbf{s}, \mathbf{m}'] \big)  P(\mathbf{s}) P(\mathbf{m}'). \label{eq:wcde}
    \end{align} 
\end{definition}
}


\begin{definition}[VAS for the WCDE] \label{def:valid}
    Given Definitions \ref{def:id_cde} and \ref{def:cde_bar}, a VAS for WCDE 
    estimation blocks (1) all backdoor paths for $\{X,Y\}$, (2) all backdoor paths for $\{\mathbf{M},Y\}$, and (3) all mediator paths for $\{X,Y\}$. 
\end{definition}

\paragraph{As a Fairness Metric}
Similar to CDE, WCDE indicates direct discrimination when its value is non-zero. However, a zero value \emph{does not} guarantee the absence of direct discrimination, as different CDE values may cancel each other out. Thus, we encourage caution when interpreting zero values.



\subsection{Mapping Causal Partitions to the SFM}
The methods introduced in this work leverage the \textit{causal partition} taxonomy defined in \citet{maasch2024local} (Table \ref{tab:partitions}). Given an exposure-outcome pair $\{X,Y\}$, any arbitrary variable set $\z$ can be uniquely partitioned into eight disjoint subsets (which may be empty) that are defined by the types of causal paths that they share with $X$ and $Y$.
By focusing structure learning
on relationships that are \textit{causally relevant} to the exposure and outcome, this partition taxonomy provides practical building blocks for local discovery. Mapping this partition taxonomy to the SFM, $\z_1$ are confounders $\mathbf{C}$ (and their proxies) and $\z_3$ are mediators $\mathbf{M}$ (and their proxies).\footnote{See \citet{maasch2024local} for more formal partition definitions. Proxy variables are not relevant in this setting,  as these cannot be parents of $Y$ and are not returned by LD3.
} 
When referring to the union of multiple partitions, we use notation of the form $\z_{1,3} \coloneqq \z_1 \cup \z_3$.

\setlength{\intextsep}{0pt}%
\begin{table}[!t]
    \centering
    \begin{adjustbox}{max width=0.48\textwidth}
    \begin{tabular}{p{0.25cm} p{8.7cm}}
    \toprule
    \multicolumn{2}{c}{\fontfamily{cmr}\textsc{Exhaustive, Disjoint Causal Partitions w.r.t. $\{X,Y\}$}} \\
    \midrule
        $\z_1$ &  Confounders and their proxies. \\
        $\z_2$ &  Colliders and their proxies.\\
        $\z_3$ &  Mediators and their proxies. \\
        $\z_4$ &  Non-descendants of $Y$ where $\z_4 \ind X$ and $\z_4 \nind X | Y$. \\
        $\z_5$ &  Instruments and their proxies. \\
        $\z_6$ &  Descendants of $Y$ s.t. active paths with $X$ are mediated by $Y$. \\
        $\z_7$ &  Descendants of $X$ s.t. active paths with $Y$ are mediated by $X$. \\
        $\z_8$ &  Nodes that share no active paths with $X$ nor $Y$. \\
    \bottomrule
    \end{tabular}
    \end{adjustbox}
    \vspace{-3mm}
    \caption{Adapted from \citet{maasch2024local}.}
    \label{tab:partitions}
\end{table}


\section{Local Discovery for Direct Discrimination} \label{sec:method}

In CFA, we can translate the fairness query \textit{``Is direct discrimination present?"} into the graphical query \textit{``Is the protected attribute a parent of the outcome?"} To answer this graphical query, we focus on two indicators of parentage: the SDC and WCDE, where the former can be directly answered by LD3 and the latter can be estimated by using the VAS returned by LD3.

The input to LD3 (Algorithm~\ref{alg:a_cde}) is a variable set $\z$ of unknown causal relation to the protected attribute $X$ and outcome $Y$. 
Instead of learning the causal graph, LD3 learns \textit{causal partition labels} (Table \ref{tab:partitions}). This local learning approach abstracts away structural information that is impertinent to direct discrimination detection, resulting in a computationally efficient discovery procedure. LD3 performs sequential CI tests to iteratively discover the partition label ($\widehat{\z}$) of each variable in $\z$. Leveraging these labels, LD3 uses a test of \textit{d}-separation to evaluate SDC (Lines 12–13). \revision{By conditioning only on $\widehat{\z}_{1,3\in pa(Y)}$, this CI test offers sample efficiency benefits relative to conditioning on all $\z$.} Additionally, LD3 returns a VAS for WCDE containing $pa(Y) \setminus X$.

\paragraph{Time Complexity}
Constraint-based discovery methods are typically analyzed by the number of CI tests performed \citep{spirtes2001causation}. 
For-loops at Lines 2–5, 7–8, and 9–10 of Algorithm \ref{alg:a_cde} perform $O(|\z|)$ tests each. All remaining lines perform a constant number of operations. Thus, the total number of CI tests is of $O(|\z|)$,  ensuring scalability in real-world CFA.

\begin{algorithm}[!t]
\caption{\textit{LD3} 
} \label{alg:a_cde}
    \begin{algorithmic}[1]
        \INPUT Exposure $X$, outcome $Y$, variable set $\z$, CI test of choice, significance level $\alpha$.
        \OUTPUT Adjustment set $\cde$, SDC results.
        \ASSUMPTIONS Sufficient conditions \ref{assumption:y_no_desc} and \ref{assumption:all_parents_y}.
        \vspace{3mm}
        \STATE $\z' \gets \z$
        \FOR{$\forall \; Z \in \z'$}
            \STATE \textbf{if} $Z \ind X \land Z \ind Y$ \textbf{then} $Z \in \widehat{\z}_8$ 
            \STATE \textbf{if} $Z \nind Y \land Z \ind Y | X$ \textbf{then} $Z \in \widehat{\z}_{5,7}$ 
            \STATE \textbf{if} $Z \ind X \land Z \nind X | Y$ \textbf{then} $Z \in \widehat{\z}_4$ 
        \ENDFOR
        \STATE $\z' \gets \z' \setminus \widehat{\z}_8 \cup \widehat{\z}_{5,7} \cup \widehat{\z}_4$
        \FOR{$\forall \; Z \in \z'$}
            \STATE \textbf{if} $Z \nind Y | X \cup \widehat{\z}_{4} \cup \{\z' \setminus Z\}$ 
             \\
            \textbf{then} $Z \in \widehat{\z}_{1\in pa(Y)} \cup \widehat{\z}_{3\in pa(Y)}$  
        \ENDFOR
        \FOR{$\forall \; \widehat{Z}_4 \in \widehat{\z}_4$}
            \STATE \textbf{if} $\widehat{Z}_4 \nind Y | X \cup \widehat{\z}_{1\in pa(Y)} \cup \widehat{\z}_{3\in pa(Y)} \cup \{\widehat{\z}_4 \setminus \widehat{Z}_4\}$ \textbf{then} $\widehat{Z}_4 \in \widehat{\z}_{4\in pa(Y)}$ 
        \ENDFOR 
        \STATE $\cde \gets \widehat{\z}_{1\in pa(Y)} \cup \widehat{\z}_{3\in pa(Y)} \cup \widehat{\z}_{4\in pa(Y)}$ 
        \\
        \STATE \textbf{if} $X \ind Y | \widehat{\z}_{1\in pa(Y)} \cup \widehat{\z}_{3\in pa(Y)}$ \textbf{then} $SDC \gets 0$ 
        \STATE \textbf{else} $SDC \gets 1$
        \RETURN $\cde, SDC$ 
    \end{algorithmic}
\end{algorithm}

\paragraph{Sufficient Conditions for Structure Learning} We assume causal Markov, faithfulness, and acyclicity. We do not impose parametric assumptions on causal functions nor distributional forms. As for all constraint-based methods, the independence test selected may impose its own parametric assumptions. When these are not well-justified, we recommend nonparametric tests (e.g., \citealt{gretton2005measuring,gretton2007kernel}). 

\revision{
\begin{theorem}
\label{theorem:algorithm_correctness}
    Asymptotic guarantees on partitioning and SDC correctness hold under Assumptions \ref{assumption:y_no_desc} and \ref{assumption:all_parents_y}. Given WCDE identifiability by Equation \ref{eq:wcde} (Remark \ref{remark:assumptions_confounding}), \ref{assumption:y_no_desc} and \ref{assumption:all_parents_y} are also sufficient for VAS discovery. 
\end{theorem}
} 

\begin{enumerate}[label=A\arabic*,leftmargin=\widthof{A1}+\labelsep] 
    \item \textit{$Y$ has no descendants in the observed variable set.} This is satisfied when $Y$ is a terminal variable in the temporal ordering (e.g., when outcome is death, a policy or algorithmic decision made at a known time point, etc.). \label{assumption:y_no_desc}
    \item \textit{All parents of $Y$ are observed.} Latent variables that are not parents of $Y$ are permissible. Thus, this is a milder condition than assuming causal sufficiency. \label{assumption:all_parents_y}
\end{enumerate}
\revision{Proof of Theorem \ref{theorem:algorithm_correctness} is in Appendix \ref{appendix:proofs}.} Note that assumptions \ref{assumption:y_no_desc}  and \ref{assumption:all_parents_y} are sufficient but not necessary. \ref{assumption:y_no_desc} has been previously used to facilitate parent and ancestor learning \citep{soleymani_causal_2022, cai2023learning}. LD3 learns causal partitions directly from data, without assumptions on temporal ordering other than 
\ref{assumption:y_no_desc} and
$Y$ cannot cause $X$. 
Assumption \ref{assumption:all_parents_y} is a consequence of the sufficient conditions for CDE identifiability, as Definition~\ref{def:id_cde} requires blocking all spurious and indirect paths into $Y$. \ref{assumption:all_parents_y} allows for unobserved variables that are not in $pa(Y)$, a more relaxed assumption than the causal sufficiency typically required in discovery (e.g., \citealt{spirtes2001causation}; \citealt{zheng2018dags}; etc.). 
We exploit \ref{assumption:all_parents_y} to evaluate the SDC, as it ensures that $X$ is conditionally $d$-separable from $Y$ when there is no direct path from $X$ to $Y$. It also ensures colliders ($\z_{2 \notin de(Y)}$) are removed from the adjustment set. 

We empirically demonstrate failure modes and robustness to violations of \ref{assumption:all_parents_y} in Appendix \ref{sec:robustness}, showing that \ref{assumption:all_parents_y} is sufficient but not necessary. We prove in Appendix \ref{appendix:proofs} that latent variables not adjacent to $Y$ do not impact correctness.  

\begin{theorem}\label{theorem:latent}
    Latent variables that are not parents of $Y$ do not affect Algorithm \ref{alg:a_cde}.
\end{theorem}

\begin{figure}[!t]
    \centering
\begin{tikzpicture}[every edge quotes/.style = {font=\footnotesize, fill=white,sloped}]
  \node[circle,WildStrawberry,very thick,draw,scale=0.7] (X) at (0, 0) {$X$};
  \node[circle,WildStrawberry,very thick,draw,scale=0.7] (Y) at (2, 0) {$Y$};
  \node[circle,black,thick,draw,scale=0.7] (Z1) at (1, 1.5) {$\z_1$};
  \node[circle,black,thick,draw,scale=0.7] (Z3) at (1, -1) {$\z_3$};
  \node[circle,black,thick,draw,scale=0.7] (Z2) at (1, -1.75) {$\z_2$};
  \node[circle,black,thick,draw,scale=0.7] (Z4) at (2, 1.5) {$\z_4$};
  \node[circle,black,thick,draw,scale=0.7] (Z5) at (0, 1.5) {$\z_5$};
  \node[circle,black,thick,draw,scale=0.7] (Z7) at (-1.75, 0) {$\z_7$};
  \node[circle,black,thick,draw,scale=0.7] (Z8) at (-1, 1) {$\z_8$};
  \draw[-{Stealth[width=5pt,length=5pt]},very thick,WildStrawberry]  (X) edge[] node[yshift=5.5] {?} (Y);
  \draw[-{Stealth[width=5pt,length=5pt]},thick,black]  (Z1) edge["\footnotesize..."] (X);
  \draw[-{Stealth[width=5pt,length=5pt]},thick,black]  (Z1) edge["\footnotesize..."] (Y);
  \draw[-{Stealth[width=5pt,length=5pt]},thick,black]  (X) edge["\footnotesize.."] (Z3);
  \draw[-{Stealth[width=5pt,length=5pt]},thick,black]  (Z3) edge["\footnotesize.."] (Y);
  \draw[-{Stealth[width=5pt,length=5pt]},thick,black]  (Z4) edge["\footnotesize.."] node {} (Y);
  \draw[-{Stealth[width=5pt,length=5pt]},thick,black]  (Z5) edge["\footnotesize.."] (X);
  \draw[-{Stealth[width=5pt,length=5pt]},thick,black]  (X) edge["\footnotesize..."] (Z7);
  \draw[{Stealth[width=5pt,length=5pt]}-{Stealth[width=5pt,length=5pt]},thick,black] (Z2) edge[bend right=30,"\footnotesize..."] (Y) node[] {};
  \draw[-{Stealth[width=5pt,length=5pt]},thick,black] (X) edge[bend right=30,"\footnotesize..."] (Z2) node[] {};
  \node[black,below=of Z7,yshift=0.5cm] {\textbf{(A)}};
\end{tikzpicture}
\hspace{5mm}
\begin{tikzpicture}[every edge quotes/.style = {font=\footnotesize, fill=white,sloped}]
  \node[circle,WildStrawberry,very thick,draw,scale=0.7] (X) at (-1.25, 0) {$X$};
  \node[circle,WildStrawberry,very thick,draw,scale=0.7] (Y) at (0, 0) {$Y$};
  \node[circle,white,thick,draw,scale=0.7] (Z2) at (0, -1.75) {$\z_2$};
  \node[circle,black,thick,draw,scale=0.7] (Z1) at (0, 1.25) {$\z_1$};
  \node[circle,black,thick,draw,scale=0.7] (Z3) at (0, -1.25) {$\z_3$};
  \node[circle,black,thick,draw,scale=0.7] (Z4) at (1.25, 0) {$\z_4$};
  \draw[-{Stealth[width=5pt,length=5pt]},very thick,WildStrawberry]  (X) edge[] node[yshift=5.5] {?} (Y);
  \draw[-{Stealth[width=5pt,length=5pt]},thick,black]  (Z1) edge[] (Y);
  \draw[-{Stealth[width=5pt,length=5pt]},thick,black]  (Z3) edge[] (Y);
  \draw[-{Stealth[width=5pt,length=5pt]},thick,black]  (Z4) edge[] (Y);
  \node[black,below=of X,yshift=0.5cm] {\textbf{(B)}};
\end{tikzpicture}
    \caption{LD3 assesses whether the edge $X \to Y$ exists. \textbf{(A)} Allowable partitions under \ref{assumption:y_no_desc} and \ref{assumption:all_parents_y}. \textbf{(B)} Parents of $Y$ returned by LD3. Nodes are partition sets or subsets. Partition interrelations and latent confounding are abstracted away. Bidirected edge $Y \leftrightarrow \z_2$ signifies $\z_{2 \notin de(Y)}$. Edges with $\cdots$ are paths of arbitrary length. Solid edges are adjacencies.}
    \label{fig:ld3_graph}
\end{figure}

\begin{figure*}[!t]
    \centering
    \includegraphics[width=\linewidth]{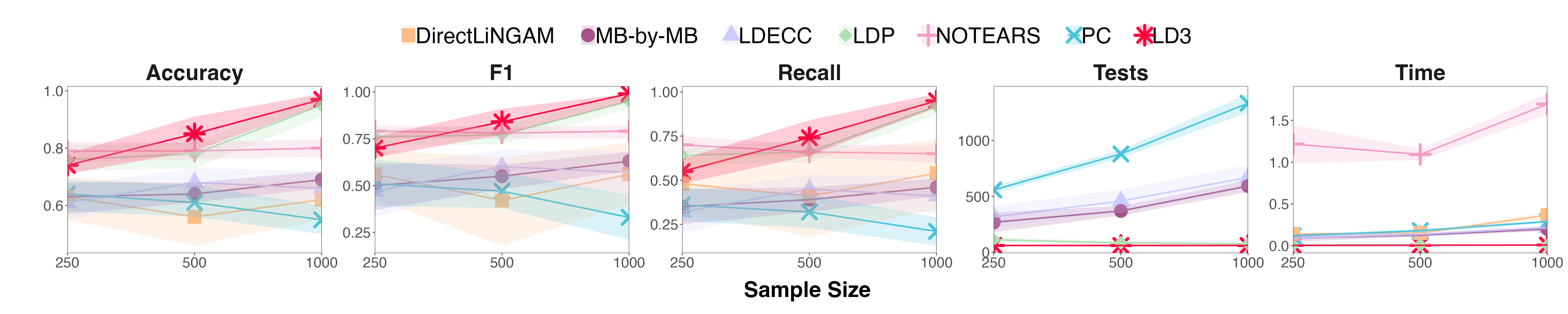}
    \caption{Baseline results for parent discovery in \textsc{Sangiovese}. Independence test count (Tests) is reported for constraint-based methods. Time is in seconds. Shaded regions denote 95\% confidence intervals over ten replicates.}
    \label{fig:sangiovese}
\end{figure*}

\paragraph{Causal Partitions in this Setting} Given \ref{assumption:y_no_desc}, there are no descendants of $Y$ in $\g$ and therefore no $\z_6$ nor $\z_{2 \in de(Y)}$. However, there may be $\z_{2 \notin de(Y)}$. Thus, $\g$ can contain the following seven (potentially empty) causal partitions with respect to $\{X,Y\}$: $\z_1$, $\z_{2 \notin de(Y)}$, $\z_3$, $\z_4$, $\z_5$, $\z_7$, and $\z_8$ (Figure \ref{fig:ld3_graph}.A). LD3 returns the partition \textit{subsets} in Figure \ref{fig:ld3_graph}.B: all $\z_1$, $\z_3$, and $\z_4$ that are directly adjacent to $Y$.

\begin{remark}[The observed WCDE under violations of \ref{assumption:all_parents_y}]
\label{remark:observed_wcde}
    While \ref{assumption:all_parents_y} is sufficient but not necessary for WCDE identifiability, each backdoor and frontdoor path must be blocked by at least one observable variable. If no variables on such a path are measured, 
    then no algorithm can identify the true WCDE. Users should note that the identifiability of any direct effect measure is fundamentally limited by the observability of backdoor and frontdoor paths. 
\end{remark}

\subsection{A Graphical Criterion for the Weighted CDE}

\revision{Under conditions where the WCDE is identifiable by Equation \ref{eq:wcde} (Remark \ref{remark:assumptions_confounding}), we propose the following criterion.}

\begin{definition}[Graphical criterion for identifying the WCDE] \label{def:adj_cde}
    Under the causal partition taxonomy defined in \citet{maasch2024local}, we define the set $\cde$ that contains all parents of the outcome:
    \begin{align}
        \cde \coloneqq \z_{1 \in pa(Y)} \cup \z_{3 \in pa(Y)} \cup \z_{4 \in pa(Y)}.
    \end{align}
\end{definition} 

\begin{theorem}[$\cde$ is a VAS for the WCDE] \label{theorem:a_cde_valid} 
$\cde$ is a valid adjustment set for the WCDE (Equation \ref{eq:wcde}), satisfying the identification conditions in Definition \ref{def:id_cde}. 
\end{theorem}


\noindent\textit{Intuition.} \; $\cde$ contains 
exactly all the parents of $Y$: confounders of $\{X,Y\}$ adjacent to $Y$ ($\z_{1 \in pa(Y)}$), mediators of $\{X,Y\}$ adjacent to $Y$ ($\z_{3 \in pa(Y)}$), and all parents of $Y$ that are marginally independent of $X$ ($\z_{4\in pa(Y)}$). Thus, at least one member of every backdoor and frontdoor path is in $\cde$. This provides conditional $d$-separation of $X$ and $Y$ if and only if there is no edge $X \to Y$. Proof is in Appendix \ref{appendix:proofs}.

\begin{remark}[The role of $\z_4$]
\label{remark:z4}
    Note that $\z_3$ and $Y$ can be confounded by $\z_1$, $\z_3$, or $\z_4$ (Figure \ref{fig:z3_confounded}). Including $\z_{4 \in pa(Y)}$ in $\cde$ helps guarantee the identifiability of WCDE without requiring exact knowledge of confounding for $\z_3$ and $Y$. 
\end{remark}

\revision{

\begin{remark}[Variance and Actionability]
\label{remark:variance_and_actionability}
     Definition \ref{def:adj_cde} defines a VAS for the WCDE in the general setting, irrespective of \ref{assumption:y_no_desc} and \ref{assumption:all_parents_y}. In settings where LD3 is used to obtain $\cde$, $\z$ itself constitutes a VAS (per \ref{assumption:y_no_desc} and \ref{assumption:all_parents_y}). However, adjusting for all $\z$ is not advised for two primary reasons:
    \begin{enumerate}
        \item Adjusting for all $\z$ risks \textit{unnecessary adjustment}, which can inflate estimator variance under finite data \citep{schisterman_overadjustment_2009}. To support statistical efficiency, we follow intuition provided by prior theorems on VAS optimality with respect to asymptotic variance, which dictate exclusion of $\z_5$, inclusion of $\z_4$, and generally favor control for parents of $Y$  \citep{rotnitzky2020efficient,henckel_graphical_2022}. 
        \item Ignoring causal structure limits the informativeness and actionability of fairness conclusions. Forgoing structure learning is a missed opportunity to identify potential structural mechanisms that could be redressed through interventions (e.g., policy change).
    \end{enumerate} 
    A full discussion of this topic is in Appendix \ref{sec:variance_actionability}. Sections \ref{sec:empirics} and \ref{sec:causal_fairness_analysis} provide empirical support for points (1) and (2).
\end{remark}
}

\section{Related Works} \label{sec:related_works}

\paragraph{Discovery for CFA}
Few works in causal discovery have centered on fairness objectives \citep{binkyte2023causal}. These include learning Suppes-Bayes causal networks with maximum likelihood estimation \citep{bonchi2017exposing} and applying PC Algorithm \citep{zhang2017causal}. To our knowledge, this work presents the first \textit{local} causal discovery method that is specifically tailored for CFA. 

\paragraph{Local Discovery of Direct Causes} Learning the direct causes of a target has primarily garnered interest in causal feature selection \citep{soleymani_causal_2022}. 
Various Markov blanket (MB) learners have been proposed, though many cannot distinguish parents from children and/or spouses \citep{yu2020causality}. MB-by-MB \citep{wang_discovering_2014}, Causal Markov Blanket (CMB; \citealt{gao_local_2015}), and Local Discovery Using Eager Collider Checks (LDECC; \citealt{gupta_local_2023}) are constraint-based methods that differentiate parents and children when unambiguous over the Markov equivalence class (MEC).  
Like the global algorithm PC \citep{spirtes2001causation}, MB-by-MB, CMB, and LDECC  have exponential time complexity with respect to variable set size in the worst case. Local Discovery by Partitioning (LDP; \citealt{maasch2024local}) causally partitions variables around $\{X,Y\}$ using a quadratic number of CI tests with respect to variable set size; results can be post-processed to identify parents of $Y$ if \ref{assumption:y_no_desc} is imposed.



\section{Empirical Validation on Synthetic Data} \label{sec:empirics}

\paragraph{Baselines} Baselines represent a range of approaches that are available as open-source Python implementations. Local baselines are MB-by-MB \citep{wang_discovering_2014}, LDECC \citep{gupta_local_2023}, and LDP \citep{maasch2024local}. Global baselines are PC \citep{spirtes2001causation}, DirectLiNGAM \citep{shimizu2011directlingam}, and NOTEARS \citep{zheng2018dags}.
Extended baseline descriptions and post-processing procedures are given in Appendix \ref{sec:appendix_baselines}. Besides LDP, all baselines assume causal sufficiency. PC, MB-by-MB, and LDECC return results in terms of the MEC. DirectLiNGAM assumes an additive noise model. DirectLiNGAM and NOTEARS assume linearity. All experiments used an Apple MacBook (M2 Pro Chip; 12 CPU cores; 16G memory).

\begin{figure*}[!t]
    \centering
    \includegraphics[width=0.48\linewidth]
    {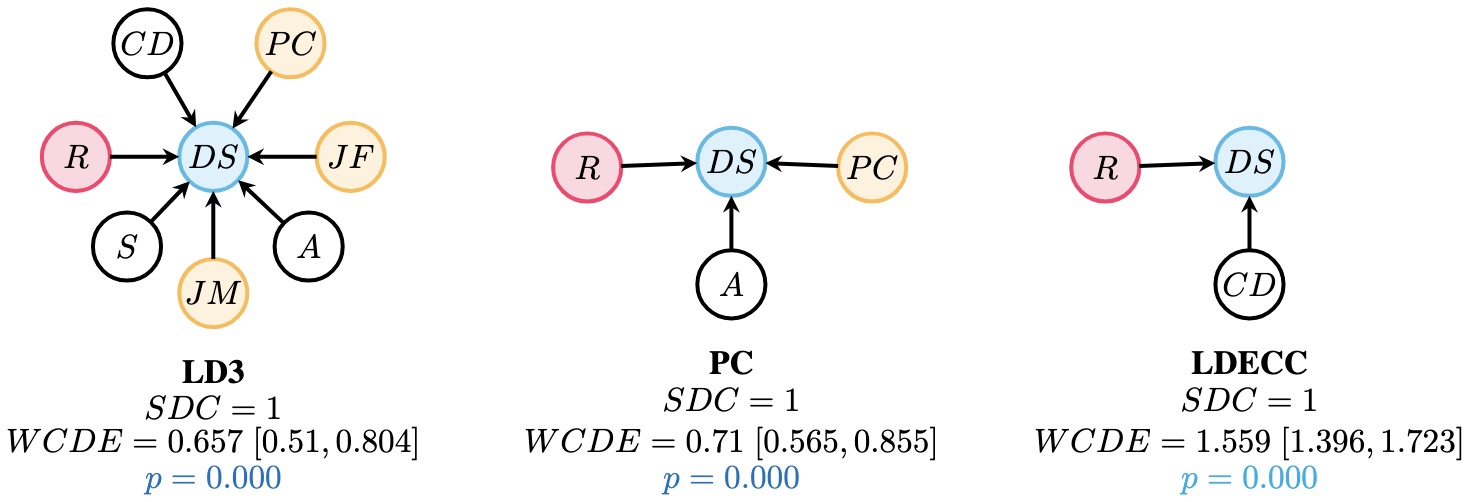}
    \hspace{5mm}
    \includegraphics[width=0.48\linewidth]{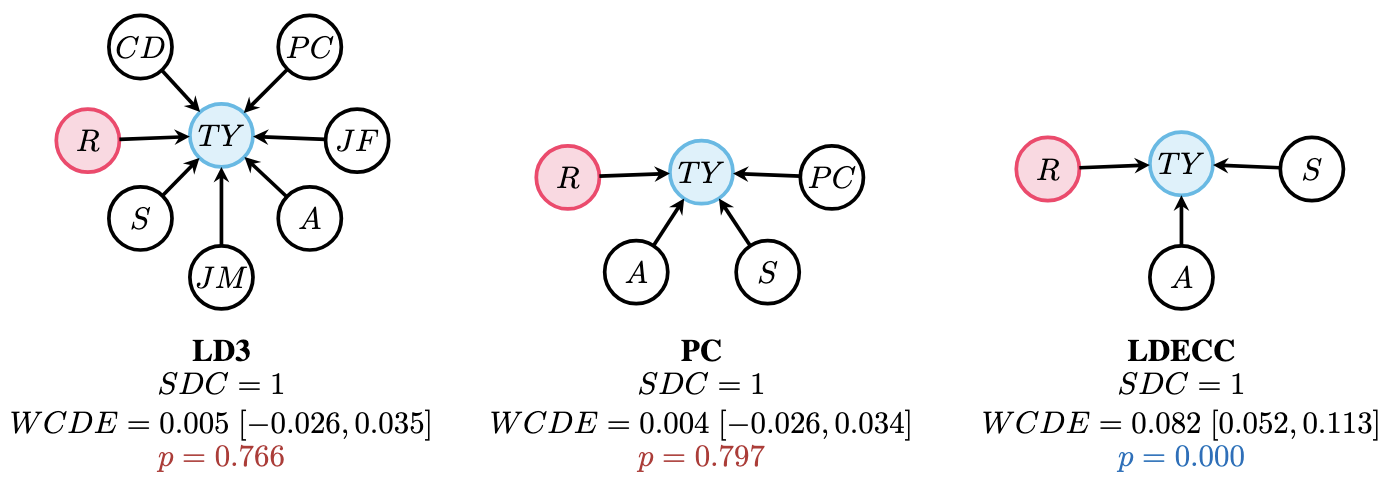}
    \caption{Predicted parent sets, SDC, and WCDE estimates for COMPAS. Exposure is race (\textit{R}; {\color{WildStrawberry}red}) and outcome is general recidivism risk decile score (\textit{DS}; {\color{CornflowerBlue}blue}). 
    Known parents of \textit{DS} are in {\color{Dandelion}yellow}. \textit{A} = age; \textit{CD} = charge degree; \textit{JF} = juvenile felonies; \textit{JM} = juvenile misdemeanors; \textit{PC} = priors count; \textit{S} = sex. WCDE is reported with $p$-value ($p$) and 95\% confidence intervals in brackets. All methods used $\chi^2$ CI tests ($\alpha = 0.05$). 
    Full results are in Figures \ref{fig:compas_ld3}–\ref{fig:compas_ldecc}.}
    \label{fig:compas_05}
\end{figure*}

\paragraph{Parent Discovery} 
We evaluated whether LD3 can recover true parent sets using an oracle independence test on random exposure-outcome pairs in 90 unique directed acyclic Erd\H{o}s-R\'enyi graphs (node counts in $[5...500]$) \citep{erdHos1960evolution}.\footnote{\texttt{https://r.igraph.org/}} 
Parent F1, recall, and precision were 100\%. Runtimes and total CI tests as node and edge cardinality scale are shown in Figure \ref{fig:time_tests}. In Appendix \ref{sec:weighted_cde_results_appendix}, we show for a linear-Gaussian SCM that WCDE estimates converged toward the true direct effect with low variance when adjusting  for $\cde$ discovered with LD3 (Figures \ref{fig:wcde}, \ref{fig:ld3_metrics}).

All baselines were assessed on the \textsc{Sangiovese} benchmark from the \texttt{bnlearn} repository \citep{scutari_learning_2010}, a linear-Gaussian model of Tuscan grape production \citep{magrini2017conditional}. Ten replicate datasets were sampled at $n = [250,500,1000]$. 
All constraint-based methods used Fisher-z tests ($\alpha = 0.01$). DirectLiNGAM assumes non-Gaussian noise and was expected to underperform. LD3 was generally most performant across metrics, with LDP performing similarly (Figure \ref{fig:sangiovese}, Table \ref{tab:grapes}). NOTEARS was significantly slower than other methods. Comparisons of LD3 to LDECC, LDP, and PC on \textsc{Asia} \citep{lauritzen_local_1988} and \textsc{Sachs} \citep{sachs2005causal} benchmarks are in Tables \ref{tab:asia} and \ref{tab:sachs}. Runtime comparisons are in Figure \ref{fig:time_tests} and Tables \ref{tab:asia_time_tests}, \ref{tab:sachs_time_tests}.


\rebuttal{\paragraph{Estimator Variance in Finite Samples}  As discussed in Remark \ref{remark:variance_and_actionability}, $\z$ itself is a VAS under \ref{assumption:y_no_desc} and \ref{assumption:all_parents_y}. However, adjusting for  $\z$ risks unnecessary adjustment, which can inflate the asymptotic variance of the causal effect estimator. We demonstrate the impacts of variance inflation as sample size scales in both a linear and nonlinear SCM (Figure \ref{fig:graph_variance}). Estimate variance using all $\z$ was $7.8\times$ to $9.6\times$ higher than using $\cde$ in the linear SCM (Table \ref{tab:variance_all_z}) and at least $12.6\times$ higher in the nonlinear SCM (Table \ref{tab:variance_all_z_learned}).

\paragraph{VAS Interpretability} Structure learning can improve the interpretability of the VAS by removing irrelevant variables. In some structures, this removal can substantially reduce VAS size. On two \texttt{bnlearn} benchmarks with moderate to large DAGs and small $\cde$ cardinality, discovery with LD3 reduced adjustment set cardinality by at least 95.4\% relative to retaining all $\z$ (Table \ref{tab:interpretability}). 

}

\section{Real-World Causal Fairness Analyses}
\label{sec:causal_fairness_analysis}

We deploy LD3 for two causal fairness settings: (1) racial discrimination in criminal recidivism prediction and (2) sex-based discrimination in healthcare. We compare the results of LD3 to PC and LDECC on the basis of $\cde$ quality and computational efficiency. Causal discovery used $\chi^2$ CI tests and WCDE estimation used double machine learning (\citealt{chernozhukov2018double}).\footnote{\texttt{https://econml.azurewebsites.net}} Estimators used random forest classifiers with a 70\% / 30\% train-test split.\footnote{\texttt{https://scikit-learn.org}} We assumed \ref{assumption:y_no_desc} and \ref{assumption:all_parents_y}. Data preprocessing is described in Appendix \ref{sec:fairness_experiments}. Note that all CFA results are only preliminary qualitative indicators, and further analyses should take place. 

\subsection{Race and COMPAS Recidivism Prediction}

\paragraph{Background} We assessed the ability of LD3 to facilitate CFA on the COMPAS dataset from ProPublica.\footnote{\texttt{github.com/propublica/compas-analysis}} Correctional Offender Management Profiling for Alternative Sanctions (COMPAS) is a commercial algorithm for case management and decision support used by the US criminal justice system to assess risk of recidivism \citep{compas_2015}. ProPublica's landmark exposé on COMPAS found that African Americans are ``almost twice as likely as whites to be labeled a higher risk but not actually re-offend'' \citep{angwin2022machine}.

We examined racial bias in the COMPAS General Recidivism Risk model. Due to data availability, we limited our analyses to the most represented racial groups (black and white). The algorithm's developer states that the model directly considers 
 prior criminal history and juvenile delinquency, among other factors
(\citealt{compas_2015}, p. 27). Our data contained multiple indicators of criminal history and juvenile delinquency, so these were used to assess the quality of causal discovery in lieu of complete ground truth. We used three significance levels for independence testing ($\alpha = 0.005, 0.01, 0.05$) to assess stability of results (Figures \ref{fig:compas_05}, \ref{fig:compas_ld3}–\ref{fig:compas_ldecc}). We selected 11 features with $n = 6150$ observations (2454 white, 3696 black; see Appendix \ref{sec:fairness_experiments}). We explored two outcomes: (1) \textit{general recidivism decile score} to probe bias in the COMPAS algorithm, and (2) \textit{actual two-year recidivism}, to examine factors in real outcomes.

\begin{figure*}[!t]
    \centering
    \includegraphics[width=0.48\linewidth]{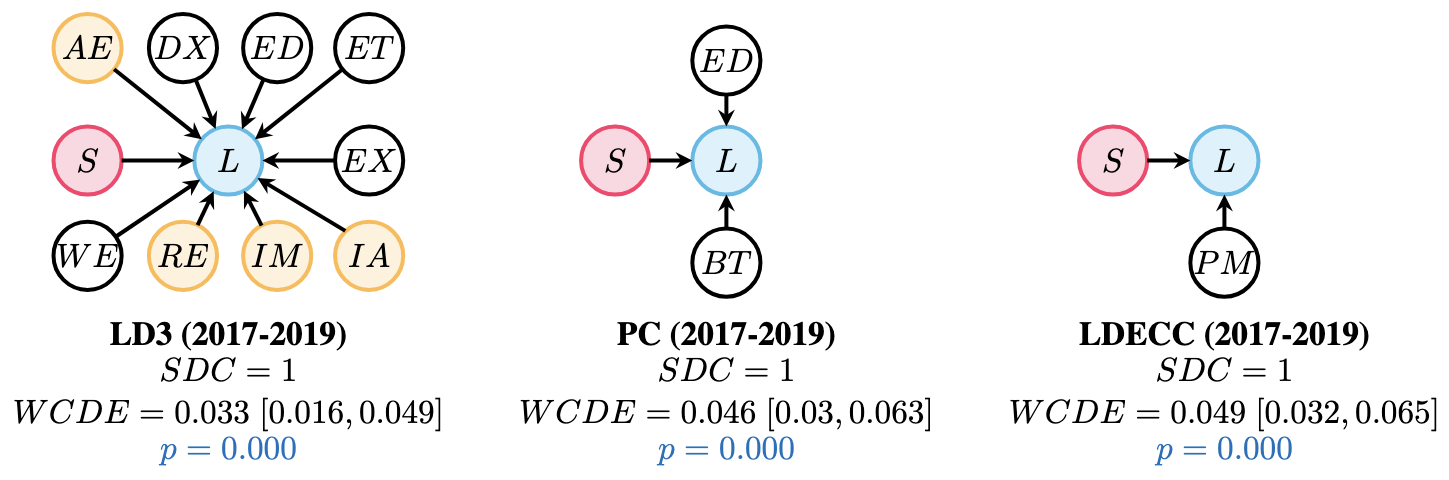}
    \hspace{2mm}
    \includegraphics[width=0.48\linewidth]{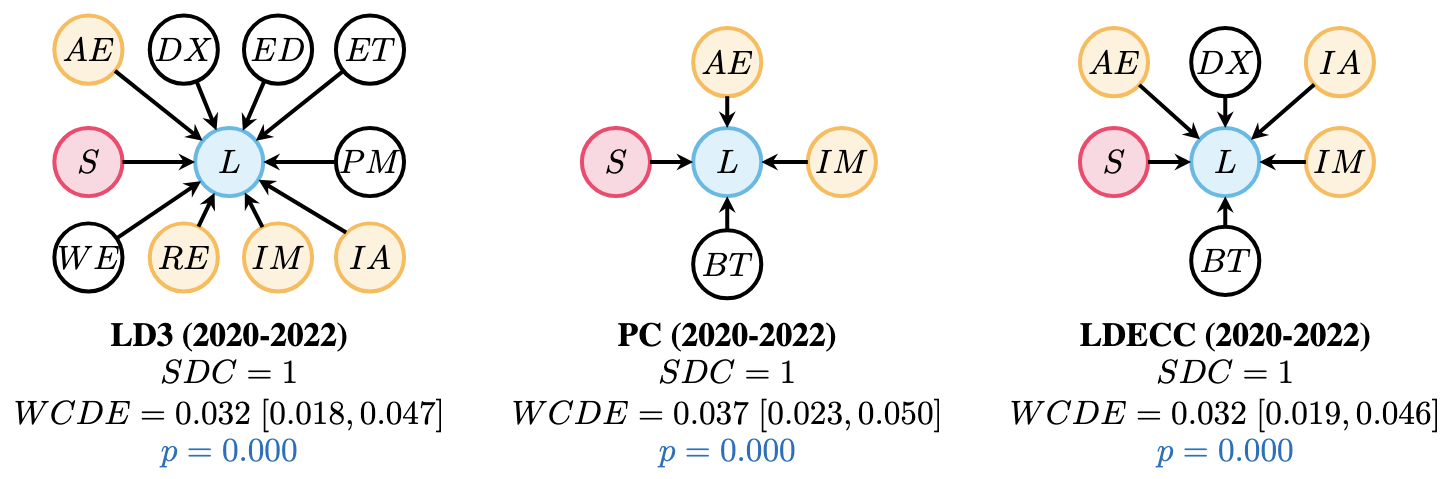}
    \caption{Predicted parent sets, SDC, and WCDE estimates for STAR liver data. Exposure is patient sex (\textit{S}; {\color{WildStrawberry}red}) and outcome is receiving a liver (\textit{L}; {\color{CornflowerBlue}blue}). Known parents of $L$ are in {\color{Dandelion}yellow}. \textit{AE} = active exception case; \textit{BT} = recipient blood type; \textit{DX} = diagnosis; \textit{ED} = education; \textit{ET} = ethnicity; \textit{EX} = exception type; \textit{IA} = initial age; \textit{IM} = initial MELD; \textit{PM} = payment method; \textit{RE} = region; \textit{WE} = weight. WCDE is reported with $p$-value ($p$) and 95\% confidence intervals in brackets. All methods used $\chi^2$ CI tests ($\alpha = 0.01$). Additional results are in Tables \ref{tab:liver}–\ref{tab:liver_ldecc}.
    }
    \label{fig:liver_in_text}
\end{figure*}

\paragraph{Results} At all significance levels, results from LD3, PC, and LDECC qualitatively agree that the effects of race on decile score are not fully explained by observed\footnote{Note that if \ref{assumption:all_parents_y} was violated, some of the observed effects might be due to unobserved variables. This is unverifiable using ProPublica's data alone. See Remark \ref{remark:observed_wcde}.} confounding and mediation (SDC = 1) and that the WCDE is signficant ($p = 0.000)$. LD3 successfully predicted that juvenile delinquency is a parent of decile score at all significance levels, while PC and LDECC never did. Age, priors count, and charge degree were nearly always predicted to be parents across methods and significance levels. For two-year recidivism, LD3 stably predicted that the WCDE of race was not significant ($p = 0.87, 0.68, 0.77$).

In general, results for PC and LDECC were less stable than LD3. Both methods wavered between strong significance ($p = 0.000$) and no significance ($p > 0.4$) for the WCDE of race on two-year recidivism. LDECC parent sets were inconsistent depending on whether they were taken as the intersection or union across graphs in the MEC, requiring additional interpretation by the user. On average, PC and LDECC took 46$\times$ longer to run and performed at least 11.7$\times$ more tests than LD3 across experiments (Figure \ref{fig:compas_time_tests}).


\subsection{Sex and Liver Transplant Allocation} \label{sec:liver}


\paragraph{Background} We apply LD3 to a real-world case study in the US healthcare system: fairness in liver transplant allocation. Liver transplantation is a critical therapeutic option for patients with end-stage chronic liver disease and acute liver failure. Demand significantly surpasses supply for donor livers \citep{srtr2020}, and patients are placed on a national waiting list managed by the United Network for Organ Sharing (UNOS). 
The distribution policy sorts waitlisted patients by multiple criteria, such as medical urgency, compatibility, and location \citep{latt2022liver}. Several key policy changes have sought to optimize distribution and improve patient outcomes (\citealt{papalexopoulos2023reshaping}; see Appendix \ref{appendix:liver}), including the \emph{model for end-stage liver disease} (MELD; \citealt{malinchoc2000model}). Despite efforts to increase fairness \citep{kim2022past}, it is widely recognized that US organ allocation suffers from disparities 
\citep{zhang2018racial}.

We explored potential sex-based discrimination in liver allocation. Sex-based disparities have been observed as statistical associations \citep{mathur2011sex,gordon2012gender,sarkar2015outcomes,oloruntoba2015gender,allen2018reduced,nephew2021racial}, 
but have not been explored through a causal lens. We use the National Standard Transplant Analysis and Research (STAR) dataset \citep{optn2024} for adult patients during 2017-2019 ($n = 21,101$) and 2020-2022 ($n= 22,807$). See Appendix \ref{appendix:liver} for feature selection and summary statistics.




\paragraph{Results} 
Under all experimental settings and time frames, results from LD3 suggest that the effects of sex on receiving a liver were not fully explained by observed confounding and mediation (Figure \ref{fig:liver_in_text}, Table \ref{tab:liver}). Causal evidence of direct discrimination was detected (SDC = 1) and corroborated by non-zero WCDE with significant $p$-values ($< 0.005$). In all settings, $\cde$ contains key factors used in the liver distribution policy, including initial MELD, initial age, region, and active exception case \citep{optn2024}. 
Our results are concordant with prior observations that differences in body size and medical condition contribute to sex-based disparities \citep{nephew2017exception}, as weight and diagnosis are in $\cde$ in all settings. Predicted $\cde$ also indicate potential discrimination with respect to ethnicity, education, and payment method. These socioeconomic factors may warrant further investigation for future policy interventions.

PC and LDECC qualitatively agree with LD3 on both discrimination metrics (Figure \ref{fig:liver_in_text}, Tables \ref{tab:liver_pc}, \ref{tab:liver_ldecc}). However, PC, LDECC, and LD3 had relatively low agreement in terms of $\cde$. Adjustment set cardinality was lower (and at times zero) for PC and LDECC. Most settings for PC and LDECC returned $\cde$ that omitted key expected variables, such as known policy criteria (e.g., initial MELD score, active exception case, and initial age). 
At both significance levels, PC returned multiple untrustworthy edges (e.g., weight $\to$ sex, height $\to$ sex, blood type $\to$ age, education $\to$ height, body mass index $\to$ height), undermining the credibility of results. Likewise, LDECC predicts that the outcome has children, which is known to be untrue. PC performed 479–1021$\times$ more tests and took 2295–5870$\times$ longer to execute than LD3 (Table \ref{tab:ld3_pc_time_tests}). LDECC performed 42–984$\times$ more tests and took 197–5774$\times$ longer to run.




\section{Conclusion}

This work advocates for increased practicality in causal fairness pipelines. We propose a time-efficient and asymptotically correct local discovery method for identifying two qualitative indicators of direct discrimination: the SDC and WCDE. 
For two real-world causal fairness analyses, LD3 returned more stable and plausible predictions with signficantly better computational efficiency relative to baselines.

\paragraph{Limitations and Future Directions} Future work could extend LD3 to allow for $Y$ with descendants. The assumption that all parents of $Y$ are observed is sufficient but not necessary for CDE identification and could be replaced with a different criterion. LD3 cannot differentiate $\z_1$ from $\z_3$, which would enable analysis of indirect and spurious discrimination. 
Future work could consider discovery for other fairness estimands, such as the natural effects \citep{pearl_interpretation_2014} and path-specific effects \citep{avin2005identifiability}. 



\section*{Acknowledgments}

This research was supported by NSF 2212175; NIH RF1AG084178, R01AG076448, R01AG080624, R01AG076234, R01AG080991 and RF1AG072449; and the NSF Graduate Research Fellowship (J. Maasch).


\bibliography{references}

\clearpage
\onecolumn

\appendix

\counterwithin{table}{section}
\counterwithin{figure}{section}
\counterwithin{definition}{section}
\counterwithin{theorem}{section}
\counterwithin{lemma}{section}

\clearpage

\section*{Ethical Statement}
Matters of unfairness and direct discrimination have real, profound impacts on human well-being and are often challenging to detect in real-world retrospective data. Researchers who deploy automated methods like LD3 should be prudent in interpreting and publicly communicating the statistical significance and uncertainty of their findings, whether evidence for or against discrimination is found.

\section*{Appendix}

\section{Further Discussion on Direct Effect Estimation} \label{appendix:direct_effects}

\subsection{Comparison of Direct Effect Measures}

Here we explore the differences and similarities between the CDE and alternative direct effect measures, and provide justification for our use of the CDE in this work. 


\paragraph{Alternative Direct Effect Measures} We elaborate on the comparison between the CDE and its most popular counterpart, the NDE \citep{pearl_direct_2001}.\footnote{The Ctf-DE is an extension of the NDE; see  \citet{zhang_fairness_2018}.} The CDE measures the expected change in outcome as the exposure changes when mediators $\mathbf{M}$ are uniformly fixed to $\mathbf{m}$, a \textit{constant value over the population} (as simulated by conditioning on $\mathbf{M}$; \citealt{pearl_causal_2012}). We express the CDE in $do$-notation as
\begin{align}
        \mathrm{CDE}(\mathbf{m}) &\coloneqq \mathbb{E}[Y_{x,\mathbf{m}} - Y_{x^*,\mathbf{m}}] \\
        &= \mathbb{E}[Y|do(X = x, \mathbf{M} = \mathbf{m})] - \mathbb{E}[Y|do(X = x^*, \mathbf{M} = \mathbf{m})]. 
    \end{align}
Assuming access to a covariate set $\mathbf{S}$ that simultaneously controls for confounding of the exposure-outcome and mediator-outcome relationships such that the conditions described in Definition \ref{def:id_cde} are satisfied,\footnote{When a single set $\mathbf{S}$ is not sufficient for valid control of confounding for both $\{X,Y\}$ and $\{\mathbf{M},Y\}$, we can apply a more complex formula that requires two covariate sets. See \citealt{vanderweele_controlled_2011} (Proposition 1) and \citealt{pearl_interpretation_2014} (Appendix C) for elaboration.} the CDE is identifiable from observational data as
\begin{align}
    \sum_{\mathbf{s}} \big( \mathbb{E}[Y|x, \mathbf{s}, \mathbf{m}] - \mathbb{E}[Y | x^*, \mathbf{s}, \mathbf{m}] \big) P(\mathbf{s}).
\end{align}
To obtain a unique estimate irrespective of potential interactions between $X$ and $\mathbf{M}$, we marginalize over the mediator values:
\begin{align}
    \mathrm{WCDE} &\coloneqq 
         \sum_{\mathbf{m}} \sum_{\mathbf{s}}  \big( \mathbb{E}[Y|x, \mathbf{s}, \mathbf{m}] - \mathbb{E}[Y | x^*, \mathbf{s}, \mathbf{m}] \big) P(\mathbf{s}) P(\mathbf{m}).
\end{align}

In contrast, the NDE measures the expected change in outcome as the exposure changes, with mediators fixed at the values that they would have been \textit{per individual} prior to this change in exposure. The NDE captures the portion of the total causal effect that would be transmitted to the outcome if the mediators did not respond to changes in the exposure.

\begin{definition}[Natural direct effect (NDE), \citealt{pearl_interpretation_2014}] \label{def:nde}
We express the average NDE as
    \begin{align}
    \mathrm{NDE}(x,x^*;Y) &= \mathbb{E}[Y_{x,\mathbf{M}_x^*} - Y_{x^*,\mathbf{M}_x^*}].
\end{align}
where $\mathbf{M}_x^*$ takes the value that the individual would have attained under $X = x^*$. The NDE requires probabilities of nested counterfactuals and a cross-world counterfactual assumption \citep{shpitser_complete_2011,pearl2012mediation, pearl_interpretation_2014,andrews2021insights}. Nevertheless, \citet{pearl_direct_2001} showed that the NDE can be expressed with \textit{do}-notation and reduced to a \textit{do}-free expression by valid covariate adjustment, allowing for identification and estimation \citep{pearl2012mediation}.  The average NDE in Markovian models is identifiable from observational data per Corollary 2 in \citet{pearl_direct_2001}, allowing counterfactual expressions to be replaced by probabilistic estimands:
\begin{align}
    \mathrm{NDE}(x,x^*;Y) &= \sum_\mathbf{m} \sum_\mathbf{s} \big( \mathbb{E}[Y|x,\mathbf{s},\mathbf{m}] - \mathbb{E}[Y|x^*,\mathbf{s},\mathbf{m}] \big) P(\mathbf{s}) P(\mathbf{m}|x^*,\mathbf{s}). \label{eq:nde_prob}
\end{align}
\end{definition}

Note that the WCDE and NDE  differ only by the way that we weigh the CDE:  for the WCDE, we weigh by $P(\mathbf{m})$, and for the NDE, we weigh by $P(\mathbf{m} | x^*,s)$. This stems from the fact that the CDE is a function of interventional distributions, while the NDE is a counterfactual quantity. Meanwhile, the Ctf-DE measures NDEs conditioned on the subpopulation $X = x$ \citep{zhang_fairness_2018}. 

\paragraph{Total Effect Decomposition} Unlike the NDE, the CDE does not admit a valid decomposition of the total causal effect into direct and indirect effects for the general nonparametric setting \citep{pearl_direct_2001}. Conversely, the total effect is always equal to the difference of the NDE and the reverse transition of the natural indirect effect (NIE), even in the presence of nonlinearities and exposure-mediator interactions \citep{pearl_interpretation_2014}. However, in the absence of exposure-mediator interaction, the CDE and NDE are equivalent and the CDE can be used for total effect decomposition \citep{vanderweele_controlled_2011}. Guaranteed total effect decomposability is a primary reason that the natural effects are employed in CFA \citep{plecko_causal_2024}. However, when qualitative indicators of direct discrimination are all that is needed, it suffices to estimate the CDE \citep{zhang_fairness_2018}.

\subsection{Graphical Criteria for Direct Effect Identification} 

Sufficient and necessary conditions for identification vary by estimand, but generally require covariate adjustment \citep{vanderweele_principles_2019}. Formal graphical rules can allow the VAS to be read directly from causal structures, even in the presence of latent confounding \citep{perkovic_complete_2015,maathuis_generalized_2015,runge_necessary_2021}. Graphical criteria for valid adjustment have been proposed for diverse causal quantities (e.g., Pearl's backdoor criterion for total effect estimation; \citealt{pearl_causal_1995}).  Often, multiple valid criteria exist per estimand. These vary in the amount of prior structural knowledge required, and criteria that require less knowledge are easier to implement in practice \citep{vanderweele_definition_2013,guo_confounder_2022}. 

Several graphical criteria have been proposed for the identification of direct effects and related quantities. \citet{shpitser_identication_2006} present a complete graphical criterion for conditional interventional distributions. \citet{pearl_direct_2001} defines a graphical criterion for identifying the NDE from observational data (Corollary 1). Discussion, debate, and confusion around the sufficient and necessary conditions for NDE identification have inspired a flurry of papers \citep{imai_identification_2010,imai_general_2010}, as summarized in \citet{pearl_interpretation_2014}. \citet{shpitser_complete_2011} present a complete graphical criterion for the NDE that reconciles some debate, showing where common criteria coincide.

Graphical criteria for direct effect estimation are often challenging for practitioners to implement \citep{shpitser_complete_2011}. To increase the practicality of covariate adjustment in complex and low-knowledge domains, we advocate for the development of graphical criteria that are (1) relatively intuitive and practical to implement in applied research, and (2) are satisfied by knowledge that can be reliably learned through data-driven methods. Thus, we apply the causal partition taxonomy of \citet{maasch2024local} to provide local graphical intuition for fast and practical adjustment set discovery.

\section{Proofs}
\label{appendix:proofs}

\subsection{Proof of Theorem \ref{theorem:algorithm_correctness} (Correctness of Algorithm \ref{alg:a_cde})} \label{sec:proofs_of_correctness}

\begin{proof} \label{proof:z2_excluded}
    We show that each set of tests is correct when all stated assumptions are met.
    \begin{itemize}\setlength\itemsep{1em}
        \item $\z_8$ (Line 3): As proven in \citep{maasch2024local}, the independence relations tested in Line 3 uniquely define partition $\z_8$ and are therefore an asymptotically correct test for identifying this partition, and this partition only.
        \item $\z_{5,7}$ (Line 4): As proven in \citep{maasch2024local}, the independence relations tested in Line 4 define partition $\z_7$ in all structures and $\z_5$ in structures for which $|\z_1| = 0$ (i.e., no confounders, latent nor observed, exist for $X$ and $Y$). Therefore, this test is asymptotically guaranteed to return all $\z_7$ for arbitrary structures, and all $\z_5$ if $\z_1$ is the empty set.
        \item $\z_4$ (Line 5): As proven in \citep{maasch2024local}, the independence relations tested in Line 5 uniquely define partition $\z_4$ and are therefore an asymptotically correct test for identifying this partition, and this partition only.
        \item $\z_{1\in pa(Y)} \cup \z_{3\in pa(Y)}$ (Line 8): At this stage, the only partitions remaining under consideration are $\z_1$, $\z_{2 \notin de(Y)}$, $\z_3$, and $\z_5$ (assuming that $\z_1$ is not the empty set). By definition, no $\z_5$ will ever be adjacent to $Y$, as this partition cannot share a causal path to $Y$ that is not mediated by $X$ \citep{hernan_instruments_2006, lousdal_introduction_2018, maasch2024local}. All $\z_{2 \notin de(Y)}$ are descended from $\z_1$, $\z_3$, and/or $\z_4$ (i.e., any partition that can enter into $Y$), but not from $Y$ itself. Thus, all $\z_2$ and $\z_5$ will be eliminated by eliminating variables not adjacent to $Y$ given $X \cup \z_{4} \cup \{\z' \setminus Z\}$. This test will also result in the elimination of $\z_1$ and $\z_3$ that are not adjacent to $Y$, including proxy confounders and proxy mediators as defined in \citet{maasch2024local}.
        \item $\z_{4\in pa(Y)}$ (Line 10): Any intervening node on the causal path from $\z_4$ to $Y$ ($\z_4 \to \cdots \to Y$) must also belong to a partition that can enter into $Y$ ($\cdots \to Y$). By definition, this is strictly $\z_1$, $\z_3$, and $\z_4$. Therefore, to obtain $d$-separation between $Y$ and any member of $\z_4$ that is \textit{not} adjacent to $Y$, a correct conditioning set will contain $X$ and members of $\z_1$, $\z_3$ and $\z_4$. Conditioning on $X$ is essential, as the remaining conditioning set can open noncausal paths between $\z_4$ and $Y$. It is therefore sufficient to condition on all $\z_1$ and $\z_3$ that are already known to be adjacent to $Y$, unioned with $X$ and all known $\z_4$. Any variable that is marginally dependent on $Y$ given this conditioning set is a member of $\z_4$ that is adjacent to $Y$.
        \item $\cde$ (Line 11): By definition of $\cde$.
        \item SDC (Line 12): By definition of the SDC (Definition \ref{def:sdc}), $SDC = \mathbf{1}(X \in pa(Y))$. If $X$ is not a parent of $Y$, it will always be $d$-separable from $Y$ when all their confounding and mediating paths are blocked. If $X$ is a parent of $Y$, $X$ and $Y$ will be always conditionally \textit{dependent} given any conditioning set. Therefore, the SDC will evaluate to 0 if and only if $X \ind Y| \z_{1\in pa(Y)} \cup \z_{3\in pa(Y)}$.
    \end{itemize}
\end{proof}

\subsection{Proof of Theorem \ref{theorem:latent}} 

We show that latent variables that are not parents of $Y$ do not impact the asymptotic guarantees of Algorithm \ref{alg:a_cde}. 

\begin{proof}
    \citet{maasch2024local} have shown previously that the marginal and conditional independence tests used to identify $\z_4$ and $\z_8$ are not affected by latent variables. Intuition for this follows from the fact that these tests only require knowledge of $\{X,Y\}$ and the candidate variable $Z$. In Algorithm \ref{alg:a_cde}, the test for $\z_{5,7}$ also relies only on knowledge of $X$, $Y$, and $Z$, and is therefore correct in the presence of latent variables. As we do not require identification of $\z_5$ itself for downstream results, this test suffices. Otherwise, latent variables could pose identifiability problems for $\z_5$, as in prior works \citep{maasch2024local}. \ref{assumption:all_parents_y} alone is sufficient to ensure $d$-separation between $Y$ and its non-parents at Lines 8 and 10, guaranteeing identification of the true parents of $Y$ in $\z_1$, $\z_3$, and $\z_4$. Therefore, latent variables that are not parents of $Y$ have no impact on Algorithm \ref{alg:a_cde}. 
\end{proof}

\subsection{Proof of Theorem \ref{theorem:a_cde_valid}}

\begin{proof} \label{proof:a_de_valid}
    This theorem claims that $\cde \coloneqq \z_{1 \in pa(Y)} \cup \z_{3 \in pa(Y)} \cup \z_{4 \in pa(Y)}$ is a VAS for WCDE identification under the conditions described in Section \ref{sec:cde_preliminaries}. Per the identifiability conditions of the WCDE (Definition \ref{def:id_cde}), a VAS must block all backdoor paths for $\{X,Y\}$ and $\{\mathbf{M},Y\}$. As shown in \citet{pearl_direct_2001}, knowledge of all mediators for $\{X,Y\}$ that are parents of $Y$ is sufficient for identifying the CDE (i.e., knowledge of every mediator in the full causal graph is unnecessary). Thus, we show that adjusting for $\cde$ (1) blocks all backdoor paths for $\{X,Y\}$, (2) blocks all backdoor paths for $\{\mathbf{M},Y\}$, (3) blocks all frontdoor paths for $\{X,Y\}$, and (4) does not open non-causal paths by conditioning on colliders. We use the language of \textit{causal partitions} defined in \citet{maasch2024local}. 

    
    \begin{enumerate}
        \item \textbf{\textit{Backdoor paths for $\{X,Y\}$ are blocked.}} Any confounder for $\{X,Y\}$ will be in causal partition $\z_1$, by definition of $\z_1$ (Table \ref{tab:partitions}; \citealt{maasch2024local}). All backdoor paths enter $Y$ by definition (i.e., $ \cdots \to Y$). As $\cde$ contains all confounders that are parents of $Y$, all backdoor paths for $\{X,Y\}$ will be blocked when conditioning on $\cde$.
        \item \textbf{\textit{Backdoor paths for $\{\mathbf{M},Y\}$ are blocked.}} Any confounder for $\mathbf{M}$ and $Y$ will be in $\z_1 \cup \z_3 \cup \z_4$ (e.g., Figure \ref{fig:z3_confounded}), as these are the only causal partitions with edges pointing into $Y$. Set $\cde$ by definition contains all members of $\z_1$, $\z_3$, and $\z_4$ that are parents of $Y$. 
        \item \textbf{\textit{Frontdoor paths for $\{X,Y\}$ are blocked.}} Any mediator for $\{X,Y\}$ will be in causal partition $\z_3$, by definition of $\z_3$ (Table \ref{tab:partitions}; \citealt{maasch2024local}). All frontdoor paths enter $Y$ by definition (i.e., $ \cdots \to Y$). As $\cde$ contains all mediators that are parents of $Y$, all frontdoor paths will be blocked. 
        \item \textbf{\textit{Non-causal paths through colliders are blocked.}} Though $\z$ can  contain colliders in $\z_{2 \notin de(Y)}$, these cannot be parents of $Y$ by definition. As defined, $\cde$ does not contain any members of $\z_2$. Thus, non-causal paths will not be opened by conditioning on $\cde$. 
    \end{enumerate}
\end{proof}

\begin{figure}[!h]
    \centering
\begin{tikzpicture}[every edge quotes/.style = {font=\scriptsize, fill=white,sloped}]
  \node[circle,black,thick,draw,scale=0.8] (X) at (0, 0) {$X$};
  \node[circle,black,thick,draw,scale=0.8] (Y) at (2, 0) {$Y$};
  \node[circle,black,thick,draw,scale=0.8] (Z1a) at (1, 1) {$Z_{1a}$};
  \node[circle,black,thick,draw,scale=0.8] (Z3a) at (1, -1) {$Z_{3a}$};
  \node[circle,black,thick,draw,scale=0.8] (Z1b) at (1, -2.25) {$Z_{1b}$};
  \draw[-{Stealth[width=5pt,length=5pt]},thick,auto,black]  (X) edge node {} (Y);
  \draw[-{Stealth[width=5pt,length=5pt]},thick,auto,black]  (Z1a) edge node {} (X);
  \draw[-{Stealth[width=5pt,length=5pt]},thick,auto,black]  (Z1a) edge node {} (Y);
  \draw[-{Stealth[width=5pt,length=5pt]},thick,auto,black]  (X) edge node {} (Z3a);
  \draw[-{Stealth[width=5pt,length=5pt]},thick,auto,black]  (Z3a) edge node {} (Y);
  \draw[-{Stealth[width=5pt,length=5pt]},ultra thick,auto,WildStrawberry]  (Z1b) edge node {} (Z3a);
  \draw[-{Stealth[width=5pt,length=5pt]},thick,black] (Z1b) edge[bend left=30] (X) node[] {};
  \draw[-{Stealth[width=5pt,length=5pt]},ultra thick,WildStrawberry] (Z1b) edge[bend right=30] (Y) node[] {};
\end{tikzpicture}
\hspace{10mm}
\begin{tikzpicture}[every edge quotes/.style = {font=\scriptsize, fill=white,sloped}]
  \node[circle,black,thick,draw,scale=0.8] (X) at (0, 0) {$X$};
  \node[circle,black,thick,draw,scale=0.8] (Y) at (2, 0) {$Y$};
  \node[circle,black,thick,draw,scale=0.8] (Z1a) at (1, 1) {$Z_{1a}$};
  \node[circle,black,thick,draw,scale=0.8] (Z3a) at (1, -1) {$Z_{3a}$};
  \node[circle,black,thick,draw,scale=0.8] (Z3b) at (1, -2.25) {$Z_{3b}$};
  \draw[-{Stealth[width=5pt,length=5pt]},thick,auto,black]  (X) edge node {} (Y);
  \draw[-{Stealth[width=5pt,length=5pt]},thick,auto,black]  (Z1a) edge node {} (X);
  \draw[-{Stealth[width=5pt,length=5pt]},thick,auto,black]  (Z1a) edge node {} (Y);
  \draw[-{Stealth[width=5pt,length=5pt]},thick,auto,black]  (X) edge node {} (Z3a);
  \draw[-{Stealth[width=5pt,length=5pt]},thick,auto,black]  (Z3a) edge node {} (Y);
  \draw[-{Stealth[width=5pt,length=5pt]},ultra thick,auto,WildStrawberry]  (Z3b) edge node {} (Z3a);
  \draw[-{Stealth[width=5pt,length=5pt]},thick,black] (X) edge[bend right=30] (Z3b) node[] {};
  \draw[-{Stealth[width=5pt,length=5pt]},ultra thick,WildStrawberry] (Z3b) edge[bend right=30] (Y) node[] {};
\end{tikzpicture}
\hspace{10mm}
\begin{tikzpicture}[every edge quotes/.style = {font=\scriptsize, fill=white,sloped}]
  \node[circle,black,thick,draw,scale=0.8] (X) at (0, 0) {$X$};
  \node[circle,black,thick,draw,scale=0.8] (Y) at (2, 0) {$Y$};
  \node[circle,black,thick,draw,scale=0.8] (Z1a) at (1, 1) {$Z_{1a}$};
  \node[circle,black,thick,draw,scale=0.8] (Z3a) at (1, -1) {$Z_{3a}$};
  \node[circle,black,thick,draw,scale=0.8] (Z4) at (1, -2.25) {$Z_{4}$};
  \draw[-{Stealth[width=5pt,length=5pt]},thick,auto,black]  (X) edge node {} (Y);
  \draw[-{Stealth[width=5pt,length=5pt]},thick,auto,black]  (Z1a) edge node {} (X);
  \draw[-{Stealth[width=5pt,length=5pt]},thick,auto,black]  (Z1a) edge node {} (Y);
  \draw[-{Stealth[width=5pt,length=5pt]},thick,auto,black]  (X) edge node {} (Z3a);
  \draw[-{Stealth[width=5pt,length=5pt]},thick,auto,black]  (Z3a) edge node {} (Y);
  \draw[-{Stealth[width=5pt,length=5pt]},ultra thick,auto,WildStrawberry]  (Z4) edge node {} (Z3a);
  \draw[-{Stealth[width=5pt,length=5pt]},ultra thick,WildStrawberry] (Z4) edge[bend right=30] (Y) node[] {};
\end{tikzpicture}
    \caption{$Y$ and members of $\z_3$ can be confounded by members of $\z_1$ (e.g., left), $\z_3$ (e.g., center), or $\z_4$ (e.g., right), as these are the only causal partitions with edges pointing into $Y$. }
    \label{fig:z3_confounded}
\end{figure}


\clearpage

\rebuttal{\section{Beyond Bias: Variance and Actionability}
\label{sec:variance_actionability}

Definition \ref{def:adj_cde} defines a VAS for the WCDE (Equation \ref{eq:wcde}) in the general setting, irrespective of \ref{assumption:y_no_desc} and \ref{assumption:all_parents_y}. However, in settings that assume \ref{assumption:y_no_desc} and \ref{assumption:all_parents_y}, bypassing structure learning and adjusting for all  $\z$ does not induce bias in WCDE estimation. However, this is not advised for two reasons: (1) this risks \textit{unnecessary adjustment}, which can inflate the variance of causal effect estimators under finite data \citep{schisterman_overadjustment_2009} and (2) ignoring causal structure limits the informativeness and actionability of fairness conclusions. 

\paragraph{Impacts on Asymptotic Variance} While $\z$ itself constitutes a VAS under Assumptions \ref{assumption:y_no_desc} and \ref{assumption:all_parents_y}, adjusting for all $\z$ with no knowledge of its causal structure can undermine statistical efficiency. We highlight some prior theorems here, using the language of causal partitions. 

Consider the optimality of a VAS with respect to the \textit{asymptotic variance} (AV) of the causal effect estimator. For total effect estimation in both linear and nonparametric settings, it has been proven that the asymptotically optimal VAS captures less or equal information about the exposure and more or equal information about the outcome than alternative VAS \citep{rotnitzky2020efficient, witte2020efficient, henckel_graphical_2022}. This implies:\footnote{See Theorem 1 and its corollaries in \citet{henckel_graphical_2022}, which were later proven for the nonparametric case in \citet{rotnitzky2020efficient}.}
\begin{enumerate}
    \item VAS containing members of $\z_5$ will have \textit{greater} or equal AV than VAS not containing any $\z_5$;
    \item VAS containing members of $\z_4$ will have \textit{lesser} or equal AV than VAS not containing any $\z_4$; and 
    \item VAS containing $\z_{1 \not \in pa(Y)}$ or $\z_{4 \not \in pa(Y)}$ will have \textit{greater} or equal AV than VAS where all $\z_1$ and $\z_4$ are in $pa(Y)$.
\end{enumerate}
While these results apply to total effect estimation, we follow this intuition for WCDE estimation using $\cde$, which excludes $\z_5$ and $\z_{1,4 \not \in pa(Y)}$. We defer formal derivation of the asymptotically optimal VAS for the WCDE to future work. In Section \ref{sec:empirics} and Appendix \ref{sec:adjusting_for_all}, we empirically demonstrate variance inflation in finite samples when adjusting for all $\z$ versus $\cde$.

\paragraph{Actionable Fairness Conclusions} Ultimately, CFA must support our ability to \textit{take actions} that remove unwanted association between the protected attribute and outcome. In CFA, we examine causal structure to (1) determine whether unfairness is present and 
(2) identify structural mechanisms that can be redressed through interventions (e.g., policy change). LD3 addresses goal (1) by using the SDC and WCDE for binary classification of the SCM (fair/unfair) with respect to direct discrimination. Through causal partition labeling, LD3 yields interpretable VAS that provide partial information for goal (2). Adjusting for all $\z$ in WCDE estimation with no prior structure learning can address (1), but not (2). Consider the following example SCMs with unknown graphical structure, where \ref{assumption:y_no_desc} and \ref{assumption:all_parents_y} hold.
\begin{enumerate}
    \item \textbf{(A) WCDE $\neq$ 0, (B) $\z_{1,3}$ is empty.} We can assess (A) by adjusting for $\z$ without structure learning, but we cannot know (B). Thus, we can conclude that the SCM is unfair with respect to direct discrimination, but not that indirect and spurious unfairness are absent. Discovery methods like LD3 address (A) and (B) by enabling precise WCDE estimation and indicating that policy interventions need only address direct mechanisms of unfairness.
    \item \textbf{(A) WCDE = 0, (B) $\z_{1,3}$ is not empty.} All pathways of unfairness in the SCM pass through $\z_{1,3 \in pa(Y)}$. We can ask the policy design question: are any variables in $\z_{1,3 \in pa(Y)}$ feasible points of intervention for mitigating unfairness? While (A) implies (B) even without structure learning, we cannot answer our policy design question without the structural knowledge returned by methods like LD3.
\end{enumerate}
Empirical demonstration of improved VAS interpretability is given in Section \ref{sec:empirics}. Examples of structural insights provided by local discovery are given in the real-world CFA described in Section \ref{sec:causal_fairness_analysis}. 

}


\clearpage

\section{Empirical Validation on Synthetic and Semi-Synthetic Data}
\label{sec:appendix_empirics}

\vspace{2mm}

\paragraph{Computing Resources} All experiments used an Apple MacBook (M2 Pro Chip; 12 CPU cores; 16G memory).

\subsection{LD3 in the Presence of Latent Variables}
\label{sec:robustness}

We explored the impacts of latent variables that do and do not violate Assumption \ref{assumption:all_parents_y} (Table \ref{tab:latent_confounding}). We iteratively dropped individual variables from the observed data represented by Figure \ref{fig:eval_graph_cde}, including variables that do and do not act as confounders.

Results illustrate that (1) Theorem \ref{theorem:latent} holds and (2) Assumption \ref{assumption:all_parents_y} is sufficient but not necessary. WCDE for latent confounders $B_1$, $M_1$, and $Z_{4a}$ remain unbiased, as these are not adjacent to $Y$ (Theorem \ref{theorem:latent}). 
Though latent $M_2$ could induce bias by allowing for the erroneous inclusion of $M_3$ in $\cde$, WCDE results are not significantly impacted in this setting. Additionally, though $Z_{3c}$ is a parent of $Y$, dropping this variable simply causes LD3 to place its parent $Z_{3b}$ in $\cde$. In this setting, the predicted $\cde$ has 100\% F1, precision, and recall with respect to the expected VAS, and the WCDE remains unbiased ($1.25 [1.24,1.26]$). Thus, Assumption \ref{assumption:all_parents_y} is sufficient but \textit{not necessary} for returning a VAS.

\vspace{5mm}

\begin{table}[!h]
    \centering
    \begin{tabular}{l c c c c c}
    \toprule
       \textsc{Latent} & \textsc{WCDE} & \textsc{SDC Acc} & $\cde$ \textsc{F1} & $\cde$ \textsc{Prec} & $\cde$ \textsc{Rec} \\
       \midrule
        {\color{Maroon}$Z_1 \in \z_1$} & 1.70 [1.66,1.75] & 1.0  & 0.80 [0.80,0.80] & 0.67 [0.67,0.67] & 1.00 [1.00,1.00] \\
        $B_1 \in \z_1$ & 1.25 [1.24,1.26] & 1.0 & 1.00 [1.00,1.00] & 1.00 [1.00,1.00] & 1.00 [1.00,1.00] \\
        {\color{Maroon}$B_2 \in \z_1$} & 1.22 [1.17,1.27] & 1.0 & 0.91 [0.89,0.93] & 0.84 [0.81,0.86] & 1.00 [1.00,1.00] \\
        {\color{Maroon}$B_3 \in \z_1$} & 1.72 [1.68,1.76] & 1.0 & 0.82 [0.8,0.840] & 0.70 [0.67,0.73] & 1.00 [1.00,1.00] \\
        $M_1 \in \z_5$ & 1.25 [1.24,1.26] & 1.0 & 0.99 [0.98,1.01] & 0.99 [0.96,1.01] & 1.00 [1.00,1.00] \\
        {\color{Maroon}$M_2 \in \z_4$} & 1.26 [1.25,1.26] & 1.0 & 0.86 [0.86,0.86] & 0.75 [0.75,0.75] & 1.00 [1.00,1.00] \\
        $Z_{4a} \in \z_4$ & 1.25 [1.24,1.26] & 1.0 & 1.00 [1.00,1.00] & 1.00 [1.00,1.00] & 1.00 [1.00,1.00] \\
        \bottomrule
    \end{tabular}
    \caption{Results with latent confounders. WCDE estimates (\textsc{WCDE}), SDC accuracy (\textsc{Acc}), $\cde$ F1, $\cde$ precision (\textsc{Prec}), and $\cde$ recall (\textsc{Rec}) for LD3 as confounders in Figure \ref{fig:eval_graph_cde} are iteratively dropped from the observed data (linear-Gaussian data; Fisher-z tests; $\alpha = 0.01$; true direct effect = 1.25). Values are means over 10 replicates with 95\% confidence intervals in brackets. Latent nodes in red are adjacent to outcome $Y$, violating Assumption \ref{assumption:all_parents_y}. $\cde$ metrics were scored by comparing the set returned by LD3 to the true $\cde$ set-minus the latent variable.} 
    \label{tab:latent_confounding}
\end{table}

\vspace{10mm}
\begin{figure}[!h]
    \centering
    \begin{tikzpicture}[scale=0.15]
\tikzstyle{every node}+=[inner sep=0pt]
\draw [WildStrawberry,very thick] (19.3,-27.4) circle (3);
\draw (19.3,-27.4) node {$X$};
\draw [WildStrawberry,very thick] (50.6,-27.4) circle (3);
\draw (50.6,-27.4) node {$Y$};
\draw [black,very thick, dashed] (34.9,-9.7) circle (3);
\draw (34.9,-9.7) node {$B_3$};
\draw [black,very thick, dashed] (25.5,-6.9) circle (3);
\draw (25.5,-6.9) node {$B_1$};
\draw [black,very thick, dashed] (44.6,-6.9) circle (3);
\draw (44.6,-6.9) node {$B_2$};
\draw [black,very thick, dashed] (34.9,-19.6) circle (3);
\draw (34.9,-19.6) node {$Z_1$};
\draw [black,very thick] (24.3,-39.9) circle (3);
\draw (24.3,-39.9) node {$Z_{3a}$};
\draw [black,very thick] (34.2,-43.4) circle (3);
\draw (34.2,-43.4) node {$Z_{3b}$};
\draw [black,very thick,dashed] (44.6,-39.9) circle (3);
\draw (44.6,-39.9) node {$Z_{3c}$};
\draw [black,very thick] (34.9,-32.1) circle (3);
\draw (34.9,-32.1) node {$Z_{3d}$};
\draw [black,very thick, dashed] (62.3,-31.5) circle (3);
\draw (62.3,-31.5) node {$Z_{4a}$};
\draw [black,very thick] (11,-19.6) circle (3);
\draw (11,-19.6) node {$Z_5$};
\draw [black,very thick, dashed] (15.3,-46.3) circle (3);
\draw (15.3,-46.3) node {$M_1$};
\draw [black,very thick, dashed] (55.6,-46.3) circle (3);
\draw (55.6,-46.3) node {$M_2$};
\draw [black,very thick] (34.9,-50.3) circle (3);
\draw (34.9,-50.3) node {$M_3$};
\draw [black,very thick] (59.8,-19.6) circle (3);
\draw (59.8,-19.6) node {$Z_{4b}$};
\draw [WildStrawberry,very thick,dotted,-{Stealth[width=5pt]}] (22.3,-27.4) -- (47.6,-27.4);
\draw [black,semithick,-{Stealth[width=5pt]}] (22.17,-28.27) -- (32.03,-31.23);
\draw [black,semithick,-{Stealth[width=5pt]}] (37.9,-31.5) -- (47.75,-28.32);
\draw [black,semithick,-{Stealth[width=5pt]}] (59.3,-31.57) -- (37.9,-32.03);
\draw [black,semithick,-{Stealth[width=5pt]}] (57.51,-21.54) -- (52.89,-25.46);
\draw [black,semithick,-{Stealth[width=5pt]}] (13.19,-21.65) -- (17.11,-25.35);
\draw [black,semithick,-{Stealth[width=5pt]}] (32.22,-20.94) -- (21.98,-26.06);
\draw [black,semithick,-{Stealth[width=5pt]}] (37.59,-20.93) -- (47.91,-26.07);
\draw [black,semithick,-{Stealth[width=5pt]}] (24.63,-9.77) -- (20.17,-24.53);
\draw [black,semithick,-{Stealth[width=5pt]}] (45.44,-9.78) -- (49.76,-24.52);
\draw [black,semithick,-{Stealth[width=5pt]}] (41.72,-7.73) -- (37.78,-8.87);
\draw [black,semithick,-{Stealth[width=5pt]}] (28.38,-7.76) -- (32.02,-8.84);
\draw [black,semithick,-{Stealth[width=5pt]}] (36.89,-11.94) -- (48.61,-25.16);
\draw [black,semithick,-{Stealth[width=5pt]}] (32.92,-11.95) -- (21.28,-25.15);
\draw [black,semithick,-{Stealth[width=5pt]}] (27.13,-40.9) -- (31.37,-42.4);
\draw [black,semithick,-{Stealth[width=5pt]}] (37.04,-42.44) -- (41.76,-40.86);
\draw [black,semithick,-{Stealth[width=5pt]}] (45.9,-37.2) -- (49.3,-30.1);
\draw [black,semithick,-{Stealth[width=5pt]}] (20.41,-30.19) -- (23.19,-37.11);
\draw [black,semithick,-{Stealth[width=5pt]}] (18.24,-46.9) -- (31.96,-49.7);
\draw [black,semithick,-{Stealth[width=5pt]}] (52.65,-46.87) -- (37.85,-49.73);
\draw [black,semithick,-{Stealth[width=5pt]}] (62.4,-28.5) -- (60.64,-22.48);
\draw [black,semithick,-{Stealth[width=5pt]}] (15.92,-43.37) -- (18.68,-30.33);
\draw [black,semithick,-{Stealth[width=5pt]}] (54.83,-43.4) -- (51.37,-30.3);
\end{tikzpicture}
    \caption{Eighteen-node DAG used to evaluate the ability of LD3 to recover the true $\cde$, assess the SDC, and estimate the weighted CDE. For WCDE experiments (Figures \ref{fig:wcde}, \ref{fig:ld3_metrics}), all nodes were observed. For latent variable experiments (Section \ref{sec:robustness}), one dashed node was treated as latent per iteration.}
    \label{fig:eval_graph_cde}
\end{figure}
\vspace{10mm}

    


\subsection{Baseline Evaluation} \label{sec:appendix_baselines}

\paragraph{Baseline Evaluation for Parent Discovery} Baselines were evaluated on the basis of F1, precision, and recall with respect to the true parents of outcome $Y$ (i.e., the members of $\cde$). Only LD3, LDECC, and MB-by-MB directly return the parents of $Y$. Additionally, PC, LDECC, and MB-by-MB return results relative to the MEC. Therefore, we evaluate each baseline with at least one form of post-processing to adjust for the task of $\cde$ prediction. Post-processing is described below.

\begin{enumerate}
    \item \textbf{NOTEARS:} Retain all parents of the $Y$ in the unique graph.
    \item \textbf{DirectLiNGAM:} Retain all parents of the $Y$ in the unique graph.
    \item \textbf{PC$_{\cap}$:} Retain the \textit{intersection} of parent sets across DAGs in the MEC.
    \item \textbf{PC$_{\cup}$:} Retain the \textit{union} of parent sets across DAGs in the MEC.
    \item \textbf{MB-by-MB:} Retain the variables labeled as parents of the target.
    \item \textbf{LDECC$_{\cap}$:} Retain the variables labeled as parents of the target.
    \item \textbf{LDECC$_{\cup}$:} Retain the variables labeled as parents of the target unioned with the variables adjacent to $Y$ that are unoriented in the MEC.
    \item \textbf{LDP:} Retain $\z_1 \cup \z_{\textsc{POST}} \cup \z_4$. As our experimental setting requires that there are no descendants of the outcome, the superset $\z_{\textsc{POST}} \coloneqq \z_{2 \in de(Y)} \cup \z_3 \cup \z_6$ defined in \citet{maasch2024local} will now only contain $\z_3$. 
    \item \textbf{LDP$_{pa}$:} Retain the variables that are not conditionally independent of $Y$ given the remaining members of the set. 
\end{enumerate}

For the \textsc{Sangiovese} benchmark, we compared LD3 to NOTEARS, DirectLiNGAM, MB-by-MB, LDECC$_{\cap}$, PC$_{\cap}$, and LDP$_{pa}$. For \textsc{Asia} and \textsc{Sachs}, we compared LD3 to PC$_{\cap}$, PC$_{\cup}$, LDECC$_{\cap}$, LDECC$_{\cup}$, LDP with no post-processing, and LDP$_{pa}$ to emphasize the impacts of different post-processing methods.


\subsection{Erd\H{o}s-R\'enyi Graphs}

\vspace{10mm}

\begin{figure}[!h]
    \centering
    \includegraphics[width=0.35\textwidth]{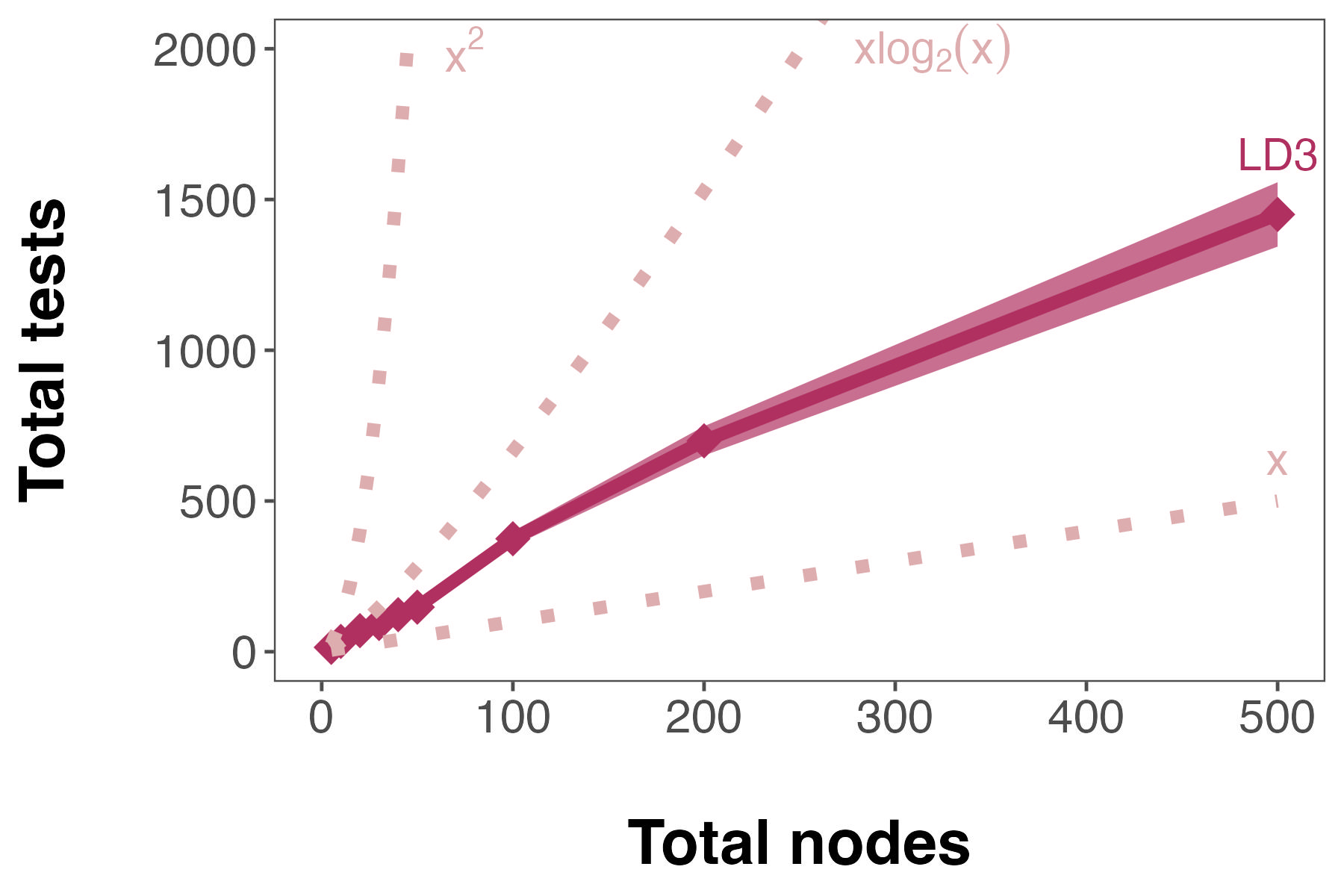} \hspace{5mm}
    \includegraphics[width=0.35\textwidth]{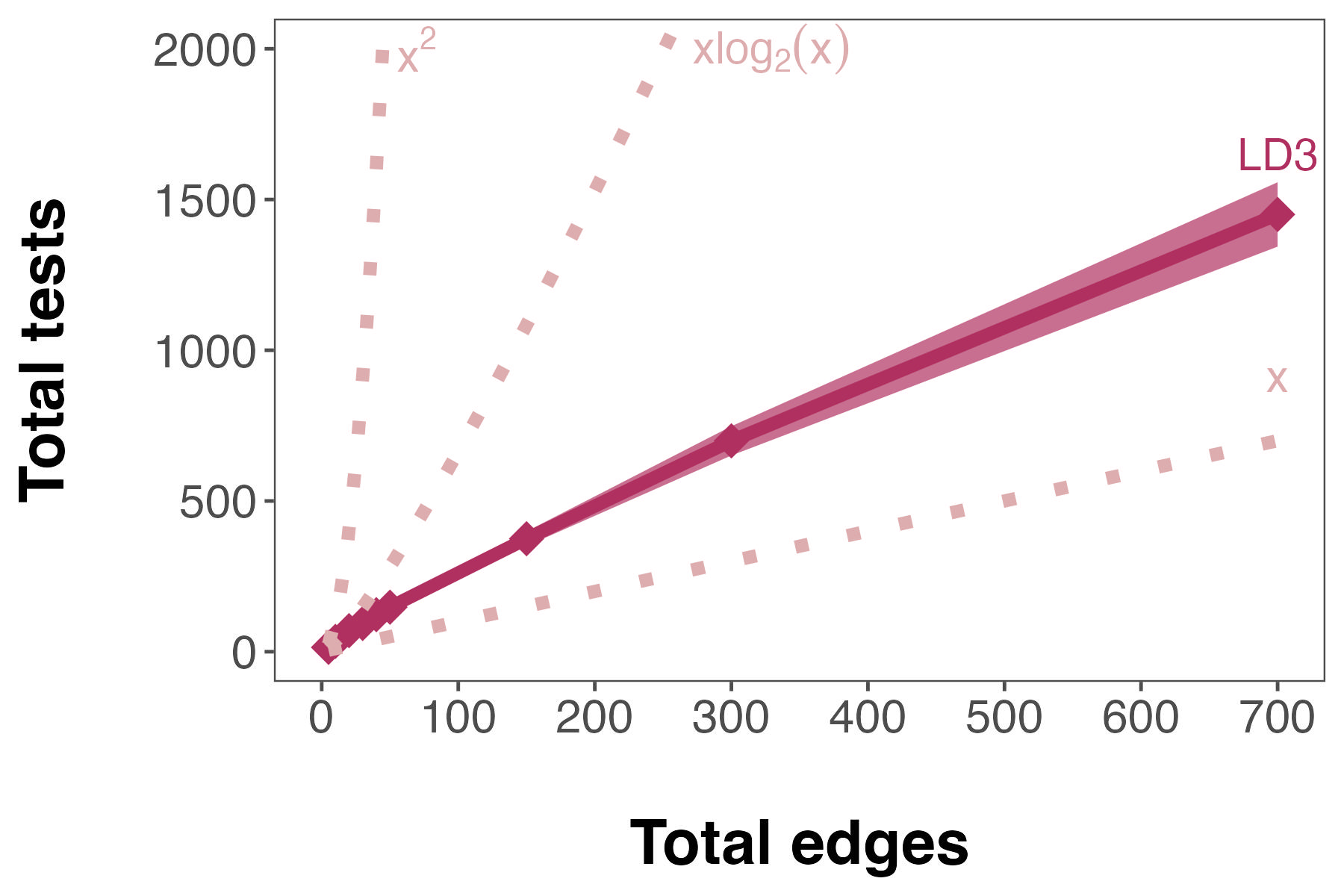} \\
    \vspace{5mm}
    \includegraphics[width=0.35\textwidth]{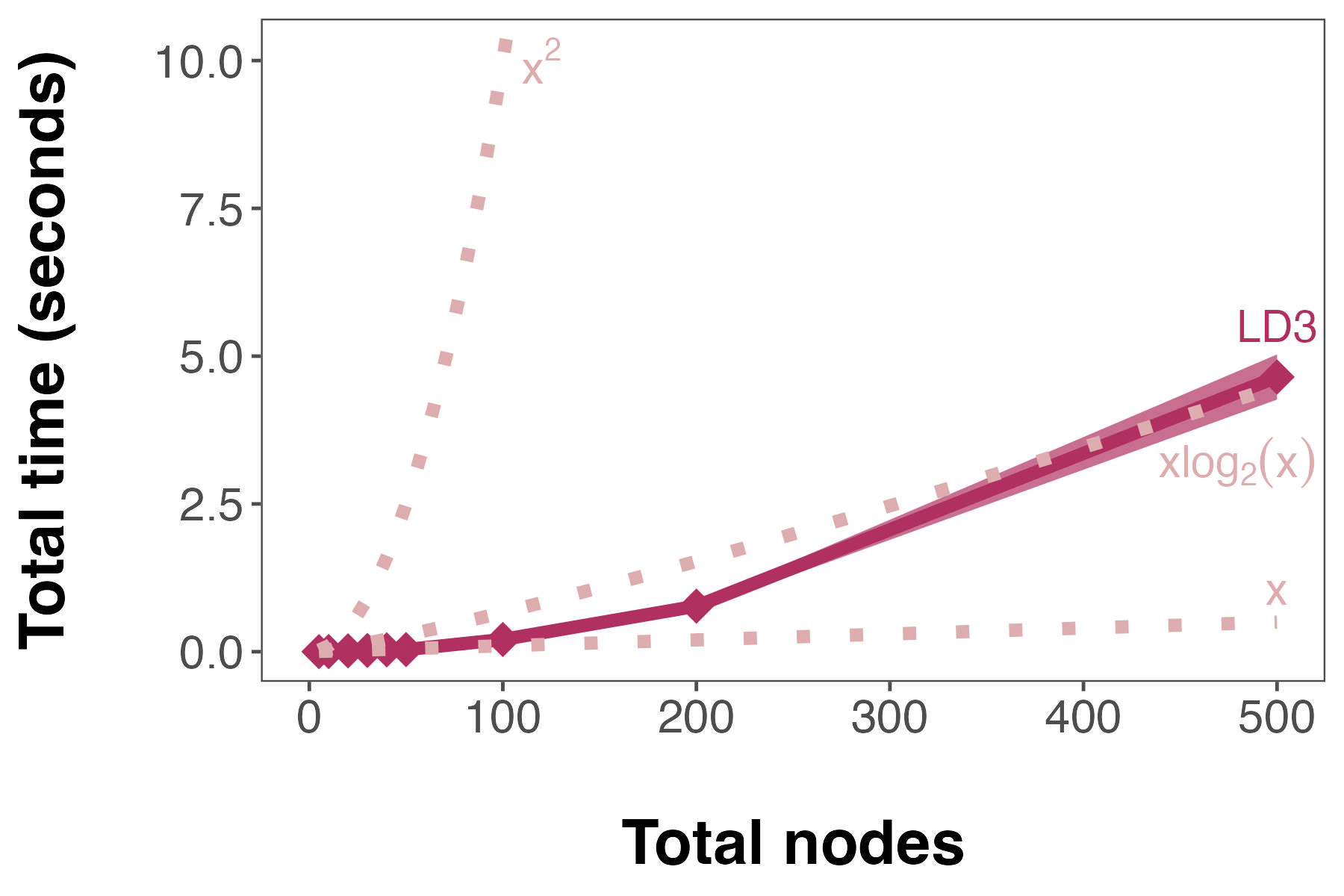}
    \hspace{5mm}
    \includegraphics[width=0.35\textwidth]{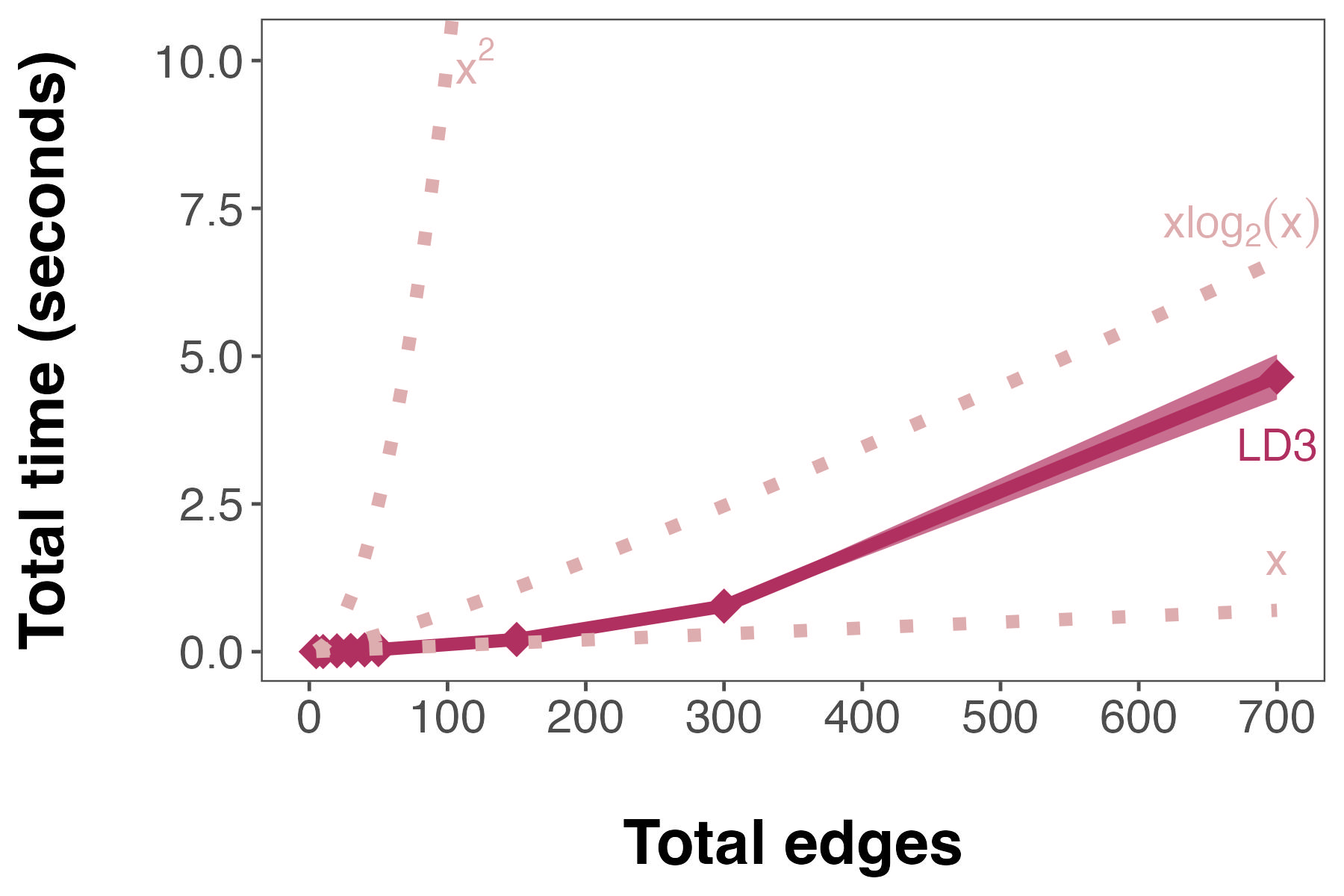}
    \caption{Total runtime and total CI tests performed with respect to total edges and total nodes, using an oracle on random directed acyclic Erd\H{o}s-R\'enyi graphs \citep{erdHos1960evolution}.}
    \label{fig:time_tests}
\end{figure}


\clearpage

\subsection{Weighted CDE Estimation}
\label{sec:weighted_cde_results_appendix}

\vspace{5mm}

\begin{figure}[!h]
    \centering
    \includegraphics[width=0.45\textwidth]{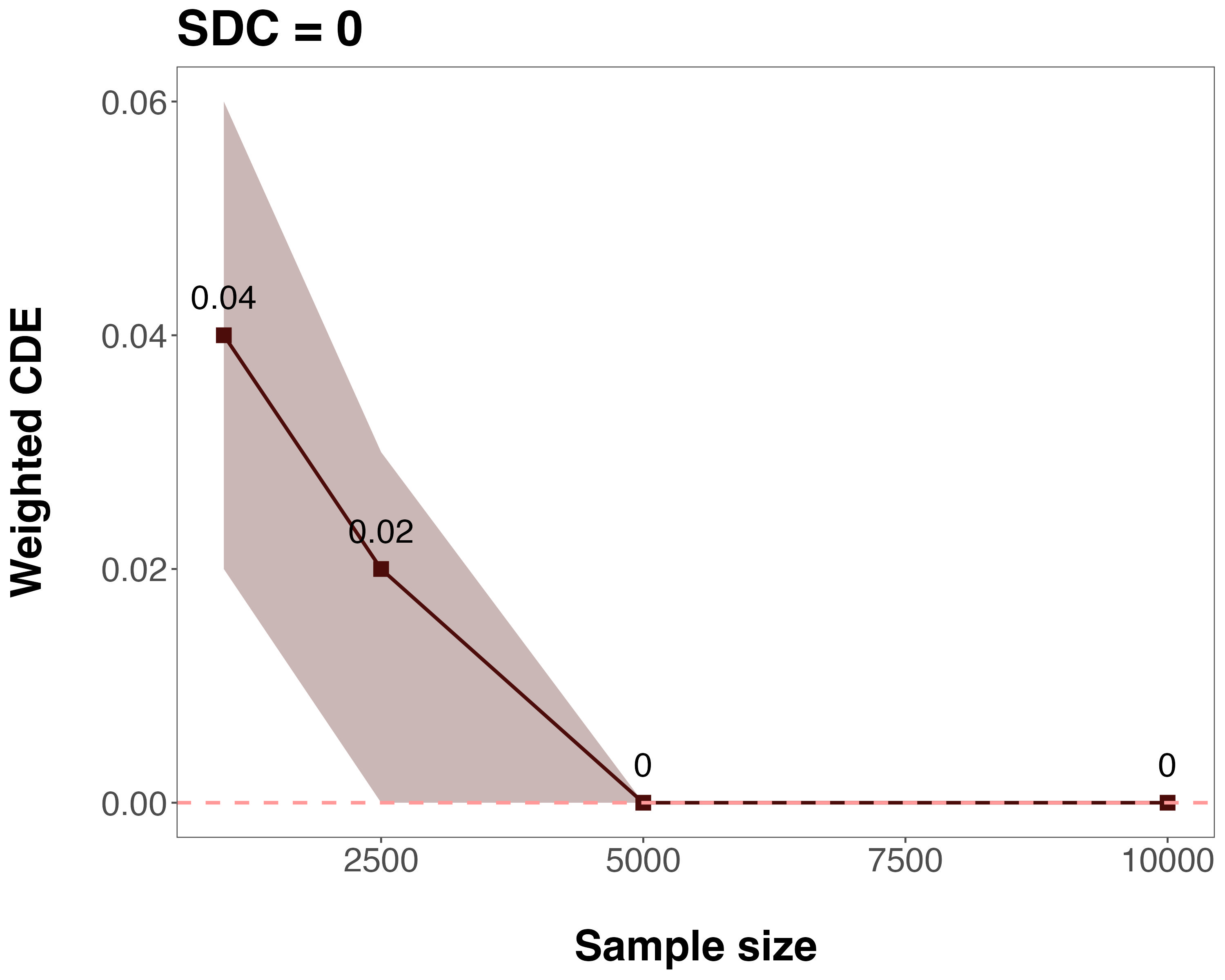}
    \hspace{5mm}
    \includegraphics[width=0.45\textwidth]{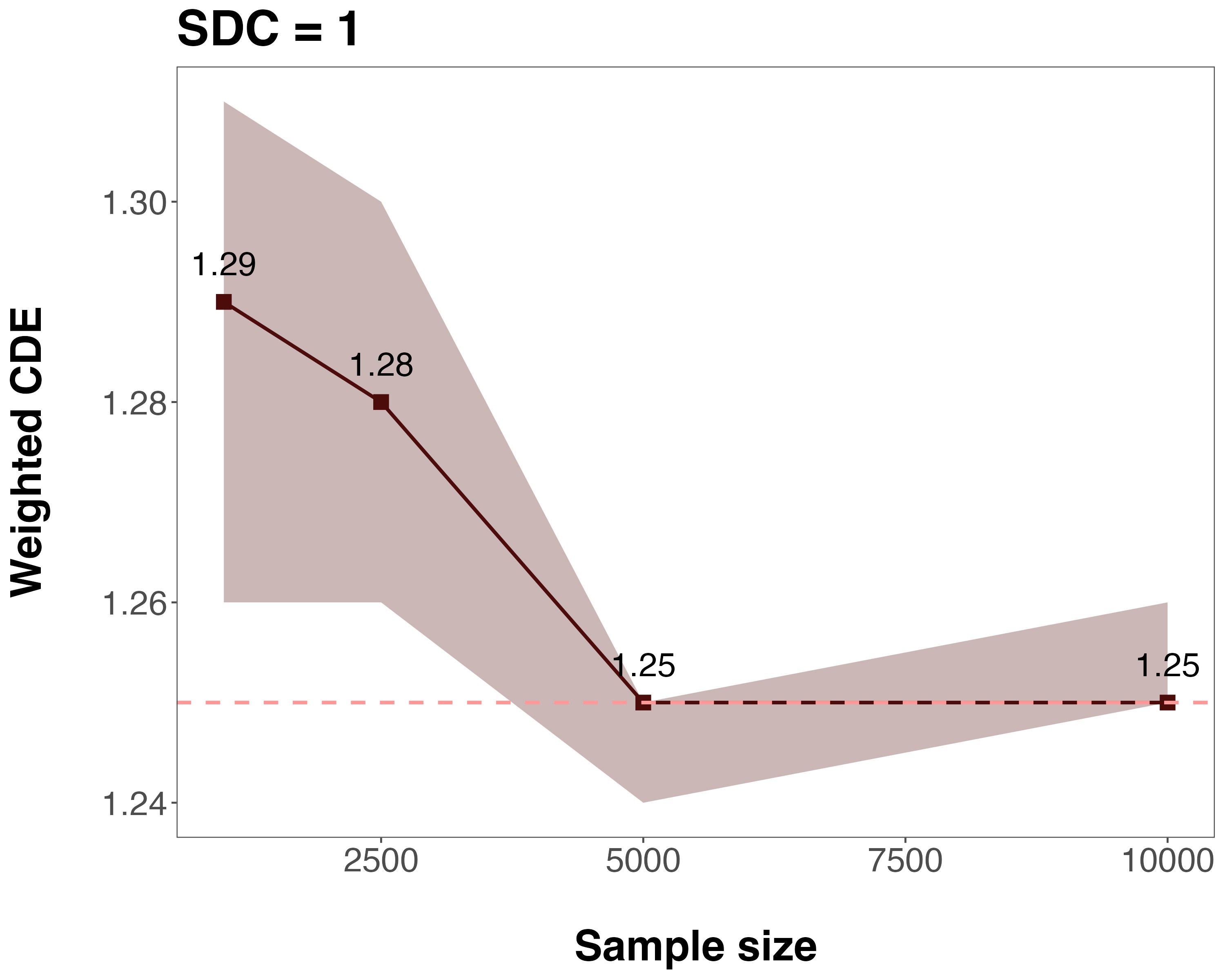}
    \caption{Mean WCDE estimates over 100 replicates per sample size for linear-Gaussian instantiations of Figure \ref{fig:eval_graph_cde} (95\% confidence intervals in shaded regions). Ground truth WCDE is denoted by the dashed line. LD3 used parametric Fisher-z tests ($\alpha = 0.01$), as the data generating process was known for this simulation. Estimates were obtained with linear regression (\texttt{https://scikit-learn.org/}). Note: This pipeline is for ease of explication in the illustrative setting only. For unknown data generating processes, we do not recommend this pipeline due to restrictive parametric assumptions.}
    \label{fig:wcde}
\end{figure}

\begin{figure}[!h]
    \centering
    \includegraphics[width=0.8\textwidth]{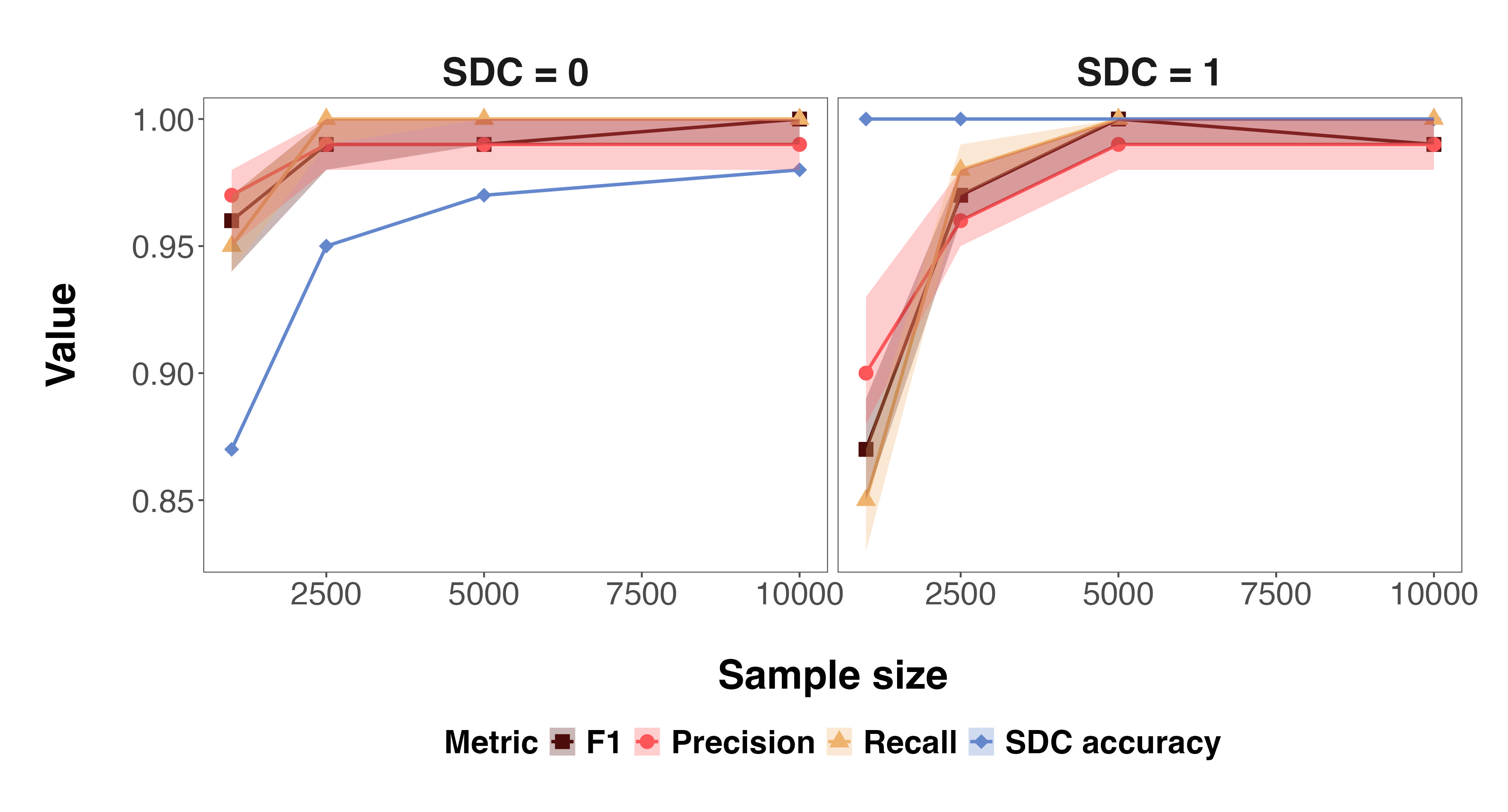}
    \caption{SDC accuracy, $\cde$ F1, precision, and recall across sample sizes when SDC = 0 and SDC = 1. Values were averaged over 100 replicates of linear-Gaussian instantiations of Figure \ref{fig:eval_graph_cde}, with 95\% confidence intervals in shaded regions. LD3 used parametric Fisher-z tests ($\alpha = 0.01$), as the data generating process was known for this simulation. }
    \label{fig:ld3_metrics}
\end{figure}

\paragraph{Results} To evaluate SDC and WCDE correctness, we tested LD3 on linear-Gaussian instantiations of an 18-node DAG with an M-structure (i.e., a member of $\z_{2 \notin de(Y)}$; \citealt{ding2015adjust}) and multiple members of $\z_1$, $\z_3$, and $\z_4$ (Figure \ref{fig:eval_graph_cde}). The WCDE was estimated using linear regression with $\cde$ as covariates. $\cde$ F1, precision, and recall and SDC accuracy converged toward 100\% as sample size increased. The WCDE converged toward the true direct effect with low variance (Figures \ref{fig:wcde}, \ref{fig:ld3_metrics}). \\
\vspace{5mm}

\vspace{5mm}

\clearpage


\subsection{Unnecessary Adjustment: Estimator Variance and VAS Interpretability}
\label{sec:adjusting_for_all}

\vspace{5mm}

\begin{figure}[!h]
    \centering
\begin{tikzpicture}[scale=0.11]
\tikzstyle{every node}+=[inner sep=0pt]
\draw [WildStrawberry,very thick] (29.3,-30.3) circle (3);
\draw (29.3,-30.3) node {$X$};
\draw [WildStrawberry,very thick] (46.5,-30.3) circle (3);
\draw (46.5,-30.3) node {$Y$};
\draw [black,very thick,fill=gray!20] (37.8,-21.9) circle (3);
\draw (37.8,-21.9) node {$Z_1$};
\draw [black,very thick,fill=gray!20] (37.8,-38.9) circle (3);
\draw (37.8,-38.9) node {$Z_3$};
\draw [black,very thick,fill=gray!20] (56.1,-20.5) circle (3);
\draw (56.1,-20.5) node {$Z_4$};
\draw [black,very thick] (27.6,-8.1) circle (3);
\draw (27.6,-8.1) node {$Z_{2a}$};
\draw [black,very thick] (37.8,-4) circle (3);
\draw (37.8,-4) node {$Z_{2b}$};
\draw [black,very thick] (49.1,-7.2) circle (3);
\draw (49.1,-7.2) node {$Z_{2c}$};
\draw [black,very thick] (17,-10.9) circle (3);
\draw (17,-10.9) node {$Z_{5a}$};
\draw [black,very thick] (17,-20.5) circle (3);
\draw (17,-20.5) node {$Z_{5b}$};
\draw [black,very thick] (17,-30.3) circle (3);
\draw (17,-30.3) node {$Z_{5c}$};
\draw [black,very thick] (17,-40.1) circle (3);
\draw (17,-40.1) node {$Z_{7a}$};
\draw [black,very thick] (17,-48.6) circle (3);
\draw (17,-48.6) node {$Z_{7b}$};
\draw [black,very thick] (29.3,-48.6) circle (3);
\draw (29.3,-48.6) node {$Z_{7c}$};
\draw [black,very thick] (46.5,-48.6) circle (3);
\draw (46.5,-48.6) node {$Z_{8a}$};
\draw [black,very thick] (56.1,-48.6) circle (3);
\draw (56.1,-48.6) node {$Z_{8b}$};
\draw [black,very thick] (56.1,-38.9) circle (3);
\draw (56.1,-38.9) node {$Z_{8c}$};
\draw [black,semithick,-{Stealth[width=5pt]}] (32.3,-30.3) -- (43.5,-30.3);
\draw [black,semithick,-{Stealth[width=5pt]}] (31.41,-32.43) -- (35.69,-36.77);
\draw [black,semithick,-{Stealth[width=5pt]}] (39.93,-36.79) -- (44.37,-32.41);
\draw [black,semithick,-{Stealth[width=5pt]}] (39.96,-23.98) -- (44.34,-28.22);
\draw [black,semithick,-{Stealth[width=5pt]}] (27.661,-27.79) arc (-151.05764:-200.18442:20.46);
\draw [black,semithick,-{Stealth[width=5pt]}] (29.076,-27.309) arc (-178.32893:-217.49211:32.949);
\draw [black,semithick,-{Stealth[width=5pt]}] (54,-22.64) -- (48.6,-28.16);
\draw [black,semithick,-{Stealth[width=5pt]}] (53.35,-19.3) -- (30.35,-9.3);
\draw [black,semithick,-{Stealth[width=5pt]}] (53.87,-18.49) -- (40.03,-6.01);
\draw [black,semithick,-{Stealth[width=5pt]}] (29.942,-27.371) arc (164.71562:114.08179:29.396);
\draw [black,semithick,-{Stealth[width=5pt]}] (54.7,-17.85) -- (50.5,-9.85);
\draw [black,semithick,-{Stealth[width=5pt]}] (29.3,-33.3) -- (29.3,-45.6);
\draw [black,semithick,-{Stealth[width=5pt]}] (27.63,-32.79) -- (18.67,-46.11);
\draw [black,semithick,-{Stealth[width=5pt]}] (26.95,-32.17) -- (19.35,-38.23);
\draw [black,semithick,-{Stealth[width=5pt]}] (20,-30.3) -- (26.3,-30.3);
\draw [black,semithick,-{Stealth[width=5pt]}] (19.35,-22.37) -- (26.95,-28.43);
\draw [black,semithick,-{Stealth[width=5pt]}] (18.61,-13.43) -- (27.69,-27.77);
\draw [black,semithick,-{Stealth[width=5pt]}] (35.67,-24.01) -- (31.43,-28.19);
\end{tikzpicture}
    \caption{The true $\cde$ for $X$ and $Y$ is in gray.}
    \label{fig:graph_variance}
\end{figure}

\vspace{5mm}

\begin{table}[!h]
    \centering
    \begin{tabular}{l l l l l}
    \toprule
        & \multicolumn{2}{c}{$\cde$} & \multicolumn{2}{c}{\textsc{All} $\z$} \\
        \cmidrule(lr){2-3}  \cmidrule(lr){4-5}
        \textit{$n$} & \textit{Mean} & \textit{Variance} & \textit{Mean} & \textit{Variance} \\
        \midrule
        100  & 6.98 & \textbf{0.83} & 6.97 & 7.79 \\
        1000 & 7.17 & \textbf{0.86} & 7.28 & 8.27 \\
        10000 & 6.91 & \textbf{1.16} & 7.25 & 9.04 \\
         \bottomrule
    \end{tabular}
    \caption{Given the DAG in Figure \ref{fig:graph_variance}, adjusting for $\z$ inflates WCDE variance relative to $\cde$ in finite samples (size $n$). True WCDE was 7. Data generating process was linear-Gaussian. 
    WCDE estimation used linear regression with 100 replicates per sample size. 
    }
    \label{tab:variance_all_z} 
\end{table}

\vspace{3mm}

\begin{table}[!h]
    \centering
    \begin{tabular}{l c c c c c c c}
    \toprule
        & \multicolumn{2}{c}{\textsc{All} $\z$} & \multicolumn{2}{c}{\textsc{True} $\cde$} & \multicolumn{3}{c}{\textsc{Pred} $\cde$}   \\
        \cmidrule(lr){2-3}  \cmidrule(lr){4-5} \cmidrule(lr){6-8}
        \textit{$n$} & \textit{Mean} & \textit{Variance} & \textit{Mean} & \textit{Variance} & \textit{Mean} & \textit{Variance} & $\cde$ \textit{F1} \\
        \midrule
        500 & 0.239 & 0.052 & 0.347 & \textbf{0.004} & 0.344 & \textbf{0.004} & 0.99 [0.98,1.0] \\
        1000 & -0.011 & 0.038 & 0.35 & \textbf{0.003} & 0.349 & \textbf{0.003} & 0.99 [0.98,1.0] \\
        10000 & 0.151 & 0.013 & 0.345 & \textbf{0.000} & 0.344 & \textbf{0.000} & 0.99 [0.98,1.0] \\
         \bottomrule
    \end{tabular}
    \caption{\textbf{Statistical Efficiency.} WCDE estimate mean and variance when adjusting for all $\z$, the true $\cde$ defined by our graphical criterion (\textsc{True} $\cde$), and the predicted $\cde$ returned by LD3 (\textsc{Pred} $\cde$) for the DAG in Figure \ref{fig:graph_variance}. Variance using all $\z$ was at least $12.6\times$ higher than using \textsc{Pred} $\cde$. F1 was high and variance for \textsc{Pred} $\cde$ was equal to the variance using the ground truth $\cde$. Variables were categorical with quadratic causal functions. WCDE estimation used double machine learning using \texttt{econml} (\texttt{https://econml.azurewebsites.net/}). Means are over 100 replicates per sample size (95\% confidence intervals in brackets). Though adjusting for all $\z$ will not theoretically induce bias in this setting, we see that mean WCDE values do diverge (and sometimes flip sign).} 
    \label{tab:variance_all_z_learned}
\end{table}



\vspace{5mm}

\begin{table}[!h]
    \centering
    \begin{adjustbox}{max width=\textwidth}
    \begin{tabular}{l c c c c c c c c}
    \toprule
         & $n$ & $|\z|$ & $|\cde|$ & Acc & F1 & Rec & Prec & Time \\
    \midrule
        \textsc{Ecoli} & $1k$ & 44 & 2.0 [2.0,2.0] & 1.00 [1.00,1.00] & 1.00 [1.00,1.00] & 1.00 [1.00,1.00] & 1.00 [1.00,1.00] & 0.016 [0.01,0.021] \\
        \textsc{Andes} & $50k$ & 221 & 4.5 [4.06,4.94] & 1.00 [1.00,1.00] & 0.95 [0.9,0.99] & 0.91 [0.83,0.99] & 1.00 [1.00,1.00] & 0.76 [0.737,0.784] \\
    \bottomrule
    \end{tabular}
    \end{adjustbox}
    \caption{\textbf{Interpretability.} $\cde$ discovered by LD3 can provide greater interpretability than adjusting for all $\z$, especially in large DAGs. We used the binary \textsc{Andes} \citep{conati1997line} and Gaussian \textsc{Ecoli} \citep{schafer2005shrinkage} benchmarks from \texttt{bnlearn}  \citep{scutari_learning_2010}. When the true parent set is small but total number of nodes is moderate (e.g., \textsc{Ecoli}) to large (e.g., \textsc{Andes}), forgoing structure learning by adjusting for $\z$ is a missed opportunity to provide mechanistic insights (e.g., causal fairness mechanisms). In the fairness setting, we can draw direct discrimination conclusions about the members of $\cde$, but not $\z$ alone. On average, eliminating extraneous variables took less than one second. \textsc{Ecoli} used Fisher-z tests ($\alpha = 0.001$) and \textsc{Andes} used $\chi^2$ ($\alpha = 0.01$). Values were averaged over ten replicates (95\% confidence intervals in brackets).}
    \label{tab:interpretability}
\end{table}


\clearpage

\subsection{\texttt{bnlearn} Benchmarks}

\vspace{10mm}

\begin{table}[!h]
    \centering
    \begin{adjustbox}{max width=\textwidth}
    \begin{tabular}{ l l l l l l l l }
    \toprule
        $n$ & \textsc{Method} & \textsc{Time} & \textsc{CI tests} & \textsc{Accuracy} & \textsc{F1} & \textsc{Precision} & \textsc{Recall}  \\
        \midrule
        250 & LD3 & \textbf{0.005 [0.003,0.008]} & \textbf{60.8 [60.41,61.19]} & 0.74 [0.71,0.78] & 0.7 [0.65,0.76] & \textbf{1.00 [1.00,1.00]} & 0.55 [0.48,0.62]  \\
         & LDECC & 0.092 [0.063,0.12] & 316.6 [220.98,412.22] & 0.61 [0.55,0.67] & 0.47 [0.33,0.6] & 0.9 [0.7,1.0] & 0.32 [0.2,0.43]  \\
         & MB-by-MB & 0.092 [0.047,0.136] & 268.4 [179.54,357.26] & 0.63 [0.57,0.68] & 0.5 [0.37,0.62] & 0.9 [0.7,1.0] & 0.35 [0.25,0.45]  \\
         & LDP & 0.009 [0.006,0.012] & 112.1 [90.82,133.38] & 0.76 [0.72,0.81] & 0.76 [0.71,0.8] & 0.94 [0.87,1.0] & 0.64 [0.59,0.68]  \\
         & PC & 0.118 [0.099,0.136] & 557.9 [524.73,591.07] & 0.64 [0.58,0.69] & 0.51 [0.39,0.64] & 0.9 [0.7,1.0] & 0.36 [0.27,0.46]  \\
         & DirectLiNGAM & 0.142 [0.138,0.145] & - & 0.63 [0.55,0.7] & 0.56 [0.43,0.69] & 0.85 [0.75,0.95] & 0.48 [0.32,0.63]  \\
         & NOTEARS & 1.216 [0.985,1.447] & - & \textbf{0.79 [0.75,0.82]} & \textbf{0.79 [0.75,0.82]} & 0.91 [0.86,0.96] & \textbf{0.7 [0.65,0.75]}  \\
         \midrule
        500 & LD3 & \textbf{0.006 [0.004,0.008]} & \textbf{60.8 [60.41,61.19]} & \textbf{0.85 [0.79,0.91]} & \textbf{0.84 [0.77,0.91]} & \textbf{1.00 [1.00,1.00]} & \textbf{0.74 [0.64,0.84]}  \\
         & LDECC & 0.137 [0.104,0.169] & 458.2 [354.85,561.55] & 0.68 [0.63,0.73] & 0.6 [0.52,0.69] & 0.98 [0.94,1.0] & 0.45 [0.37,0.53]  \\
         & MB-by-MB & 0.127 [0.113,0.141] & 370 [328.41,411.59] & 0.64 [0.61,0.68] & 0.55 [0.48,0.61] & 0.98 [0.94,1.0] & 0.39 [0.32,0.46]  \\
         & LDP & 0.01 [0.005,0.016] & 84.3 [66.81,101.79] & 0.78 [0.73,0.82] & 0.77 [0.72,0.82] & 0.94 [0.88,1.0] & 0.66 [0.6,0.73]  \\
         & PC & 0.181 [0.162,0.201] & 879.6 [843.24,915.96] & 0.61 [0.56,0.67] & 0.47 [0.38,0.57] & \textbf{1.00 [1.00,1.00]} & 0.32 [0.23,0.42]  \\
         & DirectLiNGAM & 0.152 [0.144,0.16] & - & 0.56 [0.46,0.67] & 0.42 [0.18,0.65] & 0.43 [0.19,0.67] & 0.41 [0.18,0.65]  \\
         & NOTEARS & 1.092 [1.025,1.158] & - & 0.79 [0.76,0.82] & 0.78 [0.75,0.81] & 0.95 [0.9,1.0] & 0.66 [0.63,0.7]  \\
         \midrule
        1000 & LD3 & 0.009 [0.007,0.011] & \textbf{61 [61,61]} & \textbf{0.97 [0.95,0.99]} & \textbf{0.99 [0.95,0.99]} & \textbf{1.00 [1.00,1.00]} & \textbf{0.95 [0.91,0.99]}  \\
         & LDECC & 0.217 [0.185,0.248] & 669.3 [560.92,777.68] & 0.66 [0.61,0.72] & 0.57 [0.45,0.68] & \textbf{1.00 [1.00,1.00]} & 0.41 [0.31,0.52]  \\
         & MB-by-MB & 0.201 [0.175,0.228] & 592.7 [538.57,646.83] & 0.69 [0.66,0.72] & 0.63 [0.58,0.68] & \textbf{1.00 [1.00,1.00]} & 0.46 [0.41,0.51]  \\
         & LDP & \textbf{0.006 [0.003,0.009]} & 75 [63.46,86.54] & 0.95 [0.91,0.99] & 0.95 [0.91,0.99] & 0.99 [0.96,1.0] & 0.92 [0.87,0.98]  \\
         & PC & 0.289 [0.272,0.306] & 1333.9 [1249.08,1418.72] & 0.55 [0.5,0.6] & 0.33 [0.21,0.45] & 0.8 [0.54,1.06] & 0.21 [0.13,0.29]  \\
         & DirectLiNGAM & 0.366 [0.326,0.407] & - & 0.62 [0.55,0.69] & 0.56 [0.4,0.73] & 0.66 [0.51,0.8] & 0.54 [0.34,0.73]  \\
         & NOTEARS & 1.694 [1.571,1.818] & - & 0.8 [0.77,0.83] & 0.79 [0.75,0.82] & \textbf{1.00 [1.00,1.00]} & 0.65 [0.6,0.7]  \\
        \bottomrule
    \end{tabular}
    \end{adjustbox}
    \caption{Results on the \textsc{Sangiovese} benchmark from \texttt{bnlearn} \citep{scutari_learning_2010}, a linear-Gaussian model of Tuscan Sangiovese grape production \citep{magrini2017conditional}. Ten replicate datasets were sampled at $n = [250,500,1000]$. Means are given with 95\% confidence intervals in brackets. Exposure was \textsc{pH} and outcome was \textsc{GrapeW}. All constraint-based methods used Fisher-z tests ($\alpha = 0.01$). NOTEARS hyperparameters were $\lambda_1 = 0.001$ and loss type $l_2$. DirectLiNGAM: \texttt{https://lingam.readthedocs.io/}; PC, LDECC, MB-by-MB: \texttt{https://github.com/acmi-lab/local-causal-discovery}; NOTEARS: \texttt{https://github.com/xunzheng/notears}; LDP: \texttt{https://github.com/jmaasch/ldp}.}
    \label{tab:grapes}
\end{table}

\vspace{10mm}

\begin{figure}[!h]
    \centering
    \includegraphics[width=0.45\textwidth]{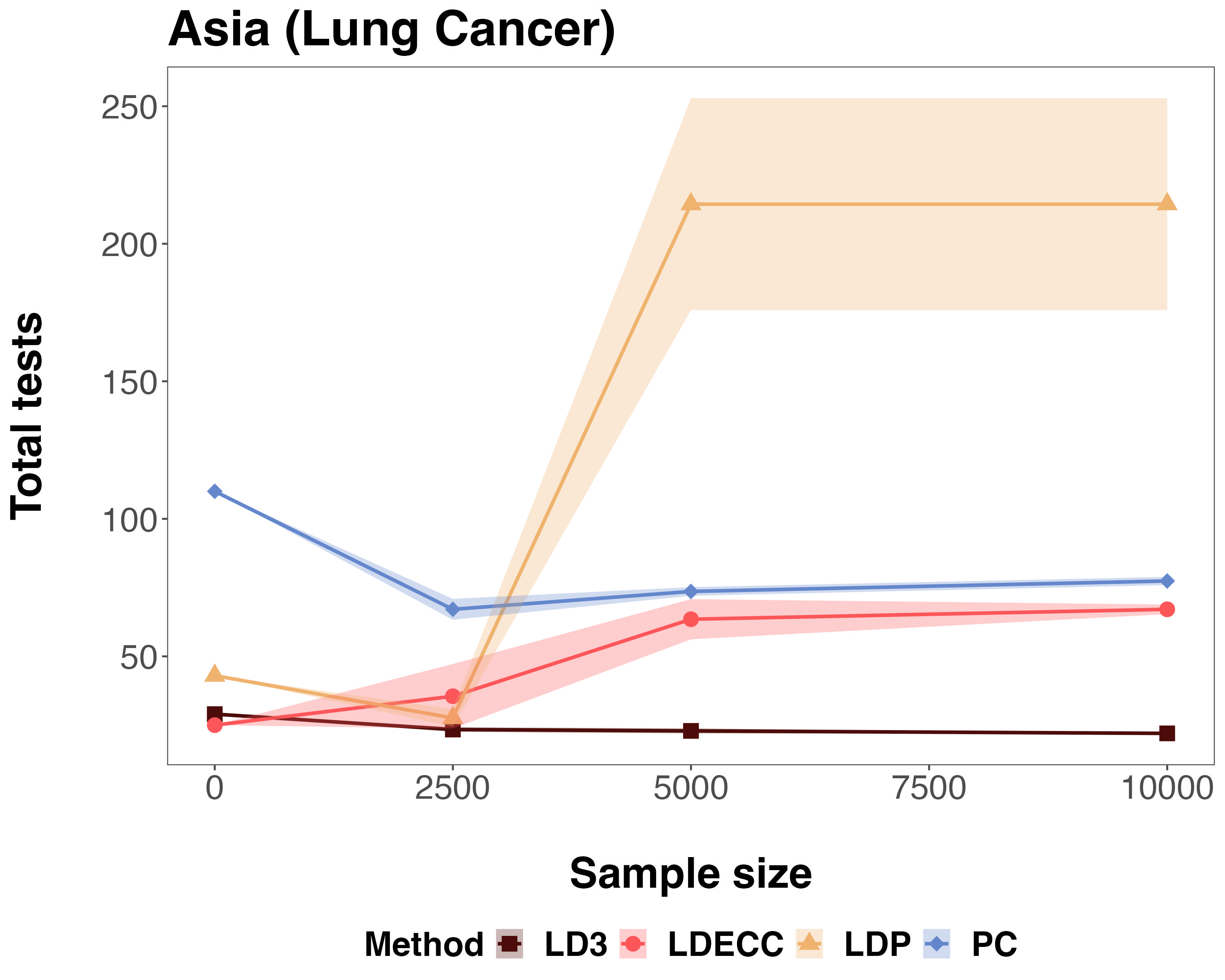}
    \hspace{5mm}
    \includegraphics[width=0.45\textwidth]{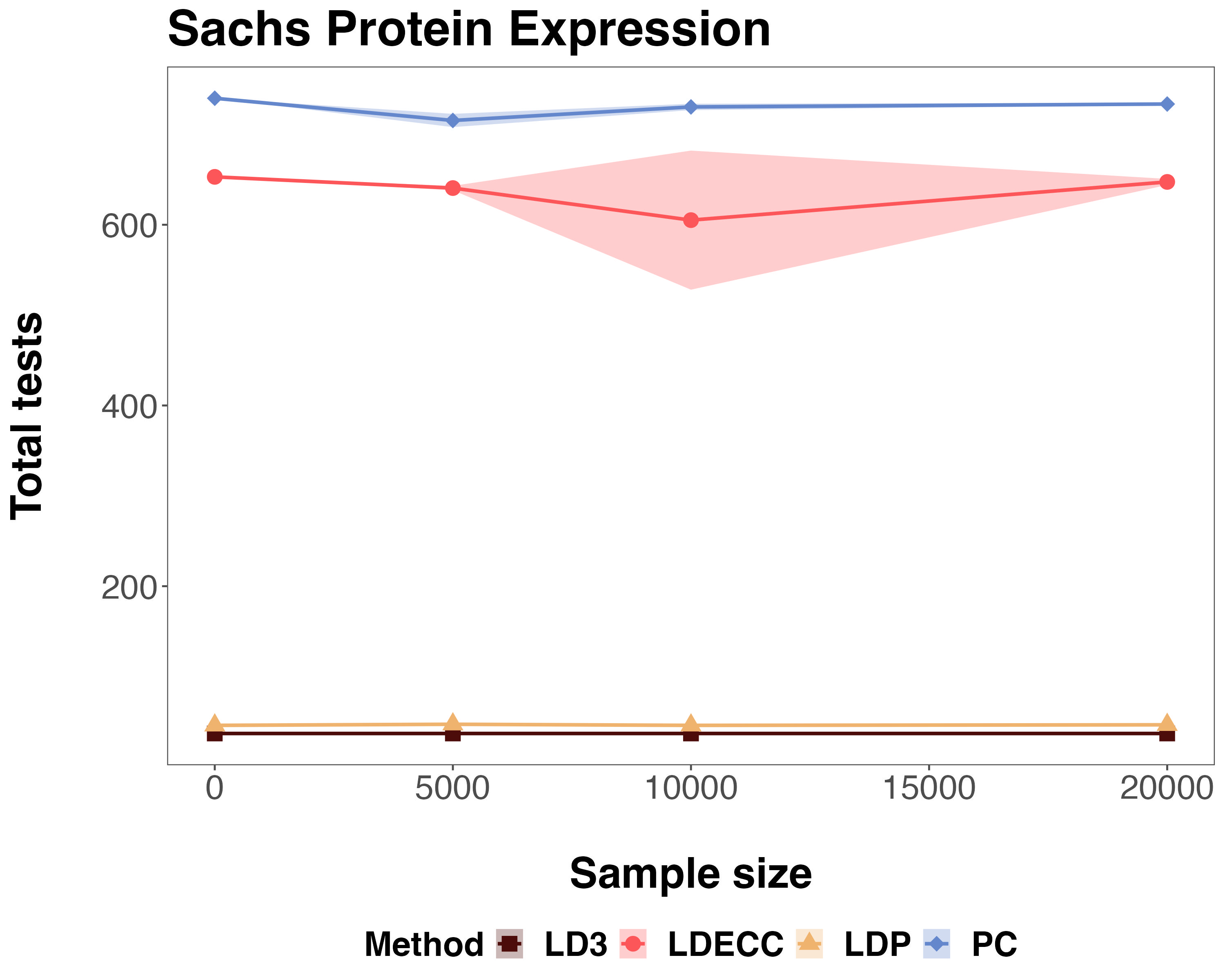}
    \caption{Total number of CI tests performed for the \textsc{Asia} \citep{lauritzen_local_1988} and \textsc{Sachs} \citep{sachs2005causal} benchmarks from \texttt{bnlearn} ({\texttt{https://www.bnlearn.com/bnrepository/}}) \citep{scutari_learning_2010} ($\chi^2$ tests; $\alpha = 0.001$; 95\% confidence intervals in shaded regions). For \textsc{Asia} (\textsc{either} $\to$ \textsc{dysp}), $|\z| = 6$ and $|\cde| = 1$. For \textsc{Sachs} (\textsc{erk} $\to$ \textsc{akt}), $|\z| = 11$ and $|\cde| = 1$. Raw data are reported in Tables \ref{tab:asia_time_tests} and \ref{tab:sachs_time_tests}.}
    \label{fig:bnlearn_tests}
\end{figure}

\begin{table}[!ht]
    \centering
    \begin{adjustbox}{max width=\textwidth}
    \begin{tabular}{l l c c c c c c}
    \toprule
    & & \multicolumn{6}{c}{\textsc{\textbf{Asia (Lung Cancer)}}} \\
    \cmidrule(lr){3-8}
    & & \multicolumn{3}{c}{SDC = 1: either $\to$ dysp} & \multicolumn{3}{c}{SDC = 0: xray $\not\to$ dysp} \\
    \cmidrule(lr){3-5}  \cmidrule(lr){6-8}
    & & \textit{F1} & \textit{Prec} & \textit{Rec} & \textit{F1} & \textit{Prec} & \textit{Rec} \\
    \midrule
    \parbox[t]{1.25cm}{\multirow{4}{*}{\textit{Oracle}}}
    & LD3  & \textbf{1.00} & \textbf{1.00} & \textbf{1.00} & \textbf{1.00} & \textbf{1.00} & \textbf{1.00} \\
    & PC$_\cup$ & \textbf{1.00} & \textbf{1.00} & \textbf{1.00} & \textbf{1.00} & \textbf{1.00} & \textbf{1.00}  \\
    & PC$_\cap$& \textbf{1.00} & \textbf{1.00} & \textbf{1.00} & \textbf{1.00} & \textbf{1.00} & \textbf{1.00}  \\
    & LDECC$_\cup$ & \textbf{1.00} & \textbf{1.00} & \textbf{1.00} & \textbf{1.00} & \textbf{1.00} & \textbf{1.00}  \\
    & LDECC$_\cap$ & \textbf{1.00} & \textbf{1.00} & \textbf{1.00} & \textbf{1.00} & \textbf{1.00} & \textbf{1.00}  \\
    & LDP$_{pa}$ & \textbf{1.00} & \textbf{1.00} & \textbf{1.00} & \textbf{1.00} & \textbf{1.00} & \textbf{1.00} \\
    & LDP & 0.50 & 0.33 & \textbf{1.00} & 0.67 & 0.50 & \textbf{1.00}\\
    \midrule
    \parbox[t]{1.25cm}{\multirow{4}{*}{$n = 2.5k$}}
    & LD3  & \textbf{1.00 [1.00,1.00]} & \textbf{1.00 [1.00,1.00]} & \textbf{1.00 [1.00,1.00]} & 0.80 [0.69,0.91] & \textbf{1.00 [1.00,1.00]} &  0.70 [0.54,0.86] \\
    & PC$_\cup$  & 0.00 [0.00,0.00] & 0.00 [0.00,0.00] & 0.00 [0.00,0.00] &  0.00 [0.00,0.00] & 0.00 [0.00,0.00] & 0.00 [0.00,0.00] \\
    & PC$_\cap$  & 0.00 [0.00,0.00] & 0.00 [0.00,0.00] & 0.00 [0.00,0.00] &  0.00 [0.00,0.00] & 0.00 [0.00,0.00] & 0.00 [0.00,0.00] \\
    & LDECC$_\cup$  & 0.75 [0.61,0.89] & 0.65 [0.46,0.84] & \textbf{1.00 [1.00,1.00]} & 0.67 [0.49,0.85] & 0.65 [0.46,0.84] & 0.70 [0.54,0.86] \\
    & LDECC$_\cap$  & 0.75 [0.61,0.89] & 0.65 [0.46,0.84] & \textbf{1.00 [1.00,1.00]} & 0.67 [0.49,0.85] & 0.65 [0.46,0.84] & 0.70 [0.54,0.86] \\
    & LDP$_{pa}$  & \textbf{1.00 [1.00,1.00]} & \textbf{1.00 [1.00,1.00]} & \textbf{1.00 [1.00,1.00]} & \textbf{0.83 [0.72,0.94]} & \textbf{1.00 [1.00,1.00]} & \textbf{0.75 [0.59,0.91]} \\
    & LDP & 0.26 [0.24,0.28] & 0.15 [0.14,0.16] & 0.97 [0.90,1.00] & 0.14 [0.13,0.14] & 0.07 [0.07,0.08] & \textbf{1.00 [1.00,1.00]} \\
    \midrule
    \parbox[t]{1.25cm}{\multirow{4}{*}{$n = 5k$}}
    & LD3  & \textbf{1.00 [1.00,1.00]} & \textbf{1.00 [1.00,1.00]} & \textbf{1.00 [1.00,1.00]} & 0.90 [0.80,1.00] & \textbf{1.00 [1.00,1.00]} & 0.85 [0.70,1.00] \\
    & PC$_\cup$  & 0.00 [0.00,0.00] & 0.00 [0.00,0.00] & 0.00 [0.00,0.00] &  0.00 [0.00,0.00] & 0.00 [0.00,0.00] & 0.00 [0.00,0.00] \\
    & PC$_\cap$ & 0.00 [0.00,0.00] & 0.00 [0.00,0.00] & 0.00 [0.00,0.00] &  0.00 [0.00,0.00] & 0.00 [0.00,0.00] & 0.00 [0.00,0.00] \\
    & LDECC$_\cup$ & 0.60 [0.47,0.73] & 0.47 [0.29,0.64] & \textbf{1.00 [1.00,1.00]} & 0.52 [0.36,0.68] & 0.47 [0.29,0.64]  & 0.60 [0.47,0.73] \\
    & LDECC$_\cap$  & 0.60 [0.47,0.73] & 0.47 [0.29,0.64] & \textbf{1.00 [1.00,1.00]} & 0.52 [0.36,0.68] & 0.47 [0.29,0.64]  & 0.60 [0.47,0.73] \\
    & LDP$_{pa}$  & \textbf{1.00 [1.00,1.00]} & \textbf{1.00 [1.00,1.00]} & \textbf{1.00 [1.00,1.00]} & \textbf{0.93 [0.85,1.00]} & \textbf{1.00 [1.00,1.00]} & \textbf{0.90 [0.77,1.00]} \\
    & LDP & 0.65 [0.62,0.68] & 0.48 [0.45,0.52] & \textbf{1.00 [1.00,1.00]} & 0.65 [0.62,0.67] & 0.48 [0.45,0.51] & \textbf{1.00 [1.00,1.00]} \\
    \midrule
    \parbox[t]{1.25cm}{\multirow{4}{*}{$n = 10k$}}
    & LD3  & \textbf{1.00 [1.00,1.00]} & \textbf{1.00 [1.00,1.00]} & \textbf{1.00 [1.00,1.00]} & \textbf{0.90 [0.80,1.00]} & \textbf{1.00 [1.00,1.00]} & \textbf{0.85 [0.70,1.00]} \\
    & PC$_\cup$  & 0.00 [0.00,0.00] & 0.00 [0.00,0.00] & 0.00 [0.00,0.00] &  0.00 [0.00,0.00] & 0.00 [0.00,0.00] & 0.00 [0.00,0.00] \\
    & PC$_\cap$  & 0.00 [0.00,0.00] & 0.00 [0.00,0.00] & 0.00 [0.00,0.00] &  0.00 [0.00,0.00] & 0.00 [0.00,0.00] & 0.00 [0.00,0.00] \\
    & LDECC$_\cup$  & 0.50 [0.50,0.50] & 0.33 [0.33,0.33] & 0.00 [0.00,0.00] & 0.40 [0.40,0.40] & 0.33 [0.33,0.33] & 0.50 [0.50,0.50] \\
    & LDECC$_\cap$  & 0.50 [0.50,0.50] & 0.33 [0.33,0.33] & 0.00 [0.00,0.00] & 0.40 [0.40,0.40] & 0.33 [0.33,0.33] & 0.50 [0.50,0.50] \\
    & LDP$_{pa}$ & \textbf{1.00 [1.00,1.00]} & \textbf{1.00 [1.00,1.00]} & \textbf{1.00 [1.00,1.00]} & \textbf{0.90 [0.80,1.00]} & \textbf{1.00 [1.00,1.00]} & \textbf{0.85 [0.70,1.00]} \\
    & LDP & 0.67 [0.67,0.67] & 0.50 [0.50,0.50] & \textbf{1.00 [1.00,1.00]} & 0.64 [0.61,0.67] & 0.47 [0.44,0.5] & \textbf{1.00 [1.00,1.00]} \\
    \bottomrule
    \end{tabular}
    \end{adjustbox}
    \caption{Empirical performance on the \textsc{Asia} (Lung Cancer) Bayesian network dataset \citep{lauritzen_local_1988} from \texttt{bnlearn} \citep{scutari_learning_2010}. Binary data were evaluated with $\chi^2$ independence tests ($\alpha = 0.001$; 95\% confidence intervals in brackets). Unexpected decreases in baseline performance for increasing sample size might be due to the difficulty of this dataset for structure learning, as noted in the \texttt{bnlearn} documentation: ``Standard learning algorithms are not able to recover the true structure of the network because of the presence of a node (E) with conditional probabilities equal to both 0 and 1" (\texttt{https://www.bnlearn.com/documentation/man/asia.html}).}
    \label{tab:asia}
\end{table}



\vspace{10mm}

\begin{table}[!ht]
    \centering
    \begin{adjustbox}{max width=0.9\textwidth}
    \begin{tabular}{l l c c c c}
    \toprule
    & & \multicolumn{4}{c}{\textsc{\textbf{Asia (Lung Cancer)}}} \\
    \cmidrule(lr){3-6}
    & & \multicolumn{2}{c}{SDC = 1: either $\to$ dysp} & \multicolumn{2}{c}{SDC = 0: xray $\not\to$ dysp} \\
    \cmidrule(lr){3-4}  \cmidrule(lr){5-6}
    & & \textit{Time} & \textit{Tests} & \textit{Time} & \textit{Tests} \\
    \midrule
    \parbox[t]{1.25cm}{\multirow{4}{*}{\textit{Oracle}}}  & LD3 & \textbf{0.002} & 29 & \textbf{0.002} & 31 \\
    & PC & 0.007 & 110 & 0.007 & 115 \\
    & LDECC & 0.003 & \textbf{25} & 0.004 & \textbf{25} \\
    & LDP & 0.003 & 43 & 0.003 & 50 \\
    \midrule
    \parbox[t]{1.25cm}{\multirow{4}{*}{$n = 2.5k$}}  & LD3 & \textbf{0.003 [0.002,0.003]} & \textbf{23.4 [22.56,24.24]} & \textbf{0.003 [0.003,0.003]} & \textbf{27.2 [26.56,27.84]} \\
    & PC & 0.032 [0.03,0.034] & 67.1 [63.31,70.89] & 0.033 [0.031,0.035] & 70.5 [66.79,74.21] \\
    & LDECC & 0.024 [0.016,0.033] & 35.5 [23.78,47.22] & 0.018 [0.013,0.023] & 35.5 [23.78,47.22] \\
    & LDP & 0.003 [0.003,0.003] & 27.6 [24.41,30.79] & 0.003 [0.003,0.003] & 31.0 [27.94,34.06] \\
    \midrule
    \parbox[t]{1.25cm}{\multirow{4}{*}{$n = 5k$}}  & LD3 & \textbf{0.004 [0.003,0.004]} & \textbf{22.9 [21.96,23.84]} & \textbf{0.004 [0.003,0.004]} & \textbf{26.6 [26.0,27.2]} \\
    & PC & 0.056 [0.054,0.057] & 73.6 [72.02,75.18] & 0.058 [0.057,0.06] &  77.1 [75.29,78.91] \\
    & LDECC & 0.049 [0.044,0.054] & 63.5 [56.21,70.79] & 0.047 [0.042,0.053] & 63.5 [56.21,70.79] \\
    & LDP & 0.073 [0.056,0.091] & 214.4 [175.9,252.9] & 0.039 [0.033,0.044] & 192.0 [167.74,216.26] \\
    \midrule
    \parbox[t]{1.25cm}{\multirow{4}{*}{$n = 10k$}}  & LD3 & \textbf{0.004 [0.004,0.004]} & \textbf{22.0 [22.0,22.0]} & \textbf{0.005 [0.005,0.005]} & \textbf{26.6 [26.0,27.2]} \\
    & PC & 0.100 [0.097,0.103] & 77.4 [75.9,78.9] & 0.105 [0.102,0.109] & 82.3 [80.38,84.22]  \\
    & LDECC & 0.096 [0.083,0.109] & 67.1 [65.34,68.86] & 0.087 [0.084,0.09] & 67.1 [65.34,68.86] \\
    & LDP & 0.073 [0.056,0.091] & 214.4 [175.9,252.9] & 0.039 [0.033,0.044] & 192.0 [167.74,216.26] \\
    \bottomrule
    \end{tabular}
    \end{adjustbox}
    \caption{Total number of independence tests performed and runtimes in seconds (95\% confidence intervals in brackets).}
    \label{tab:asia_time_tests}
\end{table}
\begin{table}[!h]
    \centering
    \begin{adjustbox}{max width=\textwidth}
    \begin{tabular}{l l c c c c c c}
    \toprule
    & & \multicolumn{6}{c}{\textsc{\textbf{Sachs Protein Expression}}} \\
    \cmidrule(lr){3-8}
    & & \multicolumn{3}{c}{SDC = 1: ERK $\to$ AKT} & \multicolumn{3}{c}{SDC = 0: JNK $\not\to$ P38} \\
    \cmidrule(lr){3-5}  \cmidrule(lr){6-8}
    & & \textit{F1} & \textit{Prec} & \textit{Rec} & \textit{F1} & \textit{Prec} & \textit{Rec} \\
    \midrule
    \parbox[t]{1cm}{\multirow{4}{*}{\textit{Oracle}}} & LD3 & \textbf{1.00} & \textbf{1.00} & \textbf{1.00} &  \textbf{1.00} & \textbf{1.00} & \textbf{1.00} \\
    & PC$_\cup$ & \textbf{1.00} & \textbf{1.00} & \textbf{1.00} & \textbf{1.00} & \textbf{1.00} & \textbf{1.00} \\
    & PC$_\cap$  & 0.00 & 0.00 & 0.00 & 0.00 & 0.00 & 0.00 \\
    & LDECC$_\cup$ & \textbf{1.00} & \textbf{1.00} & \textbf{1.00} & \textbf{1.00} & \textbf{1.00} & \textbf{1.00} \\
    & LDECC$_\cap$ & 0.00 & 0.00 & 0.00 & 0.00 & 0.00 & 0.00 \\
    & LDP$_{pa}$ & \textbf{1.00} & \textbf{1.00} & \textbf{1.00} & \textbf{1.00} & \textbf{1.00} & \textbf{1.00} \\
    & LDP & 0.29 & 0.16 & \textbf{1.00} & 0.50 & 0.33 & \textbf{1.00} \\
    \midrule
    \parbox[t]{1cm}{\multirow{4}{*}{$n = 5k$}} & LD3 & 0.97 [0.90,1.00] & 0.95 [0.85,1.00] & \textbf{1.00 [1.00,1.00]} & 0.96 [0.91,1.00] & 0.93 [0.85,1.00] & \textbf{1.00 [1.00,1.00]} \\
    & PC$_\cup$ & 0.00 [0.00,0.00] & 0.00 [0.00,0.00] & 0.00 [0.00,0.00] & 0.77 [0.51,1.00] & 0.80 [0.54,1.00] & 0.75 [0.49,1.00] \\
    & PC$_\cap$ & 0.00 [0.00,0.00] & 0.00 [0.00,0.00] & 0.00 [0.00,0.00] & 0.73 [0.48,0.99] & 0.80 [0.54,1.00] & 0.70 [0.44,0.96]  \\
    & LDECC$_\cup$ & \textbf{1.00 [1.00,1.00]} & \textbf{1.00 [1.00,1.00]} & \textbf{1.00 [1.00,1.00]} & \textbf{1.00 [1.00,1.00]} & \textbf{1.00 [1.00,1.00]} & \textbf{1.00 [1.00,1.00]} \\
    & LDECC$_\cap$ & 0.10 [-0.1,0.30] & 0.10 [-0.1,0.30] & 0.10 [-0.1,0.30] & 0.10 [-0.1,0.30] & 0.10 [-0.1,0.30] & 0.10 [-0.1,0.30] \\
    & LDP$_{pa}$ & 0.93 [0.85,1.02] & 0.90 [0.77,1.00] & \textbf{1.00 [1.00,1.00]} & 0.96 [0.91,1.00] & 0.93 [0.85,1.00] & \textbf{1.00 [1.00,1.00]} \\
    & LDP & 0.29 [0.29,0.29] & 0.17 [0.17,0.17] & \textbf{1.00 [1.00,1.00]} & 0.50 [0.50,0.50] & 0.33 [0.33,0.33] & \textbf{1.00 [1.00,1.00]} \\
    \midrule
    \parbox[t]{1cm}{\multirow{4}{*}{$n = 10k$}}
    & LD3 & \textbf{1.00 [1.00,1.00]} & \textbf{1.00 [1.00,1.00]} & \textbf{1.00 [1.00,1.00]} &  \textbf{1.00 [1.00,1.00]} & \textbf{1.00 [1.00,1.00]} & \textbf{1.00 [1.00,1.00]} \\
    & PC$_\cup$ & 0.50 [0.17,0.83] & 0.50 [0.17,0.83] & 0.50 [0.17,0.83] & \textbf{1.00 [1.00,1.00]} &\textbf{1.00 [1.00,1.00]}  & \textbf{1.00 [1.00,1.00]} \\
    & PC$_\cap$ &  0.00 [0.00,0.00] & 0.00 [0.00,0.00] & 0.00 [0.00,0.00] & 0.47 [0.16,0.78] & 0.50 [0.17,0.83] & 0.45 [0.14,0.76] \\
    & LDECC$_\cup$ & \textbf{1.00 [1.00,1.00]} & \textbf{1.00 [1.00,1.00]} & \textbf{1.00 [1.00,1.00]} &  \textbf{1.00 [1.00,1.00]} & \textbf{1.00 [1.00,1.00]} & \textbf{1.00 [1.00,1.00]} \\
    & LDECC$_\cap$ & 0.10 [-0.1,0.30] & 0.10 [-0.1,0.30] & 0.10 [-0.1,0.30] & 0.30 [0.00,0.60] & 0.30 [0.00,0.60] & 0.30 [0.00,0.60]  \\ 
    & LDP$_{pa}$ & \textbf{1.00 [1.00,1.00]} & \textbf{1.00 [1.00,1.00]} & \textbf{1.00 [1.00,1.00]} &  \textbf{1.00 [1.00,1.00]} & \textbf{1.00 [1.00,1.00]} & \textbf{1.00 [1.00,1.00]} \\
    & LDP & 0.29 [0.29,0.29] & 0.17 [0.17,0.17] & \textbf{1.00 [1.00,1.00]} & 0.50 [0.50,0.50] & 0.33 [0.33,0.33] & \textbf{1.00 [1.00,1.00]} \\
    \midrule
    \parbox[t]{1cm}{\multirow{4}{*}{$n = 20k$}}
    & LD3 & 0.97 [0.90,1.00] & 0.95 [0.85,1.00] & \textbf{1.00 [1.00,1.00]} &  \textbf{1.00 [1.00,1.00]} & \textbf{1.00 [1.00,1.00]} & \textbf{1.00 [1.00,1.00]} \\
    & PC$_\cup$ & 0.90 [0.70,1.00] & 0.90 [0.70,1.00] & 0.90 [0.70,1.00] & \textbf{1.00 [1.00,1.00]} & \textbf{1.00 [1.00,1.00]} & \textbf{1.00 [1.00,1.00]} \\
    & PC$_\cap$ & 0.00 [0.00,0.00] & 0.00 [0.00,0.00] & 0.00 [0.00,0.00] & 0.10 [-0.1,0.30] & 0.10 [-0.1,0.30] & 0.10 [-0.1,0.30] \\
    & LDECC$_\cup$ & \textbf{1.00 [1.00,1.00]} & \textbf{1.00 [1.00,1.00]} & \textbf{1.00 [1.00,1.00]} &  \textbf{1.00 [1.00,1.00]} & \textbf{1.00 [1.00,1.00]} & \textbf{1.00 [1.00,1.00]} \\
    & LDECC$_\cap$ & 0.00 [0.00,0.00] & 0.00 [0.00,0.00] & 0.00 [0.00,0.00] & 0.00 [0.00,0.00] & 0.00 [0.00,0.00] & 0.00 [0.00,0.00] \\
    & LDP$_{pa}$ & 0.97 [0.90,1.00] & 0.95 [0.85,1.00] & \textbf{1.00 [1.00,1.00]} &  \textbf{1.00 [1.00,1.00]} & \textbf{1.00 [1.00,1.00]} & \textbf{1.00 [1.00,1.00]} \\
    & LDP & 0.29 [0.29,0.29] & 0.17 [0.17,0.17] & \textbf{1.00 [1.00,1.00]} & 0.50 [0.50,0.50] & 0.33 [0.33,0.33] & \textbf{1.00 [1.00,1.00]} \\
    \bottomrule
    \end{tabular}
    \end{adjustbox}
    \caption{Performance on the \textsc{Sachs} protein dataset \citep{sachs2005causal} from \texttt{bnlearn} \citep{scutari_learning_2010}. Discrete data were evaluated with $\chi^2$ independence tests ($\alpha = 0.001$; 95\% confidence intervals in brackets).}
    \label{tab:sachs}
\end{table}



\begin{table}[!h]
    \centering
    \begin{adjustbox}{max width=\textwidth}
    \begin{tabular}{l l c c c c}
    \toprule
    & & \multicolumn{4}{c}{\textsc{\textbf{Sachs Protein Expression}}} \\
    \cmidrule(lr){3-6}
    & & \multicolumn{2}{c}{SDC = 1: ERK $\to$ AKT} & \multicolumn{2}{c}{SDC = 0: JNK $\not\to$ P38} \\
    \cmidrule(lr){3-4}  \cmidrule(lr){5-6}
    & & \textit{Time} & \textit{Tests} & \textit{Time} & \textit{Tests} \\
    \midrule
    \parbox[t]{1cm}{\multirow{4}{*}{\textit{Oracle}}}  & LD3 & \textbf{0.013} & \textbf{37} & \textbf{0.013} & \textbf{37} \\
    & PC & 0.077 & 740 & 0.079 & 740 \\
    & LDECC & 0.079 & 653 & 0.081 & 646 \\
    & LDP & 0.017 & 46 & 0.017 & 52 \\
    \midrule
    \parbox[t]{1cm}{\multirow{4}{*}{$n = 5k$}} & LD3 & \textbf{0.005 [0.005,0.006]} & \textbf{37.0 [37.0,37.0]} & \textbf{0.005 [0.005,0.005]} & \textbf{37.0 [37.0,37.0]} \\
    & PC & 0.738 [0.729,0.747] & 715.4 [708.08,722.72] & 0.757 [0.741,0.773] & 715.4 [708.08,722.72] \\
    & LDECC & 0.699 [0.688,0.709] & 640.6 [638.15,643.05] &  0.657 [0.596,0.718] &  604.7 [550.92,658.48] \\
    & LDP & 0.006 [0.006,0.006] & 47.2 [45.63,48.77] & 0.009 [0.004,0.015] & 52.4 [51.88,52.92] \\
    \midrule
    \parbox[t]{1cm}{\multirow{4}{*}{$n = 10k$}} & LD3 & \textbf{0.007 [0.007,0.007]} & \textbf{37.0 [37.0,37.0]} & \textbf{0.007 [0.006,0.007]} & \textbf{37.0 [37.0,37.0]} \\
    & PC & 1.388 [1.374,1.402] & 730.4 [726.86,733.94]  & 1.394 [1.384,1.405] & 730.4 [726.86,733.94] \\
    & LDECC & 1.199 [1.045,1.353] & 605.1 [528.14,682.06] & 1.041 [0.83,1.253] & 536.7 [428.01,645.39] \\
    & LDP & 0.008 [0.008,0.008] & 46.0 [46.0,46.0] & 0.008 [0.008,0.009] & 52.0 [52.0,52.0] \\
    \midrule
    \parbox[t]{1cm}{\multirow{4}{*}{$n = 20k$}}  & LD3 & \textbf{0.011 [0.011,0.011]} & \textbf{37.0 [37.0,37.0]}  & \textbf{0.011 [0.01,0.011]} & \textbf{37.0 [37.0,37.0]} \\
    & PC & 2.718 [2.689,2.746] & 733.6 [732.19,735.01] & 2.72 [2.676,2.763] & 733.6 [732.19,735.01] \\
    & LDECC & 2.5 [2.451,2.549] & 647.4 [644.05,650.75] & 2.472 [2.438,2.506]  &  644.4 [641.0,647.8] \\
    & LDP & 0.012 [0.012,0.013] & 46.6 [45.42,47.78] & 0.013 [0.013,0.013] & 52.0 [52.0,52.0] \\
    \bottomrule
    \end{tabular}
    \end{adjustbox}
    \caption{Total number of independence tests and runtimes (95\% confidence intervals in brackets).}
    \label{tab:sachs_time_tests}
\end{table}

\clearpage


\section{Real-World Causal Fairness Experiments}
\label{sec:fairness_experiments}

\paragraph{Computing Resources} All causal fairness experiments used an Apple MacBook (M2 Pro Chip; 12 CPU cores; 16G memory).

\subsection{COMPAS Recidivism Prediction}

\paragraph{Data Preprocessing} We used the data file \texttt{compas-scores-two-years.csv} from ProPublica,\footnote{\texttt{https://github.com/propublica/compas-analysis}} retaining the following columns for our analyses:

\begin{enumerate}
    \item \texttt{race}                (float64)
    \item \texttt{sex}                 (float64)
    \item \texttt{age\_cat}             (float64)
    \item \texttt{juv\_fel\_count}         (int64)
    \item \texttt{juv\_misd\_count}        (int64)
    \item \texttt{juv\_other\_count}       (int64)
    \item \texttt{priors\_count}          (int64)
    \item \texttt{c\_charge\_degree}     (float64)
    \item \texttt{two\_year\_recid}        (int64)
    \item \texttt{decile\_score}          (int64)
    \item \texttt{race\_binary}           (int64)
\end{enumerate}

Race was filtered to retain African American ($n = 3696$) and white individuals ($n = 2454$). For each exposure-outcome pair, $|\z| = 7$. When exposure was \texttt{race} and outcome was \texttt{decile\_score}, $\z$ contained all features other than \texttt{two\_year\_recid} (as this temporally proceeds \texttt{decile\_score}, violating \ref{assumption:y_no_desc}). When exposure was \texttt{race} and outcome was \texttt{two\_year\_recid}, $\z$ contained all features other than \texttt{decile\_score}.

\vspace{10mm}

\begin{table}[!h]
    \centering
    \begin{adjustbox}{max width=\textwidth}
    \begin{tabular}{l c c} 
    \toprule
        ~ & \textsc{White} ($n$ = 2454) & \textsc{African American} ($n$ = 3696) \\ 
        \midrule
        \textit{Age} & 37.73 (12.76) & 32.74 (10.86) \\
        \textit{Charge degree (felony)} & 0.6 (0.49) & 0.69 (0.46) \\
        \textit{Charge degree (misdemeanor)} & 0.4 (0.49) & 0.31 (0.46) \\
        \textit{General recidivism decile score} & 3.74 (2.6) & 5.37 (2.83) \\
        \textit{Juvenile felony count} & 0.03 (0.3) & 0.1 (0.51) \\
        \textit{Juvenile misdemeanor count} & 0.04 (0.28) & 0.14 (0.61) \\
        \textit{Juvenile other count} & 0.09 (0.45) & 0.14 (0.56) \\
        \textit{Priors count} & 2.59 (3.8) & 4.44 (5.58) \\
        \textit{Sex (female)} & 0.23 (0.42) & 0.18 (0.38) \\
        \textit{Sex (male)} & 0.77 (0.42) & 0.82 (0.38) \\
        \textit{Two-year recidivism (no)} & 0.61 (0.49) & 0.49 (0.5) \\
        \textit{Two-year recidivism (yes)} & 0.39 (0.49) & 0.51 (0.5) \\
        \bottomrule
    \end{tabular}
    \end{adjustbox}
    \caption{Summary statistics for African American versus white individuals in the COMPAS dataset after preprocessing. We report mean values (standard deviations). 
    }
    \label{tab:compas_summary_stats}
\end{table}

\begin{figure}[!h]
    \centering
    \fbox{\begin{tabular}{@{}c@{}}
    \includegraphics[width=0.45\linewidth]{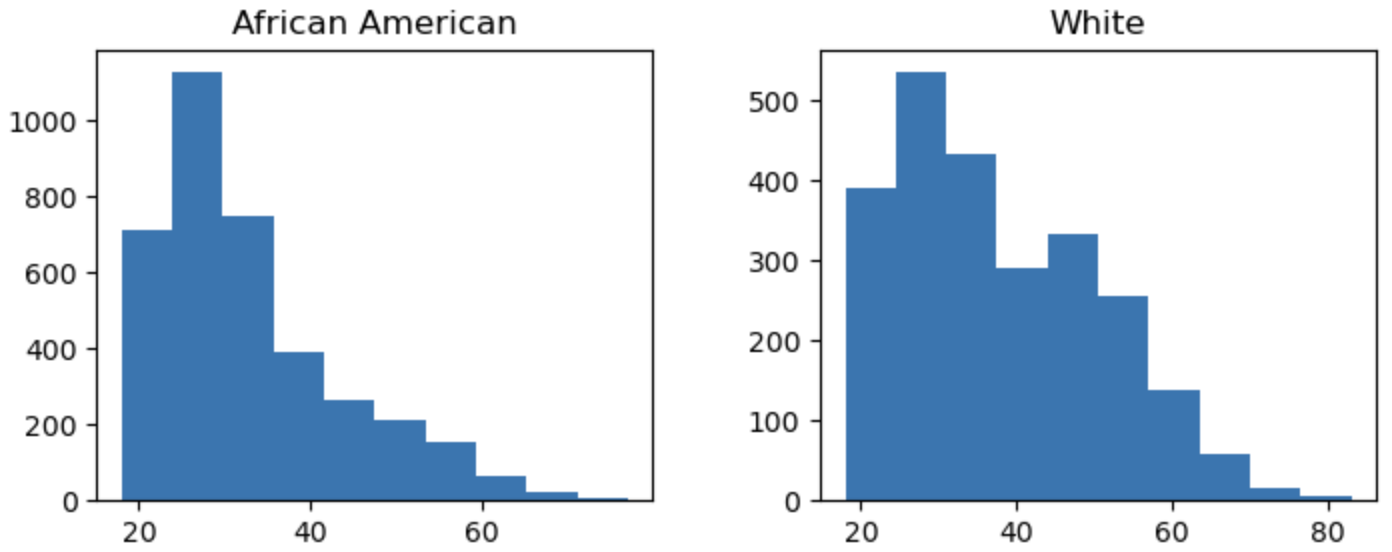} \\
    \footnotesize \textsc{Age}
    \end{tabular}}
    \fbox{\begin{tabular}{@{}c@{}}
    \includegraphics[width=0.45\linewidth]{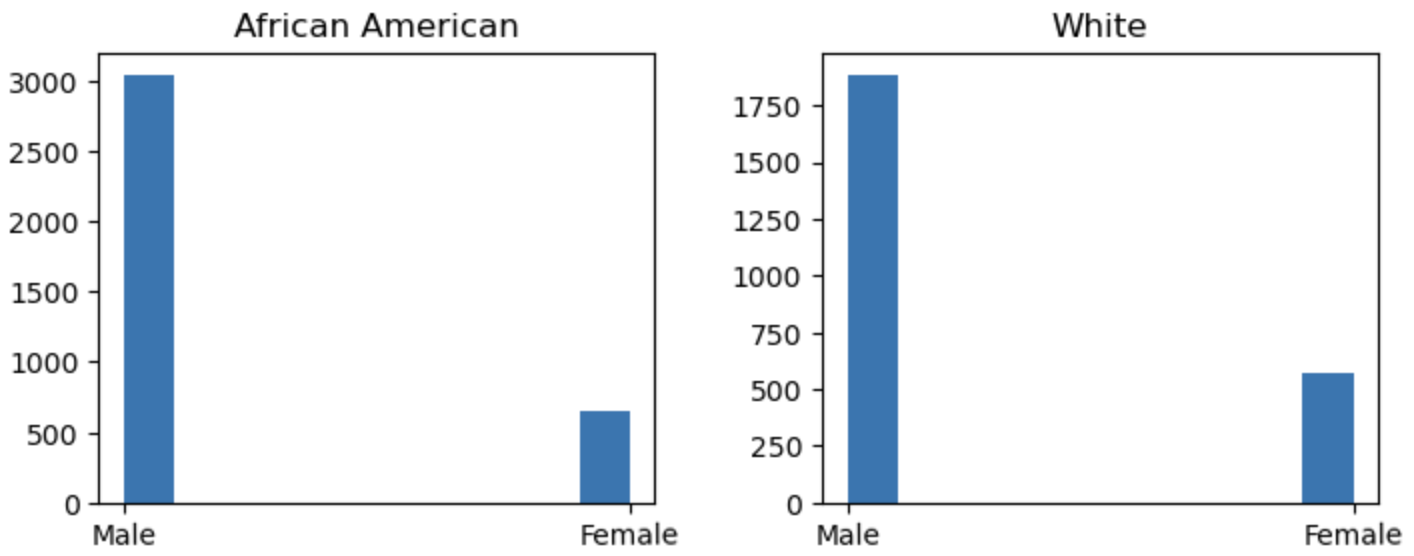} \\
    \footnotesize \textsc{Sex}
    \end{tabular}}
    \fbox{\begin{tabular}{@{}c@{}}
    \includegraphics[width=0.45\linewidth]{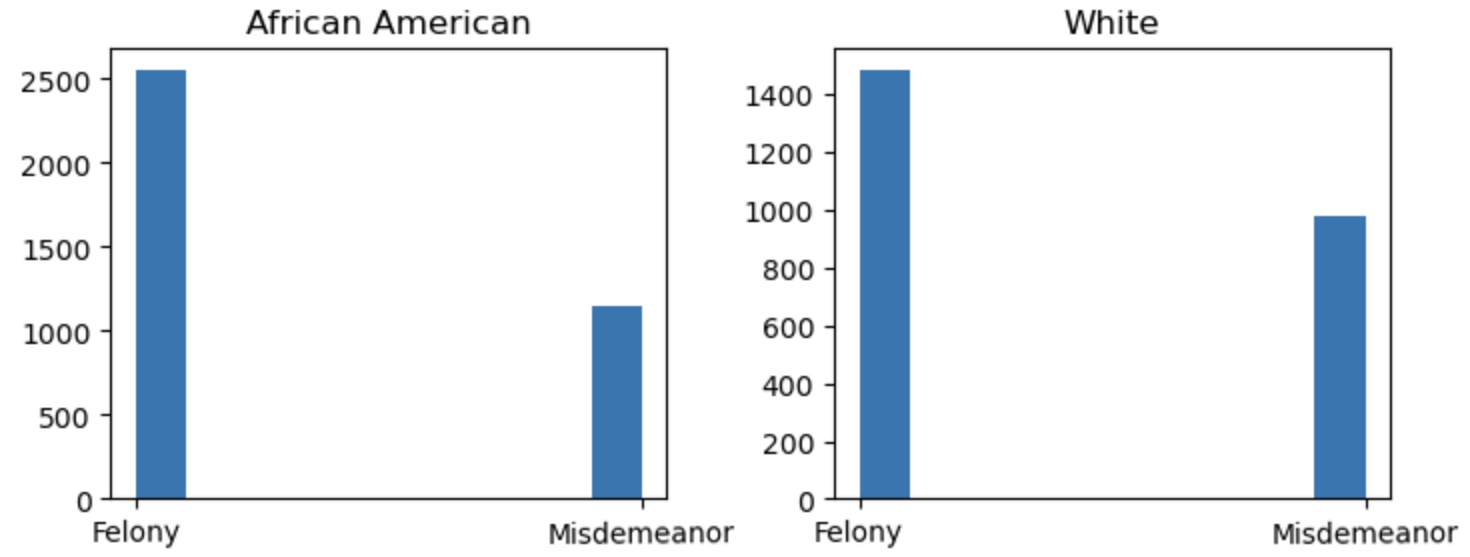} \\
    \footnotesize \textsc{Charge Degree}
    \end{tabular}}
    \fbox{\begin{tabular}{@{}c@{}}
    \includegraphics[width=0.45\linewidth]{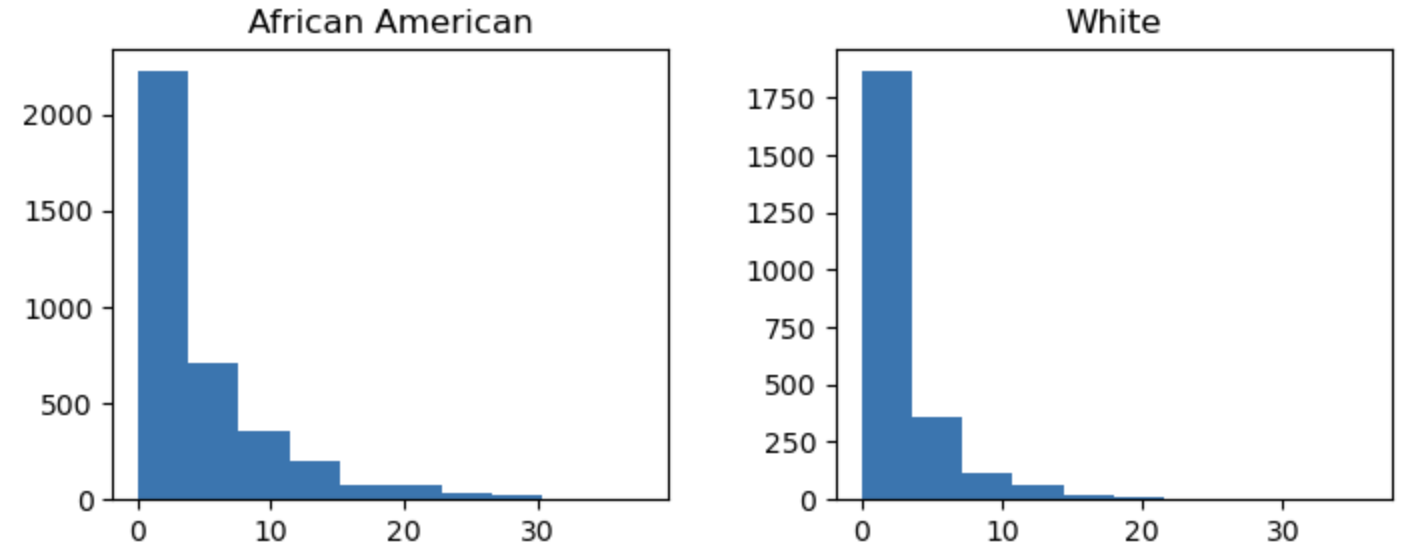} \\
    \footnotesize \textsc{Priors Count}
    \end{tabular}}
    \fbox{\begin{tabular}{@{}c@{}}
    \includegraphics[width=0.45\linewidth]{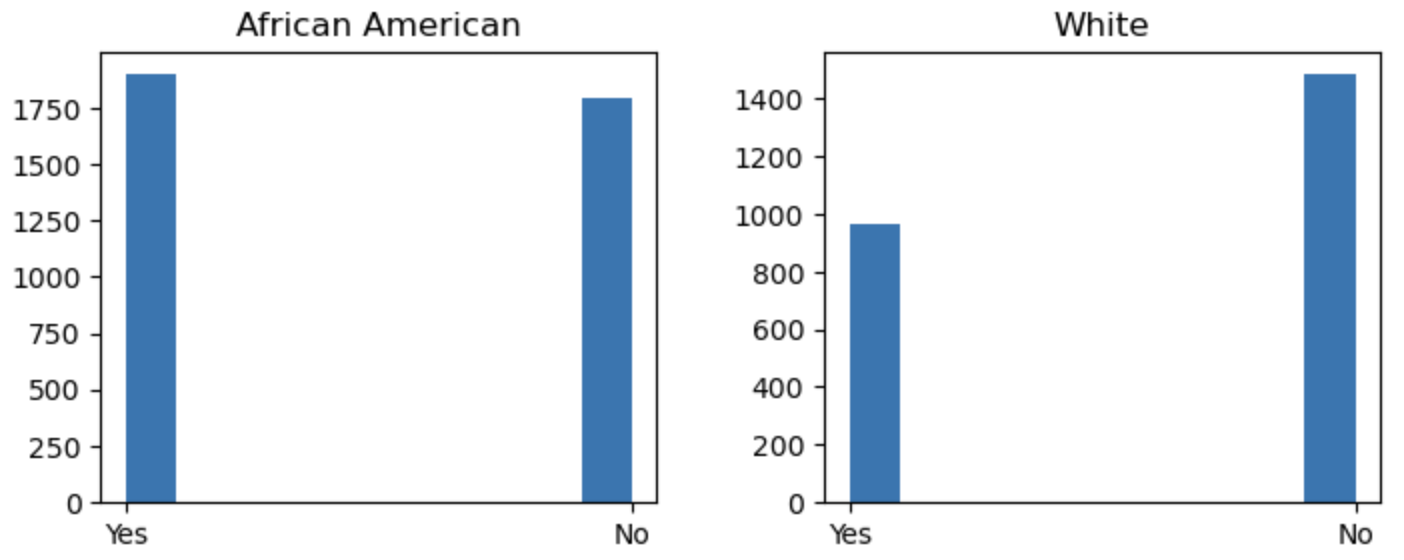} \\
    \footnotesize \textsc{Two-year Recidivism}
    \end{tabular}}
    \fbox{\begin{tabular}{@{}c@{}}
    \includegraphics[width=0.45\linewidth]{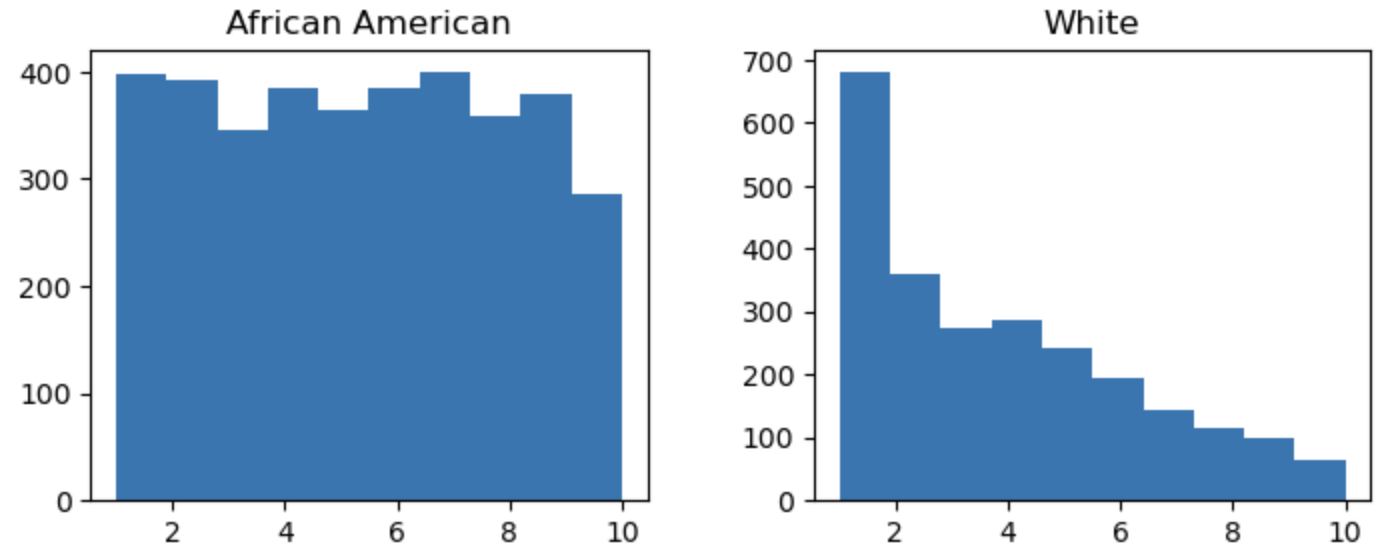} \\
    \footnotesize \textsc{General Recidivism Decile Score}
    \end{tabular}}
    \caption{Distributions for African American versus white individuals in the COMPAS dataset after preprocessing.}
    \label{fig:compas_distributions}
\end{figure}

\vspace{5mm}

\begin{figure}[!h]
    \centering
    \includegraphics[width=0.45\linewidth]{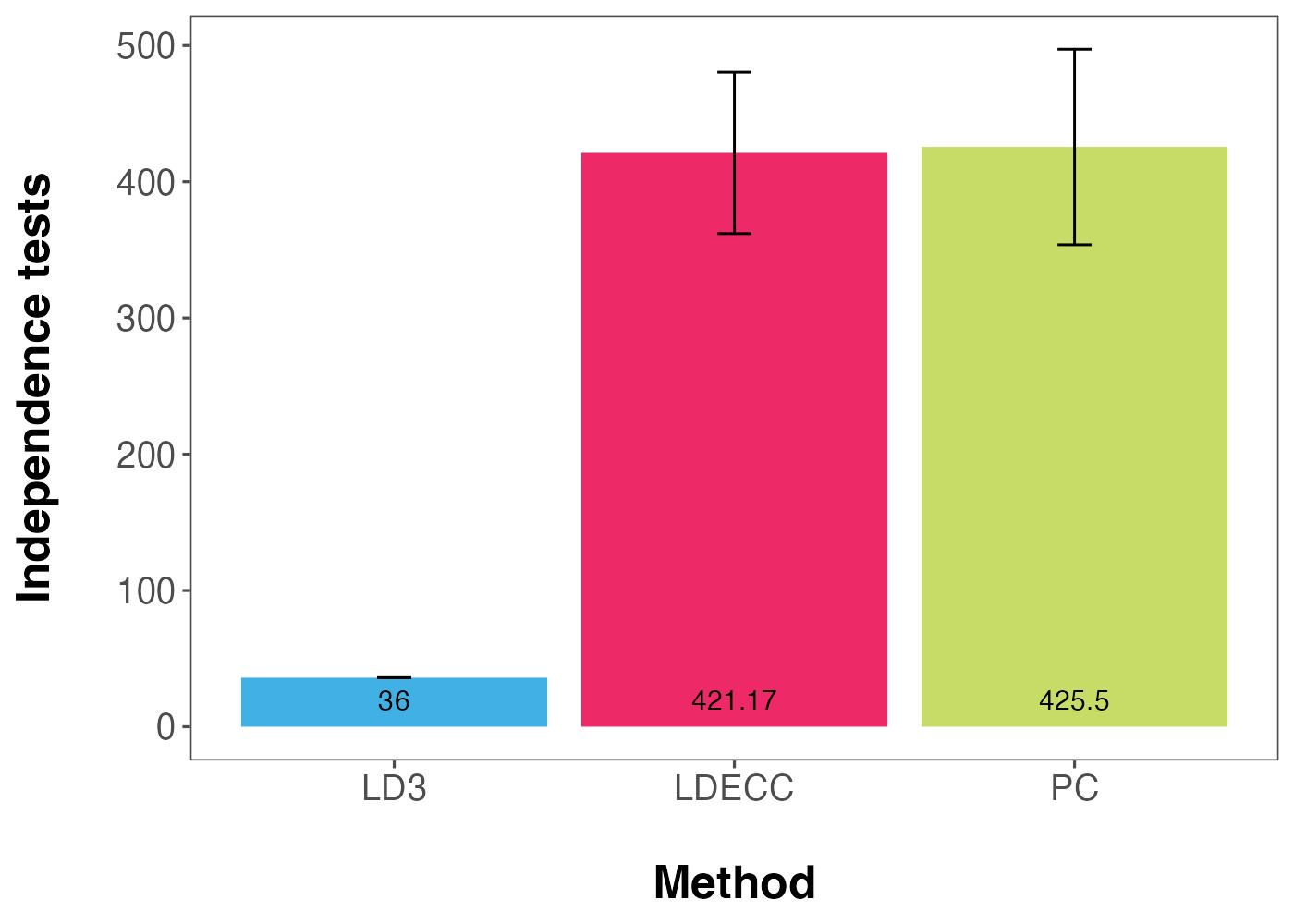}
    \hspace{3mm}
    \includegraphics[width=0.45\linewidth]{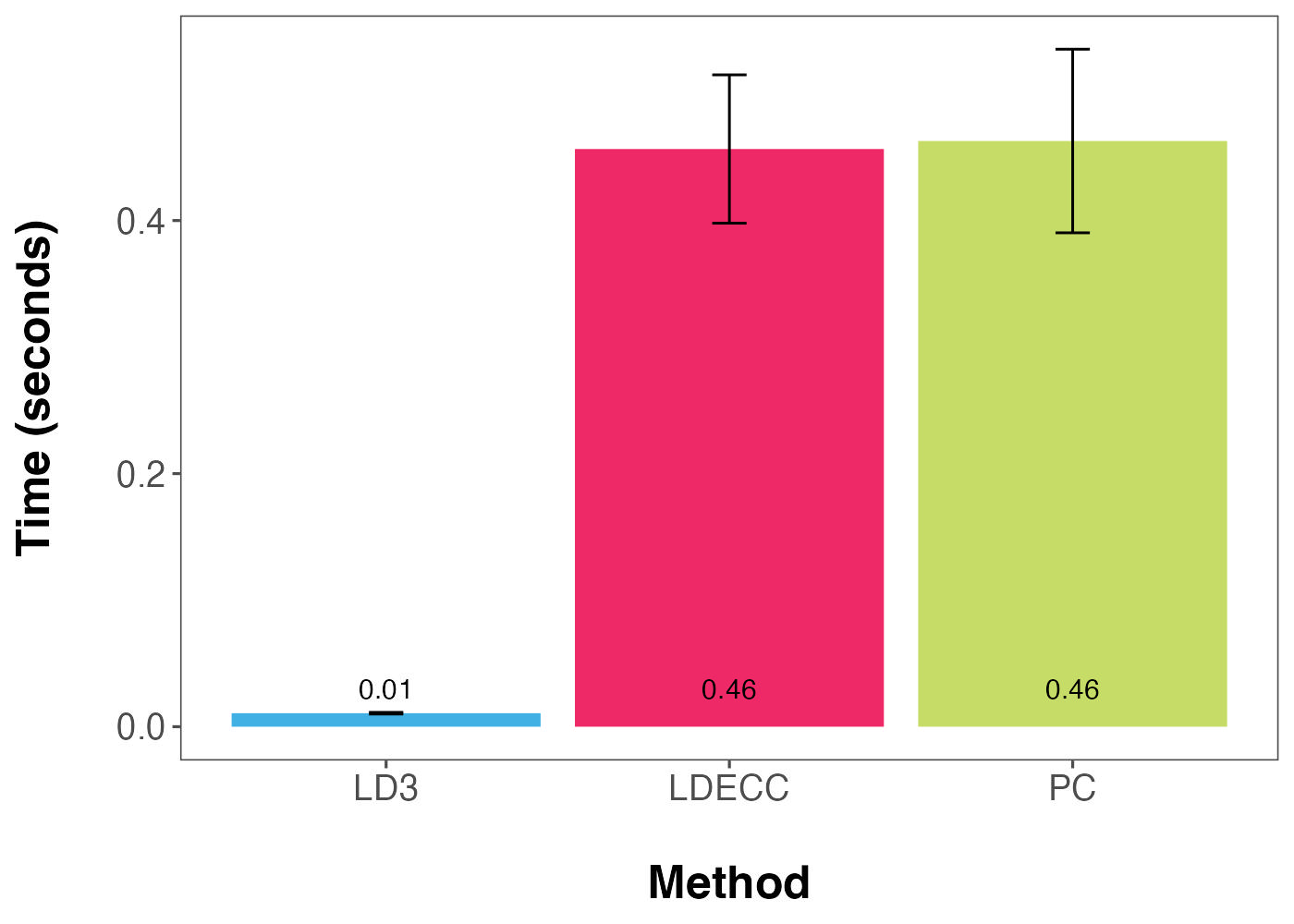}
    \caption{Average number of CI tests performed and average time (seconds) per method on COMPAS experiments.}
    \label{fig:compas_time_tests}
\end{figure}

\begin{figure}[!h]
    \centering
    \begin{tikzpicture}[scale=0.12]
\tikzstyle{every node}+=[inner sep=0pt]
\draw [CornflowerBlue,very thick,fill=CornflowerBlue,fill opacity=0.2] (27.9,-25.9) circle (3);
\draw (27.9,-25.9) node {$DS$};
\draw [black,very thick] (22.1,-15.9) circle (3);
\draw (22.1,-15.9) node {$CD$};
\draw [WildStrawberry,very thick,fill=WildStrawberry,fill opacity=0.2] (15.8,-25.9) circle (3);
\draw (15.8,-25.9) node {$R$};
\draw [black,very thick] (27.9,-36.3) circle (3);
\draw (27.9,-36.3) node {$A$};
\draw [Dandelion,very thick,fill=Dandelion,fill opacity=0.2] (40,-25.9) circle (3);
\draw (40,-25.9) node {$JF$};
\draw [Dandelion,very thick,fill=Dandelion,fill opacity=0.2] (34.7,-15.9) circle (3);
\draw (34.7,-15.9) node {$PC$};
\draw [black,very thick,-{Stealth[width=5pt,length=5pt]}] (18.8,-25.9) -- (24.9,-25.9);
\draw [black,very thick,-{Stealth[width=5pt,length=5pt]}] (27.9,-33.3) -- (27.9,-28.9);
\draw [black,very thick,-{Stealth[width=5pt,length=5pt]}] (37,-25.9) -- (30.9,-25.9);
\draw [black,very thick,-{Stealth[width=5pt,length=5pt]}] (33.01,-18.38) -- (29.59,-23.42);
\draw [black,very thick,-{Stealth[width=5pt,length=5pt]}] (23.61,-18.5) -- (26.39,-23.3);
\node [below=7mm, align=flush center,text width=2.5cm] at (27.9,-36.3) {\textbf{(A)} $\alpha = 0.005$ $SDC = 1$};
\node [below=15mm, align=flush center,text width=5cm] at (27.9,-36.3) {$WCDE = 0.694 \; [0.548,0.839]$ {\color{NavyBlue}$p = 0.000$}};
\end{tikzpicture} \hspace{5mm}
    \begin{tikzpicture}[scale=0.12]
\tikzstyle{every node}+=[inner sep=0pt]
\draw [CornflowerBlue,very thick,fill=CornflowerBlue,fill opacity=0.2] (27.9,-25.9) circle (3);
\draw (27.9,-25.9) node {$DS$};
\draw [black,very thick] (22.1,-15.9) circle (3);
\draw (22.1,-15.9) node {$CD$};
\draw [WildStrawberry,very thick,fill=WildStrawberry,fill opacity=0.2] (15.8,-25.9) circle (3);
\draw (15.8,-25.9) node {$R$};
\draw [black,very thick] (27.9,-36.3) circle (3);
\draw (27.9,-36.3) node {$A$};
\draw [Dandelion,very thick,fill=Dandelion,fill opacity=0.2] (40,-25.9) circle (3);
\draw (40,-25.9) node {$JF$};
\draw [Dandelion,very thick,fill=Dandelion,fill opacity=0.2] (34.7,-15.9) circle (3);
\draw (34.7,-15.9) node {$PC$};
\draw [black,very thick,-{Stealth[width=5pt,length=5pt]}] (18.8,-25.9) -- (24.9,-25.9);
\draw [black,very thick,-{Stealth[width=5pt,length=5pt]}] (27.9,-33.3) -- (27.9,-28.9);
\draw [black,very thick,-{Stealth[width=5pt,length=5pt]}] (37,-25.9) -- (30.9,-25.9);
\draw [black,very thick,-{Stealth[width=5pt,length=5pt]}] (33.01,-18.38) -- (29.59,-23.42);
\draw [black,very thick,-{Stealth[width=5pt,length=5pt]}] (23.61,-18.5) -- (26.39,-23.3);
\node [below=7mm, align=flush center,text width=2cm] at (27.9,-36.3) {\textbf{(B)} $\alpha = 0.01$ $SDC = 1$};
\node [below=15mm, align=flush center,text width=5cm] at (27.9,-36.3) {$WCDE = 0.695 \; [0.55,0.84]$ {\color{NavyBlue}$p = 0.000$}};
\end{tikzpicture} \hspace{5mm}
    \begin{tikzpicture}[scale=0.12]
\tikzstyle{every node}+=[inner sep=0pt]
\draw [CornflowerBlue,very thick,fill=CornflowerBlue,fill opacity=0.2] (27.9,-25.9) circle (3);
\draw (27.9,-25.9) node {$DS$};
\draw [black,very thick] (22.1,-15.9) circle (3);
\draw (22.1,-15.9) node {$CD$};
\draw [WildStrawberry,very thick,fill=WildStrawberry,fill opacity=0.2] (15.8,-25.9) circle (3);
\draw (15.8,-25.9) node {$R$};
\draw [black,very thick] (36.3,-33.8) circle (3);
\draw (36.3,-33.8) node {$A$};
\draw [Dandelion,very thick,fill=Dandelion,fill opacity=0.2] (40,-25.9) circle (3);
\draw (40,-25.9) node {$JF$};
\draw [Dandelion,very thick,fill=Dandelion,fill opacity=0.2] (34.7,-15.9) circle (3);
\draw (34.7,-15.9) node {$PC$};
\draw [black,very thick] (20.3,-33.8) circle (3);
\draw (20.3,-33.8) node {$S$};
\draw [Dandelion,very thick,fill=Dandelion,fill opacity=0.2] (27.9,-37.9) circle (3);
\draw (27.9,-37.9) node {$JM$};
\draw [black,very thick,-{Stealth[width=5pt,length=5pt]}] (18.8,-25.9) -- (24.9,-25.9);
\draw [black,very thick,-{Stealth[width=5pt,length=5pt]}] (34.11,-31.74) -- (30.09,-27.96);
\draw [black,very thick,-{Stealth[width=5pt,length=5pt]}] (37,-25.9) -- (30.9,-25.9);
\draw [black,very thick,-{Stealth[width=5pt,length=5pt]}] (33.01,-18.38) -- (29.59,-23.42);
\draw [black,very thick,-{Stealth[width=5pt,length=5pt]}] (23.61,-18.5) -- (26.39,-23.3);
\draw [black,very thick,-{Stealth[width=5pt,length=5pt]}] (22.38,-31.64) -- (25.82,-28.06);
\draw [black,very thick,-{Stealth[width=5pt,length=5pt]}] (27.9,-34.9) -- (27.9,-28.9);
\node [below=7mm, align=flush center,text width=2cm] at (27.9,-37.9){\textbf{(C)} $\alpha = 0.05$ $SDC = 1$};
\node [below=16mm, align=flush center,text width=5cm] at (27.9,-36.3) {$WCDE = 0.657 \; [0.51,0.804]$ {\color{NavyBlue}$p = 0.000$}};
\end{tikzpicture} \\
    \vspace{10mm}
    \begin{tikzpicture}[scale=0.12]
\tikzstyle{every node}+=[inner sep=0pt]
\draw [CornflowerBlue,very thick,fill=CornflowerBlue,fill opacity=0.2] (27.9,-25.9) circle (3);
\draw (27.9,-25.9) node {$TY$};
\draw [black,very thick] (20.2,-17.5) circle (3);
\draw (20.2,-17.5) node {$CD$};
\draw [black,very thick] (36.3,-33.8) circle (3);
\draw (36.3,-33.8) node {$A$};
\draw [black,very thick] (36.3,-17.5) circle (3);
\draw (36.3,-17.5) node {$PC$};
\draw [black,very thick] (20.2,-33.8) circle (3);
\draw (20.2,-33.8) node {$S$};
\draw [black,very thick,-{Stealth[width=5pt,length=5pt]}] (34.11,-31.74) -- (30.09,-27.96);
\draw [black,very thick,-{Stealth[width=5pt,length=5pt]}] (34.18,-19.62) -- (30.02,-23.78);
\draw [black,very thick,-{Stealth[width=5pt,length=5pt]}] (22.23,-19.71) -- (25.87,-23.69);
\draw [black,very thick,-{Stealth[width=5pt,length=5pt]}] (22.29,-31.65) -- (25.81,-28.05);
\node [below=17mm, align=flush center,text width=2.5cm] at (27.9,-25.9) {\textbf{(D)} $\alpha = 0.005$ $SDC = 0$};
\node [below=25mm, align=flush center,text width=5cm] at (27.9,-25.9) {$WCDE = 0.003 \; [-0.028,0.033]$ {\color{Maroon}$p = 0.87$}};
\end{tikzpicture} \hspace{5mm}
    \begin{tikzpicture}[scale=0.12]
\tikzstyle{every node}+=[inner sep=0pt]
\draw [CornflowerBlue,very thick,fill=CornflowerBlue,fill opacity=0.2] (27.9,-25.9) circle (3);
\draw (27.9,-25.9) node {$TY$};
\draw [black,very thick] (20.2,-17.5) circle (3);
\draw (20.2,-17.5) node {$CD$};
\draw [black,very thick] (36.3,-33.8) circle (3);
\draw (36.3,-33.8) node {$A$};
\draw [black,very thick] (36.3,-17.5) circle (3);
\draw (36.3,-17.5) node {$PC$};
\draw [black,very thick] (20.2,-33.8) circle (3);
\draw (20.2,-33.8) node {$S$};
\draw [WildStrawberry,very thick,fill=WildStrawberry,fill opacity=0.2] (15.2,-26.4) circle (3);
\draw (15.2,-26.4) node {$R$};
\draw [black,very thick] (40.8,-26.4) circle (3);
\draw (40.8,-26.4) node {$JM$};
\draw [black,very thick,-{Stealth[width=5pt,length=5pt]}] (34.11,-31.74) -- (30.09,-27.96);
\draw [black,very thick,-{Stealth[width=5pt,length=5pt]}] (34.18,-19.62) -- (30.02,-23.78);
\draw [black,very thick,-{Stealth[width=5pt,length=5pt]}] (22.23,-19.71) -- (25.87,-23.69);
\draw [black,very thick,-{Stealth[width=5pt,length=5pt]}] (22.29,-31.65) -- (25.81,-28.05);
\draw [black,very thick,-{Stealth[width=5pt,length=5pt]}] (18.2,-26.28) -- (24.9,-26.02);
\draw [black,very thick,-{Stealth[width=5pt,length=5pt]}] (37.8,-26.28) -- (30.9,-26.02);
\node [below=17mm, align=flush center,text width=2cm] at (27.9,-25.9) {\textbf{(E)} $\alpha = 0.01$ $SDC = 1$};
\node [below=25mm, align=flush center,text width=5cm] at (27.9,-25.9) {$WCDE = 0.007 \; [-0.024,0.037]$ {\color{Maroon}$p = 0.675$}};
\end{tikzpicture} \hspace{5mm}
    \begin{tikzpicture}[scale=0.12]
\tikzstyle{every node}+=[inner sep=0pt]
\draw [CornflowerBlue,very thick,fill=CornflowerBlue,fill opacity=0.2] (27.9,-25.9) circle (3);
\draw (27.9,-25.9) node {$TY$};
\draw [black,very thick] (20.2,-17.5) circle (3);
\draw (20.2,-17.5) node {$CD$};
\draw [black,very thick] (36.3,-33.8) circle (3);
\draw (36.3,-33.8) node {$A$};
\draw [black,very thick] (36.3,-17.5) circle (3);
\draw (36.3,-17.5) node {$PC$};
\draw [black,very thick] (20.2,-33.8) circle (3);
\draw (20.2,-33.8) node {$S$};
\draw [WildStrawberry,very thick,fill=WildStrawberry,fill opacity=0.2] (15.2,-26.4) circle (3);
\draw (15.2,-26.4) node {$R$};
\draw [black,very thick] (40.8,-26.4) circle (3);
\draw (40.8,-26.4) node {$JF$};
\draw [black,very thick] (27.9,-39.4) circle (3);
\draw (27.9,-39.4) node {$JM$};
\draw [black,very thick,-{Stealth[width=5pt,length=5pt]}] (34.11,-31.74) -- (30.09,-27.96);
\draw [black,very thick,-{Stealth[width=5pt,length=5pt]}] (34.18,-19.62) -- (30.02,-23.78);
\draw [black,very thick,-{Stealth[width=5pt,length=5pt]}] (22.23,-19.71) -- (25.87,-23.69);
\draw [black,very thick,-{Stealth[width=5pt,length=5pt]}] (22.29,-31.65) -- (25.81,-28.05);
\draw [black,very thick,-{Stealth[width=5pt,length=5pt]}] (18.2,-26.28) -- (24.9,-26.02);
\draw [black,very thick,-{Stealth[width=5pt,length=5pt]}] (37.8,-26.28) -- (30.9,-26.02);
\draw [black,very thick,-{Stealth[width=5pt,length=5pt]}] (27.9,-36.5) -- (27.9,-28.9);
\node [below=7mm, align=flush center,text width=2cm] at (27.9,-39.4) {\textbf{(F)} $\alpha = 0.05$ $SDC = 1$};
\node [below=15mm, align=flush center,text width=5cm] at (27.9,-39.4)  {$WCDE = 0.005 \; [-0.026,0.035]$ {\color{Maroon}$p = 0.766$}};
\end{tikzpicture}
    \caption{LD3 results on the COMPAS dataset at different significance levels ($\alpha = [0.005, 0.01, 0.05]$). DAGs represent the predicted parent set for the outcome only, with all other relations abstracted away. (\textbf{A–C}) Exposure race (\textit{R}; {\color{WildStrawberry}red}) and outcome general recidivism decile score (\textit{DS}; {\color{CornflowerBlue}blue}). Known parents of outcome \textit{DS} are in {\color{Dandelion}yellow}. (\textbf{D–F}) Exposure race and outcome two-year recidivism (\textit{TY}; {\color{CornflowerBlue}blue}). \textit{A} = age; \textit{CD} = charge degree; \textit{DS} = decile score; \textit{JF} = juvenile felony count; \textit{JM} = juvenile misdemeanor count; \textit{PC} = priors count; \textit{R} = race; \textit{S} = sex; \textit{TY} = two-year recidivism. Weighted CDE ($WCDE$) is reported with $p$-value ($p$) and 95\% confidence intervals in brackets.}
    \label{fig:compas_ld3}
\end{figure}

\clearpage
\begin{figure}[!h]
    \centering
    \begin{tikzpicture}[scale=0.12]
\tikzstyle{every node}+=[inner sep=0pt]
\draw [CornflowerBlue,very thick,fill=CornflowerBlue,fill opacity=0.2] (27.9,-25.9) circle (3);
\draw (27.9,-25.9) node {$DS$};
\draw [black,very thick] (27.9,-36.7) circle (3);
\draw (27.9,-36.7) node {$A$};
\draw [Dandelion,very thick,fill=Dandelion,fill opacity=0.2] (40.3,-26.4) circle (3);
\draw (40.3,-26.4) node {$PC$};
\draw [WildStrawberry,very thick,fill=WildStrawberry,fill opacity=0.2] (15.2,-26.4) circle (3);
\draw (15.2,-26.4) node {$R$};
\draw [black,very thick,-{Stealth[width=5pt,length=5pt]}] (27.9,-33.7) -- (27.9,-28.9);
\draw [black,very thick,-{Stealth[width=5pt,length=5pt]}] (37.3,-26.28) -- (30.9,-26.02);
\draw [black,very thick,-{Stealth[width=5pt,length=5pt]}] (18.2,-26.28) -- (24.9,-26.02);
\node [below=7mm, align=flush center,text width=2.5cm] at (27.9,-36.7){\textbf{(A)} $\alpha = 0.005$ $SDC = 1$};
\node [below=15mm, align=flush center,text width=5cm] at (27.9,-36.7){$WCDE = 0.709 \; [0.564,0.853]$ {\color{NavyBlue}$p = 0.000$}};
\end{tikzpicture} \hspace{5mm}
    \begin{tikzpicture}[scale=0.12]
\tikzstyle{every node}+=[inner sep=0pt]
\draw [CornflowerBlue,very thick,fill=CornflowerBlue,fill opacity=0.2] (27.9,-25.9) circle (3);
\draw (27.9,-25.9) node {$DS$};
\draw [black,very thick] (27.9,-36.7) circle (3);
\draw (27.9,-36.7) node {$A$};
\draw [Dandelion,very thick,fill=Dandelion,fill opacity=0.2] (40.3,-26.4) circle (3);
\draw (40.3,-26.4) node {$PC$};
\draw [WildStrawberry,very thick,fill=WildStrawberry,fill opacity=0.2] (15.2,-26.4) circle (3);
\draw (15.2,-26.4) node {$R$};
\draw [black,very thick,-{Stealth[width=5pt,length=5pt]}] (27.9,-33.7) -- (27.9,-28.9);
\draw [black,very thick,-{Stealth[width=5pt,length=5pt]}] (37.3,-26.28) -- (30.9,-26.02);
\draw [black,very thick,-{Stealth[width=5pt,length=5pt]}] (18.2,-26.28) -- (24.9,-26.02);
\node [below=7mm, align=flush center,text width=2.5cm] at (27.9,-36.7){\textbf{(B)} $\alpha = 0.01$ $SDC = 1$};
\node [below=15mm, align=flush center,text width=5cm] at (27.9,-36.7){$WCDE = 0.71 \; [0.565,0.855]$ {\color{NavyBlue}$p = 0.000$}};
\end{tikzpicture} \hspace{5mm}
    \begin{tikzpicture}[scale=0.12]
\tikzstyle{every node}+=[inner sep=0pt]
\draw [CornflowerBlue,very thick,fill=CornflowerBlue,fill opacity=0.2] (27.9,-25.9) circle (3);
\draw (27.9,-25.9) node {$DS$};
\draw [black,very thick] (27.9,-36.7) circle (3);
\draw (27.9,-36.7) node {$A$};
\draw [Dandelion,very thick,fill=Dandelion,fill opacity=0.2] (40.3,-26.4) circle (3);
\draw (40.3,-26.4) node {$PC$};
\draw [WildStrawberry,very thick,fill=WildStrawberry,fill opacity=0.2] (15.2,-26.4) circle (3);
\draw (15.2,-26.4) node {$R$};
\draw [black,very thick,-{Stealth[width=5pt,length=5pt]}] (27.9,-33.7) -- (27.9,-28.9);
\draw [black,very thick,-{Stealth[width=5pt,length=5pt]}] (37.3,-26.28) -- (30.9,-26.02);
\draw [black,very thick,-{Stealth[width=5pt,length=5pt]}] (18.2,-26.28) -- (24.9,-26.02);
\node [below=7mm, align=flush center,text width=2.5cm] at (27.9,-36.7){\textbf{(C)} $\alpha = 0.05$ $SDC = 1$};
\node [below=15mm, align=flush center,text width=5cm] at (27.9,-36.7){$WCDE = 0.71 \; [0.565,0.855]$ {\color{NavyBlue}$p = 0.000$}};
\end{tikzpicture} \\
    \vspace{5mm}
    \begin{tikzpicture}[scale=0.12]
\tikzstyle{every node}+=[inner sep=0pt]
\draw [CornflowerBlue,very thick,fill=CornflowerBlue,fill opacity=0.2] (27.9,-25.9) circle (3);
\draw (27.9,-25.9) node {$TY$};
\draw [black,very thick] (40.3,-26.4) circle (3);
\draw (40.3,-26.4) node {$PC$};
\draw [WildStrawberry,very thick,fill=WildStrawberry,fill opacity=0.2] (15.2,-26.4) circle (3);
\draw (15.2,-26.4) node {$R$};
\draw [black,very thick,-{Stealth[width=5pt,length=5pt]}] (37.3,-26.28) -- (30.9,-26.02);
\draw [black,very thick,-{Stealth[width=5pt,length=5pt]}] (18.2,-26.28) -- (24.9,-26.02);
\node [below=7mm, align=flush center,text width=2.5cm] at (27.9,-25.9){\textbf{(D)} $\alpha = 0.005$ $SDC = 1$};
\node [below=15mm, align=flush center,text width=5cm] at (27.9,-25.9){$WCDE = 0.058 \; [0.029,0.088]$ {\color{NavyBlue}$p = 0.000$}};
\end{tikzpicture} \hspace{5mm}
    \begin{tikzpicture}[scale=0.12]
\tikzstyle{every node}+=[inner sep=0pt]
\draw [CornflowerBlue,very thick,fill=CornflowerBlue,fill opacity=0.2] (27.9,-25.9) circle (3);
\draw (27.9,-25.9) node {$TY$};
\draw [black,very thick] (40.3,-26.4) circle (3);
\draw (40.3,-26.4) node {$PC$};
\draw [WildStrawberry,very thick,fill=WildStrawberry,fill opacity=0.2] (15.2,-26.4) circle (3);
\draw (15.2,-26.4) node {$R$};
\draw [black,very thick,-{Stealth[width=5pt,length=5pt]}] (37.3,-26.28) -- (30.9,-26.02);
\draw [black,very thick,-{Stealth[width=5pt,length=5pt]}] (18.2,-26.28) -- (24.9,-26.02);
\node [below=7mm, align=flush center,text width=2.5cm] at (27.9,-25.9){\textbf{(E)} $\alpha = 0.01$ $SDC = 1$};
\node [below=15mm, align=flush center,text width=5cm] at (27.9,-25.9){$WCDE = 0.059 \; [0.029,0.089]$ {\color{NavyBlue}$p = 0.000$}};
\end{tikzpicture} \hspace{5mm}
    \begin{tikzpicture}[scale=0.12]
\tikzstyle{every node}+=[inner sep=0pt]
\draw [CornflowerBlue,very thick,fill=CornflowerBlue,fill opacity=0.2] (27.9,-25.9) circle (3);
\draw (27.9,-25.9) node {$TY$};
\draw [black,very thick] (40.3,-26.4) circle (3);
\draw (40.3,-26.4) node {$PC$};
\draw [WildStrawberry,very thick,fill=WildStrawberry,fill opacity=0.2] (15.2,-26.4) circle (3);
\draw (15.2,-26.4) node {$R$};
\draw [black,very thick] (21.7,-35.4) circle (3);
\draw (21.7,-35.4) node {$A$};
\draw [black,very thick] (34.2,-35.4) circle (3);
\draw (34.2,-35.4) node {$S$};
\draw [black,very thick,-{Stealth[width=5pt,length=5pt]}] (37.3,-26.28) -- (30.9,-26.02);
\draw [black,very thick,-{Stealth[width=5pt,length=5pt]}] (18.2,-26.28) -- (24.9,-26.02);
\draw [black,very thick,-{Stealth[width=5pt,length=5pt]}] (23.34,-32.89) -- (26.26,-28.41);
\draw [black,very thick,-{Stealth[width=5pt,length=5pt]}] (32.54,-32.9) -- (29.56,-28.4);
\node [below=19mm, align=flush center,text width=2.5cm] at (27.9,-25.9){\textbf{(F)} $\alpha = 0.05$ $SDC = 1$};
\node [below=27mm, align=flush center,text width=5cm] at (27.9,-25.9){$WCDE = 0.004 \; [-0.026,0.034]$ {\color{Maroon}$p = 0.797$}};
\end{tikzpicture}
    \caption{PC results on the COMPAS dataset at different significance levels ($\alpha = [0.005, 0.01, 0.05]$). DAGs represent the predicted parent set for the outcome only, with all other relations abstracted away. (\textbf{A–C}) Exposure race (\textit{R}; {\color{WildStrawberry}red}) and outcome general recidivism decile score (\textit{DS}; {\color{CornflowerBlue}blue}). Known parents of outcome \textit{DS} are in {\color{Dandelion}yellow}. (\textbf{D–F}) Exposure race and outcome two-year recidivism (\textit{TY}; {\color{CornflowerBlue}blue}). \textit{A} = age; \textit{DS} = decile score; \textit{PC} = priors count; \textit{R} = race; \textit{S} = sex; \textit{TY} = two-year recidivism. Weighted CDE ($WCDE$) is reported with $p$-value ($p$) and 95\% confidence intervals in brackets.}
    \label{fig:compas_pc}
\end{figure}
\vspace{15mm}
\begin{figure}[!h]
    \centering
    \begin{tikzpicture}[scale=0.12]
\tikzstyle{every node}+=[inner sep=0pt]
\draw [CornflowerBlue,very thick,fill=CornflowerBlue,fill opacity=0.2] (27.9,-25.9) circle (3);
\draw (27.9,-25.9) node {$DS$};
\draw [black,very thick] (27.9,-36.7) circle (3);
\draw (27.9,-36.7) node {$A$};
\draw [Dandelion,very thick,fill=Dandelion,fill opacity=0.2] (40.3,-26.4) circle (3);
\draw (40.3,-26.4) node {$PC$};
\draw [WildStrawberry,very thick,fill=WildStrawberry,fill opacity=0.2] (15.2,-26.4) circle (3);
\draw (15.2,-26.4) node {$R$};
\draw [black,very thick,-{Stealth[width=5pt,length=5pt]}] (27.9,-33.7) -- (27.9,-28.9);
\draw [black,very thick,-{Stealth[width=5pt,length=5pt]}] (37.3,-26.28) -- (30.9,-26.02);
\draw [black,very thick,-{Stealth[width=5pt,length=5pt]}] (18.2,-26.28) -- (24.9,-26.02);
\node [below=7mm, align=flush center,text width=2.5cm] at (27.9,-36.7){\textbf{(A)} $\alpha = 0.005$ $SDC = 1$};
\node [below=15mm, align=flush center,text width=5cm] at (27.9,-36.7){$WCDE = 0.711 \; [0.566,0.856]$ {\color{NavyBlue}$p = 0.000$}};
\end{tikzpicture} \hspace{5mm}
    \begin{tikzpicture}[scale=0.12]
\tikzstyle{every node}+=[inner sep=0pt]
\draw [CornflowerBlue,very thick,fill=CornflowerBlue,fill opacity=0.2] (27.9,-25.9) circle (3);
\draw (27.9,-25.9) node {$DS$};
\draw [black,very thick] (27.9,-36.7) circle (3);
\draw (27.9,-36.7) node {$A$};
\draw [Dandelion,very thick,fill=Dandelion,fill opacity=0.2] (40.3,-26.4) circle (3);
\draw (40.3,-26.4) node {$PC$};
\draw [WildStrawberry,very thick,fill=WildStrawberry,fill opacity=0.2] (15.2,-26.4) circle (3);
\draw (15.2,-26.4) node {$R$};
\draw [black,very thick,-{Stealth[width=5pt,length=5pt]}] (27.9,-33.7) -- (27.9,-28.9);
\draw [black,very thick,-{Stealth[width=5pt,length=5pt]}] (37.3,-26.28) -- (30.9,-26.02);
\draw [black,very thick,-{Stealth[width=5pt,length=5pt]}] (18.2,-26.28) -- (24.9,-26.02);
\node [below=7mm, align=flush center,text width=2.5cm] at (27.9,-36.7){\textbf{(B)} $\alpha = 0.01$ $SDC = 1$};
\node [below=15mm, align=flush center,text width=5cm] at (27.9,-36.7){$WCDE = 0.709 \; [0.564,0.853]$ {\color{NavyBlue}$p = 0.000$}};
\end{tikzpicture} \hspace{5mm}
    \begin{tikzpicture}[scale=0.12]
\tikzstyle{every node}+=[inner sep=0pt]
\draw [CornflowerBlue,very thick,fill=CornflowerBlue,fill opacity=0.2] (27.9,-25.9) circle (3);
\draw (27.9,-25.9) node {$DS$};
\draw [black,very thick] (27.9,-36.7) circle (3);
\draw (27.9,-36.7) node {$CD$};
\draw [WildStrawberry,very thick,fill=WildStrawberry,fill opacity=0.2] (15.2,-26.4) circle (3);
\draw (15.2,-26.4) node {$R$};
\draw [black,very thick,-{Stealth[width=5pt,length=5pt]}] (27.9,-33.7) -- (27.9,-28.9);
\draw [black,very thick,-{Stealth[width=5pt,length=5pt]}] (18.2,-26.28) -- (24.9,-26.02);
\node [below=7mm, align=flush center,text width=2.5cm] at (27.9,-36.7){\textbf{(C)} $\alpha = 0.05$ $SDC = 1$};
\node [below=15mm, align=flush center,text width=5cm] at (27.9,-36.7){$WCDE = 1.559 \; [1.396,1.723]$ {\color{Cerulean}$p = 0.000$}};
\end{tikzpicture} \\
    \vspace{5mm}
    \begin{tikzpicture}[scale=0.12]
\tikzstyle{every node}+=[inner sep=0pt]
\draw [CornflowerBlue,very thick,fill=CornflowerBlue,fill opacity=0.2] (27.9,-25.9) circle (3);
\draw (27.9,-25.9) node {$TY$};
\draw [black,very thick] (27.9,-36.7) circle (3);
\draw (27.9,-36.7) node {$A$};
\draw [black,very thick] (40.3,-26.4) circle (3);
\draw (40.3,-26.4) node {$PC$};
\draw [black,very thick,-{Stealth[width=5pt,length=5pt]}] (27.9,-33.7) -- (27.9,-28.9);
\draw [black,very thick,-{Stealth[width=5pt,length=5pt]}] (37.3,-26.28) -- (30.9,-26.02);
\node [below=7mm, align=flush center,text width=2.5cm] at (27.9,-36.7){\textbf{(D)} $\alpha = 0.005$ $SDC = 0$};
\node [below=15mm, align=flush center,text width=5cm] at (27.9,-36.7){$WCDE = 0.012 \; [-0.018,0.042]$ {\color{Maroon}$p = 0.445$}};
\end{tikzpicture} \hspace{5mm}
    \begin{tikzpicture}[scale=0.12]
\tikzstyle{every node}+=[inner sep=0pt]
\draw [CornflowerBlue,very thick,fill=CornflowerBlue,fill opacity=0.2] (27.9,-25.9) circle (3);
\draw (27.9,-25.9) node {$TY$};
\draw [black,very thick] (40.3,-26.4) circle (3);
\draw (40.3,-26.4) node {$PC$};
\draw [WildStrawberry,very thick,fill=WildStrawberry,fill opacity=0.2] (15.2,-26.4) circle (3);
\draw (15.2,-26.4) node {$R$};
\draw [black,very thick] (21.7,-35.4) circle (3);
\draw (21.7,-35.4) node {$A$};
\draw [black,very thick] (34.2,-35.4) circle (3);
\draw (34.2,-35.4) node {$S$};
\draw [black,very thick,-{Stealth[width=5pt,length=5pt]}] (37.3,-26.28) -- (30.9,-26.02);
\draw [black,very thick,-{Stealth[width=5pt,length=5pt]}] (18.2,-26.28) -- (24.9,-26.02);
\draw [black,very thick,-{Stealth[width=5pt,length=5pt]}] (23.34,-32.89) -- (26.26,-28.41);
\draw [black,very thick,-{Stealth[width=5pt,length=5pt]}] (32.54,-32.9) -- (29.56,-28.4);
\node [below=19mm, align=flush center,text width=2.5cm] at (27.9,-25.9){\textbf{(E)} $\alpha = 0.01$ $SDC = 1$};
\node [below=27mm, align=flush center,text width=5cm] at (27.9,-25.9){$WCDE = 0.006 \; [-0.024,0.036]$ {\color{Maroon}$p = 0.693$}};
\end{tikzpicture} \hspace{5mm}
    \begin{tikzpicture}[scale=0.12]
\tikzstyle{every node}+=[inner sep=0pt]
\draw [CornflowerBlue,very thick,fill=CornflowerBlue,fill opacity=0.2] (27.9,-25.9) circle (3);
\draw (27.9,-25.9) node {$TY$};
\draw [black,very thick] (27.9,-36.7) circle (3);
\draw (27.9,-36.7) node {$A$};
\draw [black,very thick] (40.3,-26.4) circle (3);
\draw (40.3,-26.4) node {$S$};
\draw [WildStrawberry,very thick,fill=WildStrawberry,fill opacity=0.2] (15.2,-26.4) circle (3);
\draw (15.2,-26.4) node {$R$};
\draw [black,very thick,-{Stealth[width=5pt,length=5pt]}] (27.9,-33.7) -- (27.9,-28.9);
\draw [black,very thick,-{Stealth[width=5pt,length=5pt]}] (37.3,-26.28) -- (30.9,-26.02);
\draw [black,very thick,-{Stealth[width=5pt,length=5pt]}] (18.2,-26.28) -- (24.9,-26.02);
\node [below=7mm, align=flush center,text width=2.5cm] at (27.9,-36.7){\textbf{(F)} $\alpha = 0.05$ $SDC = 1$};
\node [below=15mm, align=flush center,text width=5cm] at (27.9,-36.7){$WCDE = 0.082 \; [0.052,0.113]$ {\color{NavyBlue}$p = 0.000$}};
\end{tikzpicture}
    \caption{LDECC results on the COMPAS dataset at different significance levels ($\alpha = [0.005, 0.01, 0.05]$). DAGs represent the predicted parent set for the outcome only, with all other relations abstracted away. (\textbf{A–C}) Exposure race (\textit{R}; {\color{WildStrawberry}red}) and outcome general recidivism decile score (\textit{DS}; {\color{CornflowerBlue}blue}). Known parents of outcome \textit{DS} are in {\color{Dandelion}yellow}. (\textbf{D–F}) Exposure race and outcome two-year recidivism (\textit{TY}; {\color{CornflowerBlue}blue}). \textit{A} = age; \textit{CD} = charge degree; \textit{DS} = decile score; \textit{PC} = priors count; \textit{R} = race; \textit{S} = sex; \textit{TY} = two-year recidivism. Weighted CDE ($WCDE$) is reported with $p$-value ($p$) and 95\% confidence intervals in brackets.}
    \label{fig:compas_ldecc}
\end{figure}

\clearpage

\subsection{Liver Transplant Allocation}
\label{appendix:liver}




\paragraph{Additional Background} The US liver transplant system has seen several key policy changes designed to optimize organ distribution and improve patient outcomes. In 2002, the \emph{model for end-stage liver disease} (MELD) was adopted to determine patients' priority for liver transplants based on the severity of their illness \citep{malinchoc2000model}. In June 2013, UNOS implemented the \emph{Share 35} policy to increase access to organs for patients with MELD scores $\geq 35$, given the high waitlist mortality in this subgroup \citep{wald2013new}. In January 2016, the MELD-sodium (i.e., MELD-Na) score incorporated serum sodium levels into the calculation \citep{UNOSMeldNa}. In February 2020, the acuity-circle (AC) policy for distributing livers replaced the longstanding donation service area-based policy \citep{UNOSAC}. The AC policy uses a series of concentric circles, centered on the donor hospital, to determine the distribution range for available livers. In July 2023, the calculation formula for MELD was updated to include additional variables, including female sex and serum albumin \citep{UNOSMeld3}. The new model, known as MELD 3.0, was expected to further refine the sorting of transplant candidates based on their medical urgency for transplant \citep{kim2021meld}.

\paragraph{Data Preprocessing and Feature Selection} 

We used the National Standard Transplant Analysis and Research (STAR) dataset for patient-level information on waitlisted transplant candidates, recipients, and donors from the Organ Procurement \& Transplantation Network (OPTN) up to October 2022 \citep{optn2024}. We filtered to include adult patients awaiting or having undergone a whole liver transplant from deceased donors in 2017-2019 and 2020-2022, keeping features relevant for discrimination discovery (Appendix \ref{appendix:liver}). Time frames were selected as stable periods without changes in the UNOS allocation policy. 

We used the following features in the case study. Except for the liver allocation outcome, all features were determined at time of registration or listing. Thus, \ref{assumption:y_no_desc} holds. These features include all the possible parents to the liver allocation outcome available in the OPTN dataset, thus satisfying \ref{assumption:all_parents_y} to the best of our knowledge. For time period 2017-2019, total observations was $n = 21,101$ patients. For time period 2020-2022, total observations was $n = 22,807$ patients. Below, we provide the definition of each variable included in our causal analysis. Summary statistics are reported in Table \ref{tab:liver_summary}. \\

\begin{enumerate}
    \item \textbf{Sex (exposure, protected attribute):} Recipient sex.
    \item \textbf{Liver allocation (outcome):} Did the candidate receive a liver transplant?
    \item \textbf{Recipient blood type:} Recipient blood group at registration.
    \item \textbf{Initial age:} Age in years at time of listing.
    \item \textbf{Ethnicity:} Recipient ethnicity category.
    \item \textbf{Hispanic/Latino:} Is the recipient Hispanic/Latino?
    \item \textbf{Education:} Recipient highest educational level at registration.
    \item \textbf{Initial MELD:} Initial waiting list MELD/PELD lab score.
    \item \textbf{Active exception case:} Was this an active exception case?
    \item \textbf{Exception type:} Type of exception relative to hepatocellular carcinoma (HCC).
    \item \textbf{Diagnosis:} Primary diagnosis at time of listing.
    \item \textbf{Initial status:} Initial waiting list status code.
    \item \textbf{Number of previous transplants:} Number of prior transplants that the recipient received.
    \item \textbf{Weight:} Recipient weight (kg) at registration.
    \item \textbf{Height:} Recipient height at registration.
    \item \textbf{BMI:} Recipient body mass index (BMI) at listing.
    \item \textbf{Payment method:} Recipient primary projected payment type at registration.
    \item \textbf{Region:} Waitlist UNOS/OPTN region where recipient was listed or transplanted.
\end{enumerate}



\clearpage

\begin{table*}[!h]
    \centering
    \begin{adjustbox}{max width=\textwidth}
    \begin{tabular}{p{1.1cm} p{3.6cm} p{3.6cm} p{3.6cm} p{3.6cm}}
    \toprule[1pt]
    & \multicolumn{2}{c}{\textsc{\textbf{Unos Policy (2017–2019)}}} & \multicolumn{2}{c}{\textsc{\textbf{Unos Policy (2020–2022)}}} \\
    \cmidrule(lr){2-3} \cmidrule(lr){4-5}
    & \multicolumn{1}{c}{LD3 ($\alpha = 0.01$)} & \multicolumn{1}{c}{LD3 ($\alpha = 0.05$)} & \multicolumn{1}{c}{LD3 ($\alpha = 0.01$)} & \multicolumn{1}{c}{LD3 ($\alpha = 0.05$)} \\
    \midrule[1pt]
    $\cde$ & 
 {\footnotesize active exception case,
 diagnosis,
 education,
 ethnicity,
 exception type,
 initial age,
 initial MELD,
 region,
 weight} & 
 {\footnotesize
 active exception case,
 diagnosis,
 education,
 ethnicity,
 exception type,
 initial age,
 initial MELD,
 recipient blood type,
 region,
 weight} & 
 {\footnotesize
 active exception case,
 diagnosis,
 education,
 ethnicity,
 initial age,
 initial MELD,
 payment method,
 region,
 weight} & 
 {\footnotesize
 active exception case,
 diagnosis,
 education,
 ethnicity,
 initial age,
 initial MELD,
 payment method,
 region,
 weight} \\
    \midrule
    SDC & 1 & 1 & 1 & 1 \\
    WCDE & 0.033 [0.016,0.049] & 0.025 [0.008,0.042] & 0.032 [0.018,0.047] & 0.032 [0.018,0.047] \\
    $p$-value & 0.000 & 0.003 & 0.000 & 0.000 \\
    \bottomrule[1pt]
    \end{tabular}
    \end{adjustbox}
    \caption{Results obtained with nonparametric $\chi^2$ independence tests and double machine learning. Total features is 18 (Appendix \ref{appendix:liver}). For time period 2017-2019, total observations is $n = 21101$ patients. For time period 2020-2022, total observations is $n = 22807$ patients. With no covariate control, SDC = 1 and the weighted CDE is 0.049 [0.033,0.066] ($p$-value = 0.000).
    }
    \label{tab:liver}
\end{table*}

\vspace{10mm}

\begin{table}[!h]
    \centering
    \begin{adjustbox}{max width=\textwidth}
    \begin{tabular}{l p{3cm} p{3cm} p{3cm} p{3cm}}
    \toprule[1pt]
    & \multicolumn{2}{c}{\textsc{\textbf{Unos Policy (2017–2019)}}} & \multicolumn{2}{c}{\textsc{\textbf{Unos Policy (2020–2022)}}} \\
    \cmidrule(lr){2-3} \cmidrule(lr){4-5}
    & \multicolumn{1}{c}{PC ($\alpha = 0.01$)} & \multicolumn{1}{c}{PC ($\alpha = 0.05$)} & \multicolumn{1}{c}{PC ($\alpha = 0.01$)} & \multicolumn{1}{c}{PC ($\alpha = 0.05$)} \\
    \midrule[1pt]
    $\cde$ & education, recipient blood type
  & None 
  & active exception case, intial MELD, recipient blood type & active exception case, initial age, recipient blood type, region
  \\
    \midrule
    SDC & 1 & 1 & 1 & 1 \\
    WCDE & 0.046 [0.03,0.063] & 0.049 [0.033,0.066] & 0.037 [0.023,0.050] & 0.028 [0.015,0.041] \\
    $p$-value & 0.000 & 0.000 & 0.000 & 0.000 \\
    \bottomrule[1pt]
    \end{tabular}
    \end{adjustbox}
    \caption{SDC assessment and weighted CDE estimation with double machine learning using results from PC Algorithm \citep{spirtes2001causation}, as implemented by \citet{kalisch_estimating_2007}. We define $\cde$ using the union method (PC$_\cup$; Appendix \ref{sec:appendix_baselines}), though results for $\cde$ were exactly the same using PC$_\cap$.}
    \label{tab:liver_pc}
\end{table}

\vspace{10mm}

\begin{table}[!h]
    \centering
    \begin{adjustbox}{max width=\textwidth}
    \begin{tabular}{l p{3cm} p{3cm} p{3cm} p{3cm}}
    \toprule[1pt]
    & \multicolumn{2}{c}{\textsc{\textbf{Unos Policy (2017–2019)}}} & \multicolumn{2}{c}{\textsc{\textbf{Unos Policy (2020–2022)}}} \\
    \cmidrule(lr){2-3} \cmidrule(lr){4-5}
    & \multicolumn{1}{c}{LDECC ($\alpha = 0.01$)} & \multicolumn{1}{c}{LDECC ($\alpha = 0.05$)} & \multicolumn{1}{c}{LDECC ($\alpha = 0.01$)} & \multicolumn{1}{c}{LDECC ($\alpha = 0.05$)} \\
    \midrule[1pt]
    $\cde$ & payment method
  &  payment method, region
  &  active exception case,
 diagnosis,
 initial age,
 initial MELD,
 recipient blood type
  & diagnosis, recipient blood type, region
  \\
    \midrule
    SDC & 1 & 1 & 1 & 1 \\
    WCDE & 0.049 [0.032,0.065] & 0.048 [0.032,0.064] & 0.032 [0.019,0.046] & 0.032 [0.017,0.048] \\
    $p$-value & 0.000 & 0.000 & 0.000 & 0.000 \\
    \bottomrule[1pt]
    \end{tabular}
    \end{adjustbox}
    \caption{SDC assessment and weighted CDE estimation with double machine learning using results from LDECC using the union method for selection $\cde$ (LDECC$_\cup$; Appendix \ref{sec:appendix_baselines}) \citep{gupta_local_2023}. Results for $\cde$ were almost concordant using LDECC$_{\cap}$, with the exception of 2020-2022 when $\alpha = 0.05$ (where the intersection method excluded diagnosis).}
    \label{tab:liver_ldecc}
\end{table}

\begin{table}[!h]
    \centering
    \begin{adjustbox}{max width=\textwidth}
    \begin{tabular}{c c c c c c c c c c c c}
    \toprule
       & & \multicolumn{2}{c}{\textsc{\textbf{LD3}}} & \multicolumn{2}{c}{\textsc{\textbf{LDECC}}} & \multicolumn{2}{c}{\textsc{\textbf{PC}}} & \multicolumn{2}{c}{\textsc{\textbf{Ratio PC/LD3}}} & \multicolumn{2}{c}{\textsc{\textbf{Ratio LDECC/LD3}}} \\
       \cmidrule(lr){3-4} \cmidrule(lr){5-6} \cmidrule(lr){7-8} \cmidrule(lr){9-10} \cmidrule(lr){11-12}
       \textit{Policy} & $\alpha$ & \textit{Time} & \textit{Tests} & \textit{Time} & \textit{Tests} & \textit{Time} & \textit{Tests} & \textit{Time} & \textit{Tests} & \textit{Time} & \textit{Tests} \\
       \midrule
       \parbox[t]{1.5cm}{\multirow{2}{*}{2017-2019}} & 0.01 & \textbf{0.100} & \textbf{75} & 20.304 & 3303 & 237.324 & 35943 & 2373.24 & 479.24 & 203.04 & 44.04 \\
       & 0.05 & \textbf{0.101} & \textbf{75} & 19.947 & 3188 & 231.844 & 36512 & 2295.48 & 486.83 & 197.50 & 42.51 \\
       \midrule
       \parbox[t]{1.5cm}{\multirow{2}{*}{2020-2022}} & 0.01 & \textbf{0.107} & \textbf{75} & 47.402 & 6959 &  465.278 & 63225 & 4348.39 & 843.00 & 443.01 & 92.79 \\
       & 0.05 & \textbf{0.090} & \textbf{71} & 519.707 & 69926 & 528.372  & 72517 & 5870.80 & 1021.37 & 5774.52 & 984.87 \\
    \bottomrule
    \end{tabular} 
    \end{adjustbox}
    \vspace{3mm}
    \caption{Total number of independence tests performed and runtimes in seconds for LD3 vs PC \citep{kalisch_estimating_2007} and LDECC \citep{gupta_local_2023} on the liver transplant allocation dataset. While PC is worst-case exponential in time complexity \citep{spirtes2001causation}, LDECC is worst-case exponential for some structures and polynomial for others \citep{gupta_local_2023}. This could explain the wide variation in LDECC runtimes across experimental settings.}
    \label{tab:ld3_pc_time_tests}
\end{table}

\begin{table}[!h]
    \centering
    \begin{adjustbox}{max width=0.9\textwidth}
    \begin{tabular}{l c c c c}
    \toprule[1pt]
        & \multicolumn{4}{c}{\textsc{\textbf{Unos Policy (2017-2019)}}} \\
        \cmidrule(lr){2-5}
        ~ & \textit{Female} (\textit{n} = 7679) & \textit{Male} (\textit{n} = 13422) & \textit{p-value} & \textit{Test} \\
        \midrule
        \textit{Active exception case} & 0.36 (0.73) & 0.48 (0.83) & 7.241e-28 & t-test \\ 
        \textit{Diagnosis 1 (PSC: Primary Sclerosing Cholangitis)} & 0.03 (0.18) & 0.04 (0.2) & 0.037 & $\chi^2$ \\
        \textit{Diagnosis 6 (AHF: acute hepatic failure)} & 0.06 (0.23) & 0.02 (0.15) & 0.004 & $\chi^2$ \\ 
        \textit{Diagnosis 7 (Cancer)} & 0.09 (0.28) & 0.16 (0.37) & 0.010 & $\chi^2$ \\ 
        \textit{Height} & 161.9 (7.46) & 175.91 (8.51) & 0.000 & t-test \\ 
        \textit{Initial MELD} & 20.5 (10.21) & 18.83 (9.87) & 1.588e-31 & t-test \\ 
       \textit{ Payment method} & 0.53 (0.5) & 0.54 (0.5) & 0.012 & $\chi^2$ \\
        \textit{Recipient age }& 54.46 (12.42) & 56.03 (10.74) & 4.091e-22 & t-test \\ 
        \textit{Weight} & 75.97 (18.53) & 90.63 (19.59) & 0.000 & t-test \\
    \toprule[1pt]
        & \multicolumn{4}{c}{\textsc{\textbf{Unos Policy (2020-2022)}}} \\
        \cmidrule(lr){2-5}
        ~ & \textit{Female} (\textit{n} = 8574) & \textit{Male} (\textit{n} = 14233) & \textit{p-value} & Test \\
        \midrule
        \textit{Active exception case} & 0.39 (0.58) & 0.43 (0.63) & 3.629e-07 & t-test \\
        \textit{Ethnicity 9 (Multiracial, non-hispanic)} & 0.01 (0.08) & 0.0 (0.07) & 0.022 & Fisher's exact \\ 
        \textit{Exception type 1 (Unknown)} & 0.29 (0.45) & 0.28 (0.45) & 0.022 & $\chi^2$ \\ 
        \textit{Height} & 161.96 (7.8) & 176.36 (8.25) & 0.000 & t-test \\ 
        \textit{Initial MELD} & 22.13 (10.52) & 20.99 (10.48) & 1.862e-15 & t-test \\ 
        \textit{Initial status} & 0.04 (0.2) & 0.02 (0.12) & 2.858e-31 & t-test \\ 
        \textit{Recipient age} & 53.83 (12.75) & 54.74 (11.64) & 3.059e-08 & t-test \\ 
        \textit{Weight} & 75.76 (18.71) & 91.1 (20.32) & 0.000 & t-test \\ 
    \bottomrule[1pt]
    \end{tabular}
    \end{adjustbox}
    \vspace{3mm}
    \caption{Mean values (standard deviations) for features with statistically significant differences between males and females ($\alpha = 0.05$). Summary statistics for all features are available on GitHub (\texttt{https://anonymous.4open.science/r/LD3-4440}).}
    \label{tab:liver_summary}
\end{table}

\end{document}